\documentclass[]{jair}

\setcopyright{cc}
\copyrightyear{2026}
\acmDOI{10.1613/jair.1.22539} % submission number; JAIR DOIs follow 10.1613/jair.1.<submission no.> (verified against article 21715)

\JAIRAE{Marek Petrik}
\JAIRTrack{} % Insert JAIR Track Name Here only if part of a special track
\acmVolume{87}
\acmArticle{18}
\acmMonth{10}
\acmYear{2026}

\RequirePackage[
  datamodel=acmdatamodel,
  style=acmauthoryear,
  backend=biber,
  giveninits=true,
  uniquename=init
  ]{biblatex}

\usepackage{tikz}
\usetikzlibrary{arrows.meta,automata,calc,fit,matrix,patterns,positioning,shadows,shapes.arrows,shapes.geometric,svg.path}
\usepackage{marvosym}
\usepackage[marvosym]{tikzsymbols}
\usepackage{xspace}
\usepackage{subcaption}
\usepackage[capitalise,noabbrev]{cleveref}
\usepackage{amsmath}

\usepackage{amsthm}%
\usepackage{mathrsfs}%
\usepackage[title]{appendix}%
\usepackage{textcomp}%
\usepackage{manyfoot}%
\usepackage{booktabs}%
\usepackage{longtable}
\usepackage{algorithm}%
\usepackage{algorithmicx}%
\usepackage[noend]{algpseudocode}%
\usepackage{listings}%

\usepackage{url}

\usepackage{latexsym}

\usepackage{pifont}

\usepackage{mathtools}
\usepackage{enumerate}
\usepackage{enumitem}
\usepackage{multirow}
\usepackage{multicol}
\usepackage{siunitx}
\usepackage{etoolbox}
\usepackage{rotating}

\usepackage{arydshln}
\usepackage{makecell}

\newtheorem*{remarkcustom}{Remark}

\newtheorem{example}{Example}

\newtheorem{assumption}{Assumption}  % unnumbered version

\newcommand{\set}[1]{\{#1\}}

\newcommand{\simplex}[1]{\Delta(#1)}
\newcommand{\arbitraryset}{\mathcal{X}}
\newcommand{\empseq}[1]{{#1}^\ast}
\newcommand{\nonempseq}[1]{{#1}^+}
\newcommand{\powerset}[1]{2^{#1}}
\DeclareMathOperator{\codeif}{\mathtt{:-} }

\newcommand{\noisy}[1]{\Tilde{#1}}
\newcommand{\event}[1]{{#1}=\top}
\newcommand{\notevent}[1]{{#1}=\bot}
\newcommand{\eventdetected}[1]{\noisy{#1}=\top}
\newcommand{\noteventdetected}[1]{\noisy{#1}=\bot}
\newcommand{\singleproposition}{l}
\newcommand{\singlepropositiondetected}{\tilde{l}}

\newcommand{\trace}{\lambda}
\newcommand{\noisytrace}{\tilde{\lambda}}
\newcommand{\sampletrace}{\lambda'}
\newcommand{\states}{\mathcal{S}}
\newcommand{\statee}{s}
\newcommand{\stateterm}{\termfunction^T}
\newcommand{\stategoal}{\termfunction^G}
\newcommand{\initialstate}{s_0}

\newcommand{\actions}{\mathcal{A}}
\newcommand{\action}{a}
\newcommand{\rewardfunction}{r}
\newcommand{\termfunction}{\tau}
\newcommand{\discountfactor}{\gamma}
\newcommand{\propositions}{\mathcal{P}}
\newcommand{\noisysensordetections}{\noisy{\propositions}}
\newcommand{\labelingfunction}{\mathcal{L}}
\newcommand{\noisylabelingfunction}{\widetilde{\mathcal{L}}}
\newcommand{\rmstates}{U}
\newcommand{\rmstate}{u}
\newcommand{\transitionprob}{p}
\newcommand{\rmstatebelief}{\tilde{u}}

\newcommand{\initialrmstate}{\rmstate_0}
\newcommand{\acceptingrmstate}{\rmstate_A}

\newcommand{\rejectingrmstate}{\rmstate_R}

\newcommand{\rmtransitionfunction}{\delta^\rmstate}
\newcommand{\observationlabel}{L}

\newcommand{\shapingreward}{r_s}
\newcommand{\reals}{\mathbb{R}}

\newcommand{\probability}[1]{P(#1)}
\newcommand{\append}[2]{#1 \oplus #2}
\newcommand{\history}{h}
\newcommand{\rewardmachine}{\mathcal{RM}}
\newcommand{\rmstatetransitionfun}{\delta^u}
\newcommand{\rmrewardtransitionfunbelief}{\tilde{\delta}^r}
\newcommand{\rmrewardtransitionfun}{\delta^r}
\newcommand{\rmrewardtransitionfunRS}{\tilde{\delta}^\text{RS}}
\newcommand{\policy}{\pi}
\newcommand{\step}{t}

\newcommand{\rmtraversal}[2]{{#1}(#2)}  
\newcommand{\rmtraversalelem}{v}  

\newcommand{\thesystem}{\mbox{\textsc{Prob-IRM}}\xspace}
\newcommand{\isa}{\mbox{\textsc{ISA}}\xspace}
\newcommand{\recurrentbaseline}{\textsc{Recurrent PPO}\xspace}

\newcommand{\logicaltransitionfn}{\varphi}
\newcommand{\notlogicaltransitionfn}{\hat{\varphi}}
\newcommand{\logicaltransitiondisjunct}{\textnormal{disj}_i}
\newcommand{\numdisjuncts}[1]{|\logicaltransitionfn#1|}

\newcommand{\maxrmstates}{\rmstates_{\max}}
\newcommand{\traceoriginmultiplier}{w_{t_o}}

\newcommand{\asp}[1]{M(#1)}
\newcommand{\aspp}[2]{M_{#1}(#2)}

\newcommand{\exampleset}{E}
\newcommand{\modebias}{S_M}
\newcommand{\rmrelearningthreshold}{\beta}
\newcommand{\trajectory}{\mathcal{T}}
\newcommand{\rmlearnercls}{\textsc{RMLearner}}
\newcommand{\rmlearner}{\texttt{rm\_learner}}

\newcommand{\maxedgesconst}{\kappa}
\newcommand{\numsamplespertrace}{n_{\sampletrace}}

\newcommand{\ilaspbackground}{B}
\newcommand{\hypothesis}{H}
\newcommand{\exid}{e_{id}}
\newcommand{\expen}{e_{pen}}
\newcommand{\penaltythreshold}{t_{\expen}}
\newcommand{\exinc}{e_{inc}}
\newcommand{\exexc}{e_{exc}}
\newcommand{\exctx}{e_{ctx}}

\newcommand{\classweight}{w_c}

\newcommand{\false}{\bot}
\newcommand{\true}{\top}

\newcommand{\ow}{\textsc{OfficeWorld}\xspace}
\newcommand{\owcoffee}{\textnormal{\Coffeecup}\xspace}
\newcommand{\owagent}{\Strichmaxerl[1.25]}
\newcommand{\owcella}{\ensuremath{A}\xspace}
\newcommand{\owcellb}{\ensuremath{B}\xspace}
\newcommand{\owcellc}{\ensuremath{C}\xspace}
\newcommand{\owcelld}{\ensuremath{D}\xspace}
\newcommand{\owdecor}{\ensuremath{\ast}\xspace}
\newcommand{\owoffice}{\ensuremath{o}\xspace}
\newcommand{\owmail}{\textnormal{\Letter}\xspace}
\newcommand{\owcoffeetask}{\textsc{Coffee}\xspace}
\newcommand{\owcoffeemailtask}{\textsc{CoffeeMail}\xspace}
\newcommand{\owvisit}{\textsc{VisitABCD}\xspace}

\newcommand{\wwrgbtask}{\textsc{RGB}\xspace}
\newcommand{\wwrgminusbtask}{\textsc{RG-B}\xspace}
\newcommand{\wwrgbctask}{\textsc{RGBC}\xspace}

\newcommand{\owup}{\textnormal{up}}
\newcommand{\owdown}{\textnormal{down}}
\newcommand{\owleft}{\textnormal{left}}
\newcommand{\owright}{\textnormal{right}}
\newcommand{\ownoop}{\textnormal{noop}}

\newcommand{\ww}{\textsc{WaterWorld}\xspace}

\definecolor{gold}{RGB}{255,215,0}
\definecolor{goldenrod}{RGB}{218,165,32}
\definecolor{maroon}{rgb}{0.5, 0.0, 0.0}

\tikzset{
	in place/.style={
		auto=false,
		fill=white,
		inner sep=2pt,
	},
	cascaded/.style = {%
		general shadow = {%
			shadow scale = 1,
			shadow xshift = -1ex,
			shadow yshift = 1ex,
			draw,
			fill = white},
		general shadow = {%
			shadow scale = 1,
			shadow xshift = -.5ex,
			shadow yshift = .5ex,
			draw,
			fill = white},
		fill = white, 
		draw,
		minimum width = 1.5cm,
		minimum height = 2cm
	},
	none/.style = {
		inner sep=0mm
	},
	every loop/.append style={-Latex}
}

\newcommand{\bernoulli}[1]{Ber(#1)}

\newcommand{\hd}{\hdashline[0.5pt/5pt]}

\begin{document}

\title{Robust Reward Machine Learning and Execution in Noisy Environments}

%% used to denote shared contribution to the research and/or corresponding author.

\author{Roko Parac}
\orcid{0009-0009-2916-2410}
\email{roko.parac18@ic.ac.uk}

\affiliation{%
  \institution{Imperial College London}
  \city{London}
  \country{United Kingdom}
}

\author{Lorenzo Nodari}
\orcid{0009-0008-7997-8170}
\email{lorenzo.nodari@unibs.it}
\affiliation{%
  \institution{University of Brescia}
  \city{Brescia}
  \country{Italy}
}

\author{Daniel Furelos-Blanco}
\orcid{0000-0001-7461-1910}
\email{daniel.furelos.blanco@gmail.com}
\affiliation{%
  \institution{Imperial College London}
  \city{London}
  \country{United Kingdom}
}

\author{Leo Ardon}
\orcid{0000-0003-4400-7127}
\email{leo.ardon19@ic.ac.uk}
\affiliation{%
  \institution{Imperial College London}
  \city{London}
  \country{United Kingdom}
}

\author{Federico Cerutti}
\orcid{0000-0003-0755-0358}
\email{federico.cerutti@unibs.it}
\affiliation{%
  \institution{University of Brescia}
  \city{Brescia}
  \country{Italy}
}

\author{Alessandra Russo}
\orcid{0000-0002-3318-8711}
\email{a.russo@ic.ac.uk}
\affiliation{%
  \institution{Imperial College London}
  \city{London}
  \country{United Kingdom}
}

\renewcommand{\shortauthors}{Parac, Nodari, Furelos-Blanco, Ardon, Cerutti \& Russo}

\begin{abstract}
{\bf Background:} 
    Reward machines (RMs) represent non-Markovian reward functions as finite-state machines over high-level propositional events. Traditionally, these events are generated by a deterministic labelling function that maps environment states to proposition sets. However, real-world labelling functions are inherently noisy.

    {\bf Objectives:}
    Design and implement a method that learns RMs in the presence of noisy labelling functions and exploits them for policy learning in both discrete and continuous environments.
    
    {\bf Methods:}
    We propose \thesystem, a method that learns RMs from noisy proposition histories using inductive logic programming (ILP) and exploits them for policy learning. Noisy proposition histories are represented as training examples for ILP, which learns an RM that maximises coverage of low-uncertainty histories. RM learning is interleaved with policy optimisation. \thesystem uses a belief over RM states to determine when to relearn the RM and to provide additional guiding rewards.
   
    {\bf Results:}
    \thesystem is evaluated on environments with discrete and continuous state spaces under varying levels of labelling noise. \thesystem performs comparably to baselines using handcrafted RMs and outperforms approaches that use recurrent neural networks to encode history, despite operating with noisy labels.
    
    {\bf Conclusions:}
    \thesystem represents a step towards deploying reward machines in noisy, real-world environments.
\end{abstract}

%% JAIR Note: 
%% Do not include ACM CCS Concepts or Keywords

%% To be updated by authors.
\received{15 April 2026}
%\received[revised]{15 April 2026}
\received[accepted]{01 June 2026}

\maketitle

\section{Introduction}
Reinforcement learning~\cite[RL;][]{SuttonB18} is a machine learning paradigm that models how agents learn to make decisions by interacting with an environment, improving their behaviour through experience. The objective of RL is to learn a policy that maximises cumulative reward over time. RL has played a central role in several landmark AI advances, including AlphaGo~\cite{SilverHMGSDSAPL16} and the training of large language models~\cite{ouyang2022training}. However, despite these successes, challenges remain in extending RL to a broader range of real-world applications~\cite{DBLP:journals/ml/Dulac-ArnoldLML21}. Key challenges include operating robustly under partial observability and noisy observations, generalising across tasks, and improving policy interpretability. 

Reward machines~\cite[RMs;][]{ToroIcarteKVM18} constitute an effective method for addressing some of these challenges. They are finite-state machines that represent non-Markovian reward functions in terms of propositional high-level events abstracting environment states. An RM serves as memory for the agent, decomposing a complex task into a structured sequence of intermediate sub-tasks. By augmenting the environment state space with RM states, the reward function becomes Markovian, enabling the application of standard RL algorithms. The structure of an RM is interpretable and, when combined with appropriate techniques such as reward shaping or hierarchical RL, can mitigate sparse reward challenges and facilitate policy transfer across related tasks.

Recent studies~\cite{ToroIcarteWKVCM19,XuGAMNTW20,FurelosBlancoLJBR21,HasanbeigJAMK21} have introduced methods for learning RMs, reducing manual engineering effort and enhancing their applicability across tasks. However, most assume an oracle that perfectly detects high-level events; that is, a \emph{perfect labelling function} mapping environment states to propositions. This assumption is a key limitation, as real-world event detection through sensors or AI models is rarely perfect due to technological and environmental constraints. Crucially, any errors in detecting high-level events lead to learning incorrect RMs.

Our goal is to devise a method that handles noisy sensor detections when learning and exploiting RMs for non-Markovian reward tasks. 
We introduce \thesystem (\textbf{Prob}abilistic \textbf{I}nduction of \textbf{R}eward \textbf{M}achines), the first method that directly learns and exploits RMs when the labelling function is \emph{noisy}. The agent's belief about event occurrences is modelled using a Bayesian posterior conditioned on sensor outputs. To learn RMs from noisy traces, \thesystem uses ILASP~\cite{law2015ilasp}, a state-of-the-art inductive logic programming system capable of learning from noisy examples. Pivotal to \thesystem is the interleaving of RM and policy learning: a new RM is learned whenever the RL
agent generates a trace not believed to be accepted by the current RM. To accelerate policy learning, \thesystem employs a probabilistic reward shaping formulation based on posterior beliefs derived from the traces. This paper extends our prior work~\cite{parac2024learning} with the following novel contributions:
\begin{itemize}
    \item \emph{Complete ILASP encoding for learning RMs from noisy traces.}
    We present the ILASP encoding for the RM learning task, clearly identifying the changes needed compared to learning from noise-free traces~\cite{FurelosBlancoLJBR21}. 
    \item \emph{Enhanced generality and scalability.} 
    Our prior work used tabular action-value functions, requiring the discretisation of RM state beliefs and limiting the applicability to discrete state spaces. We address this by approximating policies with neural networks trained using Proximal Policy Optimisation~\cite[PPO;][]{schulman2017proximal}, enabling seamless integration with continuous domains and improved robustness~\cite{yu2022surprising}. We also modified the architecture to enable parallel trace collection, significantly improving scalability.
    \item \emph{Extensive evaluation and ablation study.} 
    We evaluate \thesystem on tasks with discrete and continuous state spaces and sparse non-Markovian rewards, providing in-depth ablation analysis of design choices.
    We compare against baselines using handcrafted RMs and a fully neural baseline~\cite[\recurrentbaseline;][]{schulman2017proximal,recurrentppostablebaselines}.
    Results demonstrate that across diverse noise configurations, \thesystem performs comparably to the handcrafted-RM baselines and outperforms \recurrentbaseline. 
\end{itemize}

The paper is organised as follows. \Cref{sec:background} provides background on RL, RMs, and ILASP. \Cref{sec:noisy_labelled_mdps} formalises the noisy event detection setting used throughout. \Cref{sec:learning_rms_from_noisy_traces} describes our method for learning RMs from noisy traces. \Cref{section:rm_exploitation} discusses RM exploitation, including probabilistic reward shaping. \Cref{sec:interleaving_learning_and_exploitation} explains how we interleave RM and policy learning. \Cref{experiments} presents experimental results and ablation studies. \Cref{sec:related} discusses related work, \Cref{sec:discussion} highlights limitations, and \Cref{sec:conclusions} concludes the paper and proposes suggestions for future work.

\section{Background}
\label{sec:background}
This section provides background on reinforcement learning, reward machines, and inductive logic programming. We first establish notation: given a set $\arbitraryset$, $\simplex{\arbitraryset}$ is the probability simplex over $\arbitraryset$, $\empseq{\arbitraryset}$ denotes (possibly empty) sequences of elements from $\arbitraryset$, $\nonempseq{\arbitraryset}$ denotes non-empty sequences of elements from $\arbitraryset$, and $\powerset{\arbitraryset}$ is the power set of $\arbitraryset$. We use $\oplus$ to denote the concatenation of two sequences. The complete notation used throughout the paper is given in \Cref{app:notation}.

\subsection{Reinforcement Learning}
\label{Reinforcementlearning}
We formalise RL tasks as \emph{labelled Markov decision processes}~\cite{FuT14,FurelosBlancoLJBR23}.
\begin{definition}
[Labelled Markov decision process]
A labelled Markov decision process (MDP) is a tuple $\langle\states, \initialstate,\allowbreak\actions,\allowbreak\transitionprob,\allowbreak\rewardfunction,\allowbreak\termfunction,\allowbreak\discountfactor,\allowbreak\propositions,\allowbreak\labelingfunction \rangle$ where:
\begin{itemize}
 \item $\states$ is a set of states,
 \item $\initialstate \in \states$ is an initial state,
 \item $\actions$ is a set of actions,
 \item $\transitionprob:\states\times\actions\to\simplex{\states}$ is a probability transition function,
 \item $\rewardfunction:\nonempseq{(\states\times\actions)}\times\states\to\mathbb{R}$ is a reward function,
 \item $\termfunction:\empseq{(\states\times\actions)}\times\states\to\set{\false,\true}\times\set{\false,\true}$ is a termination function,
 \item $\discountfactor\in[0,1)$ is a discount factor,
 \item $\propositions$ is a finite set of \emph{propositions} representing high-level events, and 
 \item $\labelingfunction:\states\times\actions\times\states\to\powerset{\propositions}$ is a \emph{(perfect) labelling function} mapping state-action-state triplets to sets of propositions in $\propositions$. 
 We refer to these sets as \emph{labels}.
\end{itemize}
\label{def:labeled-mdp}
\end{definition}
The transition function $\transitionprob$ is Markovian, whereas the reward function $\rewardfunction$ and termination function $\termfunction$ are not (i.e.,~they are history-dependent). 
The termination function $\termfunction$ returns a pair $\langle \stateterm, \stategoal \rangle$, where $\stateterm$ indicates whether the current history is terminal, and $\stategoal$ indicates whether the goal is achieved.

Given a state-action \emph{history} $\history_\step=\langle \statee_0,\allowbreak\action_0,\allowbreak\ldots,\allowbreak\statee_\step\rangle \in \empseq{(\states\times\actions)}\times\states$, a \emph{trace} $\trace_\step=\langle\labelingfunction(\emptyset,\emptyset,\statee_0),\allowbreak\ldots,\allowbreak\labelingfunction(\statee_{\step-1},\action_{\step-1},\statee_\step)\rangle\in\nonempseq{(\powerset{\propositions})}$ assigns a label to each transition in $\history_\step$, where $\labelingfunction(\emptyset,\emptyset,\statee_0)$ represents the initial state label. The goal is to find a \emph{policy} $\policy:\nonempseq{(\powerset{\propositions})}\times\states\to\simplex{\actions}$ that maps trace-state pairs to a probability distribution over actions that maximises the expected cumulative discounted reward (or \emph{return}) $R_\step=\mathbb{E}_\policy[\sum^n_{k=\step}\discountfactor^{k-\step}\rewardfunction(\history_{k+1})]$, where $n$ is the episode's last step. The labelling function must preserve sufficient information from histories such that reward and termination can be determined from the corresponding traces. If so, the reward and termination functions can be written in terms of traces instead of history as $\rewardfunction(\history_{\step+1})=\rewardfunction(\trace_{\step+1},\statee_{\step+1})$ and $\termfunction(\history_\step)=\termfunction(\trace_\step,\statee_\step)$, respectively.

The agent-environment interaction is as follows. At time $\step$, the (label) trace is $\trace_\step\in\nonempseq{(\powerset{\propositions})}$ and the agent observes a tuple $\langle \statee_\step, \stateterm_\step, \stategoal_\step\rangle$, where $\statee_\step\in\states$ is the state, and $(\stateterm_\step,\stategoal_\step)=\termfunction(\trace_\step,\statee_\step)$ is the termination information with $\stateterm_\step\in\set{\false,\true}$ and $\stategoal_\step\in\set{\false,\true}$ indicating termination and goal status, respectively.
The agent also observes a label $\observationlabel_\step=\labelingfunction(\statee_{\step-1},\action_{\step-1},\statee_\step)$. If the history is non-terminal, the agent takes an action $\action_\step\in\actions$, and the environment transitions to state $\statee_{\step+1} \sim \transitionprob(\cdot\mid\statee_\step,\action_\step)$. The agent then observes a new tuple $\langle \statee_{\step+1}, \stateterm_{\step+1}, \stategoal_{\step+1}\rangle$ and label $\observationlabel_{\step+1}$, extends the trace as $\trace_{\step+1}=\trace_\step \oplus \observationlabel_{\step+1}$, and receives reward $\rewardfunction(\trace_{\step+1},\statee_{\step+1})$.

Traces can be classified based on the terminal and goal indicators at time $\step$: a trace $\trace_\step$ is a \emph{goal trace} if $\langle\stateterm_\step,\stategoal_\step\rangle=\langle\true,\true\rangle$, a \emph{dead-end trace} if $\langle\stateterm_\step,\stategoal_\step\rangle=\langle\true,\false\rangle$, and an \emph{incomplete trace} if $\stateterm_\step=\false$.

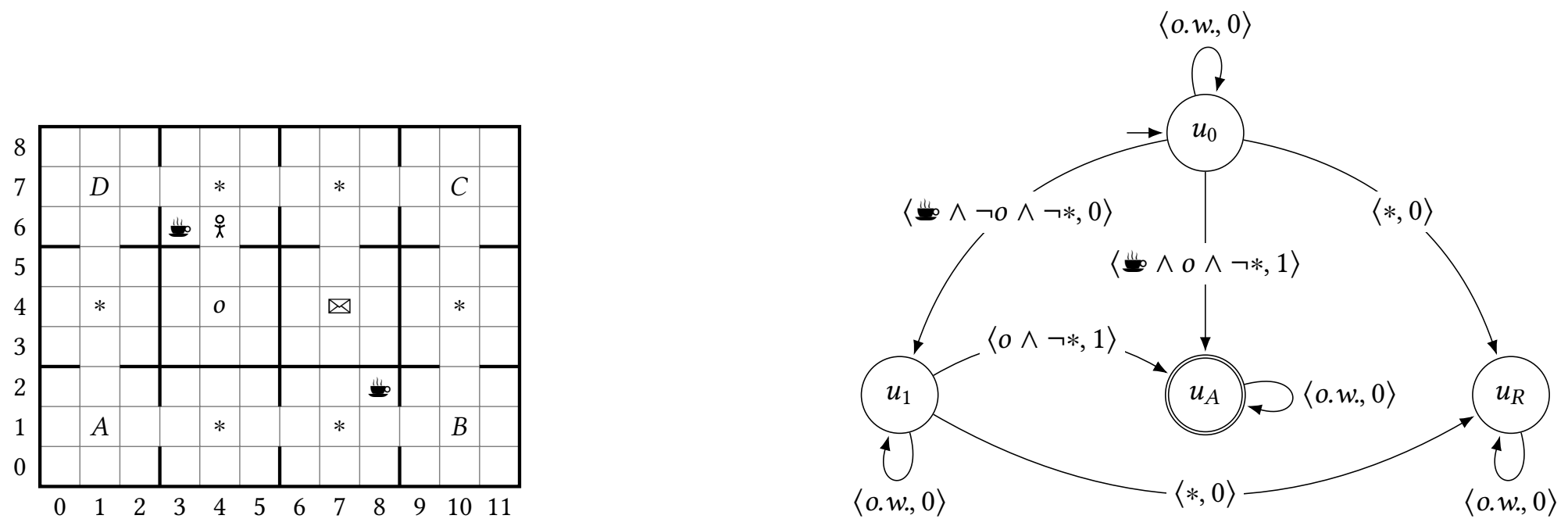
\begin{figure*}
    \centering
    \begin{subfigure}[b]{0.495\linewidth}
        \centering
        \resizebox{0.6\linewidth}{!}{
        \tikzset{digit/.style = { minimum height = 5mm, minimum width=5mm, anchor=center }}
\newcommand{\setcell}[3]{\edef\x{#2 - 0.5}\edef\y{9.5 - #1}\node[digit,name={#1-#2}] at (\x, \y) {#3};}
\begin{tikzpicture}[scale=0.5]
    \draw[white] (0, 0) grid (13, 10);
    
    \setcell{9}{2}{0} \setcell{9}{3}{1} \setcell{9}{4}{2} \setcell{9}{5}{3} \setcell{9}{6}{4} \setcell{9}{7}{5} \setcell{9}{8}{6} \setcell{9}{9}{7} \setcell{9}{10}{8} \setcell{9}{11}{9} \setcell{9}{12}{10} \setcell{9}{13}{11}
    
    \setcell{8}{1}{0} \setcell{7}{1}{1} \setcell{6}{1}{2} \setcell{5}{1}{3} \setcell{4}{1}{4} \setcell{3}{1}{5} \setcell{2}{1}{6} \setcell{1}{1}{7} \setcell{0}{1}{8}
    
    \draw[gray] (1, 1) grid (13, 10);
    \draw[very thick, scale=3] (1/3, 1/3) rectangle (13/3, 10/3);
    
    \draw[very thick, scale=1] (4, 1) rectangle (4, 2); \draw[very thick, scale=1] (4, 9) rectangle (4, 10);
    \draw[very thick, scale=1] (4, 3) rectangle (4, 8);
    \draw[very thick, scale=1] (7, 1) rectangle (7, 2); \draw[very thick, scale=1] (7, 3) rectangle (7, 8); \draw[very thick, scale=1] (7, 9) rectangle (7, 10);
    \draw[very thick, scale=1] (10, 1) rectangle (10, 2); \draw[very thick, scale=1] (10, 3) rectangle (10, 8); \draw[very thick, scale=1] (10, 9) rectangle (10, 10);
    \draw[very thick, scale=1] (1, 4) rectangle (2, 4); \draw[very thick, scale=1] (3, 4) rectangle (11, 4); \draw[very thick, scale=1] (12, 4) rectangle (13, 4);
    \draw[very thick, scale=1] (1, 7) rectangle (2, 7); \draw[very thick, scale=1] (3, 7) rectangle (5, 7); \draw[very thick, scale=1] (6, 7) rectangle (8, 7); \draw[very thick, scale=1] (9, 7) rectangle (11, 7); \draw[very thick, scale=1] (12, 7) rectangle (13, 7);
    \setcell{1}{3}{\owcelld} \setcell{1}{6}{\owdecor} \setcell{1}{9}{\owdecor} \setcell{1}{12}{\owcellc}
    \setcell{2}{5}{\owcoffee} \setcell{2}{6}{\owagent}
    \setcell{4}{3}{\owdecor} \setcell{4}{6}{\owoffice} \setcell{4}{9}{\owmail} \setcell{4}{12}{\owdecor}
    \setcell{6}{10}{\owcoffee}
    \setcell{7}{3}{\owcella} \setcell{7}{6}{\owdecor} \setcell{7}{9}{\owdecor} \setcell{7}{12}{\owcellb}
\end{tikzpicture}
        }
    \end{subfigure}
    \hfill
    \begin{subfigure}[b]{0.495\linewidth}
        \centering
        \resizebox{0.8\linewidth}{!}{
	\begin{tikzpicture}[shorten >=1pt,node distance=2.06cm,on grid,auto,every initial by arrow/.style ={-Latex}]
            \node[state,initial,initial text=] (u_0)   {$\initialrmstate$};
            \node[state,accepting] (u_acc) [below =3cm of u_0]  {$\acceptingrmstate$};
            \node[state] (u_1) [left =3.5cm of u_acc]   {$\rmstate_1$};
            \node[state] (u_rej) [right =3.5cm of u_acc]  {$\rejectingrmstate$};
            
            \path[-Latex] (u_0) edge [loop above] node {$\langle \emph{o.w.}, 0 \rangle$} ();
            \path[-Latex] (u_1) edge [loop below] node {$\langle \emph{o.w.}, 0 \rangle$} ();
            \path[-Latex] (u_acc) edge [loop right] node {$\langle \emph{o.w.}, 0 \rangle$} ();
            \path[-Latex] (u_rej) edge [loop below] node {$\langle \emph{o.w.}, 0 \rangle$} ();
            
            \path[-Latex] (u_0) edge [bend right] node[in place] {$ \langle \owcoffee \land \neg \owoffice \land \neg \owdecor, 0 \rangle$} (u_1);
            \path[-Latex] (u_0) edge [bend left] node[in place] {$\langle \owdecor, 0 \rangle$} (u_rej);
            \path[-Latex] (u_1) edge[bend right] node[in place] {$\langle \owdecor, 0 \rangle$} (u_rej);
            \path[-Latex] (u_1) edge[bend left] node[in place] {$\langle \owoffice\land\neg\owdecor, 1 \rangle$} (u_acc);
            \path[-Latex] (u_0) edge node[in place] {$\langle \owcoffee \land \owoffice \land\neg\owdecor, 1 \rangle$} (u_acc);
         \end{tikzpicture}
         }
    \end{subfigure}
    \caption{An \ow instance (left) and a reward machine for the \owcoffeetask task (right), where \emph{o.w.} stands for \emph{otherwise}.}
    \Description{Left: a 12 by 9 grid world with walls, in which special cells are marked with symbols for the coffee machines, the mail, the office, the locations A, B, C and D, and decorations, together with the agent. Right: a reward machine with four states: the initial state u0, an intermediate state u1, the accepting state uA and the rejecting state uR. Observing coffee without the office or a decoration moves u0 to u1 with reward 0; observing the office without a decoration moves u1 to uA with reward 1; observing coffee and the office together without a decoration moves u0 directly to uA with reward 1; observing a decoration moves u0 or u1 to uR with reward 0; every state has a self-loop labelled otherwise with reward 0.}
    \label{fig:ow_rm}
\end{figure*}

\begin{example}
\label{example-domain} 
The \ow~\cite{ToroIcarteKVM18} domain, illustrated in \Cref{fig:ow_rm} (left), is a $12\times 9$ grid labelled with designated special locations. At each step, the agent $\owagent$ observes its current position in the grid and moves in one of the four cardinal directions; that is, $\states=\set{0,\ldots,11}\times\set{0,\ldots,8}$ and $\actions=\set{\textnormal{up},\textnormal{down},\textnormal{left},\textnormal{right}}$. The agent always moves in the intended direction and remains in place when attempting to move into a wall. The set of propositions $\propositions=\set{\owcoffee,\owmail,\owoffice,\owcella,\owcellb,\owcellc,\owcelld,\owdecor}$ consists of the special locations. The labelling function maps a $\langle s,a,s'\rangle\in\states\times\actions\times\states$ triplet to the set of propositions observed in $s'$, e.g.,~$\labelingfunction(\langle 4,6\rangle,\textnormal{left},\langle 3,6\rangle)=\set{\owcoffee}$. 
Let us consider the \owcoffeetask task, which consists of going to the coffee machine (\owcoffee) and then to the office (\owoffice).
A reward of 1 is obtained for completing the task; otherwise, the reward is 0.
Rewards are non-Markovian since the current state (i.e.,~position on the grid) alone cannot determine the reward. The history $\history_4=\langle\langle4,6\rangle,\allowbreak\textnormal{left},\allowbreak\langle3,6\rangle,\allowbreak\textnormal{right},\allowbreak\langle4,6\rangle,\allowbreak\textnormal{down},\allowbreak\langle4,5\rangle,\allowbreak\textnormal{down},\allowbreak\langle4,4\rangle\rangle$ yields a reward of 1 and is mapped to the goal trace $\lambda_4=\langle\set{},\allowbreak\set{\owcoffee},\allowbreak\set{},\allowbreak\set{},\allowbreak\set{\owoffice}\rangle$.
\end{example}

\subsection{Reward Machines}
\label{sec:background_rms}
Reward machines provide a structured representation of reward functions by encoding task structure as a finite-state machine over propositional events.

\begin{definition}[Reward machine]
A \emph{reward machine}~\cite[RM;][]{ToroIcarteKVM18,ToroIcarteKVM22} is a tuple $\rewardmachine=\langle\rmstates,\propositions,\rmstatetransitionfun,\rmrewardtransitionfun, \initialrmstate,\acceptingrmstate,\rejectingrmstate\rangle$, where:
\begin{itemize}
\item $\rmstates$ is a set of states, 
\item $\propositions$ is a set of propositions forming the RM's alphabet,
\item $\rmstatetransitionfun:\rmstates\times\powerset{\propositions} \to \rmstates$ is a state-transition function,
\item $\rmrewardtransitionfun:\rmstates\times\rmstates\to\mathbb{R}$ is a reward-transition function,
\item $\initialrmstate, \acceptingrmstate, \rejectingrmstate \in\rmstates$ are the initial, accepting, and rejecting states, respectively.
\end{itemize}
\end{definition}

The accepting state indicates successful task completion, while the rejecting state indicates unrecoverable failure.

\begin{example}
    \Cref{fig:ow_rm} (right) shows the RM for the \owcoffeetask task in \ow introduced in \Cref{example-domain}. Edges are labelled with propositional formulas over $\propositions=\set{\owcoffee,\owmail,\owoffice,\owcella,\owcellb,\owcellc,\owcelld,\owdecor}$ and associated transition rewards. To verify whether a formula is satisfied by a label $\observationlabel\in\powerset{\propositions}$, $\observationlabel$ is used as a truth assignment: propositions contained in the label are true, and all others are false. For example, the label $\set{\owcoffee}$ satisfies the formula $\owcoffee\land\neg \owoffice\land\neg\owdecor$.
\end{example}

During the agent-environment interaction, the agent tracks its current state in the RM alongside its environment state. Starting from the RM's initial state $\initialrmstate$, the agent transitions through RM states according to the state-transition function $\rmstatetransitionfun$ based on observed labels, receiving rewards via the reward-transition function $\rmrewardtransitionfun$.

Given an RM $\rewardmachine$ and a trace $\trace=\langle\observationlabel_0,\ldots,\observationlabel_n\rangle$, a \emph{traversal} $\rmtraversal{\rewardmachine}{\trace}=\langle v_0, v_1, \ldots, v_{n+1} \rangle$ is the unique sequence of RM states such that $v_0=\initialrmstate$, and $v_{i+1}=\rmtransitionfunction(v_i,\observationlabel_i)$ for $i=0,\ldots,n$.
Traversals of goal traces terminate in $\acceptingrmstate$, traversals of dead-end traces terminate in $\rejectingrmstate$, and traversals of incomplete traces terminate in neither.

\begin{example}
    Given the RM in \Cref{fig:ow_rm} (right) and the goal trace $\trace=\langle\set{},\allowbreak\set{\owcoffee},\allowbreak\set{},\allowbreak\set{},\allowbreak\set{\owoffice}\rangle$, the traversal is $\langle\initialrmstate,\rmstate_0,\rmstate_1,\rmstate_1,\rmstate_1,\acceptingrmstate\rangle$. As expected for a goal trace, the traversal terminates in the accepting state.
    \label{ex_trace}
\end{example}

To verify that an RM correctly encodes task structure, we check whether its traversals align with trace categories. An RM $\rewardmachine$ is \emph{valid} with respect to trace $\trace = \langle \observationlabel_0, \ldots, \observationlabel_n \rangle$ if
\begin{itemize}
\item $\rmtraversalelem_{n+1} = \acceptingrmstate$ and the trace $\trace$ is a \emph{goal} trace,
\item $\rmtraversalelem_{n+1} = \rejectingrmstate$ and the trace $\trace$ is a \emph{dead-end} trace, or
\item $\rmtraversalelem_{n+1} \notin \set{\acceptingrmstate, \rejectingrmstate}$ and the trace $\trace$ is an \emph{incomplete} trace,
\end{itemize}
where $\rmtraversalelem_{n+1}$ is the last RM state in the traversal $\rmtraversal{\rewardmachine}{\trace}$.

Reward machines are compact trace representations: each RM state encodes how far the agent has advanced toward the goal. Consequently, the reward function becomes Markovian over the joint space of MDP and RM states $\states\times\rmstates$. This enables the application of standard RL algorithms such as Q-learning~\cite{watkins1992q}, DDPG~\cite{LillicrapHPHETS16}, and PPO~\cite{schulman2017proximal}. These algorithms have been successfully applied to RM-based learning in prior work~\cite[e.g.,][]{ToroIcarteKVM18,ToroIcarteKVM22,li2024reward,ardon2024texttt}.

\subsection{Learning ASP Programs from (Noisy) Examples}
\label{sec:background_ilasp}
Our work follows the methodology proposed by \citet{FurelosBlancoLJBR21} for learning RMs using \emph{inductive logic programming}~\cite[ILP;][]{muggleton1991inductive}. Specifically, we encode RMs using \emph{answer set programming}~\cite[ASP;][]{lifschitz2019answer} and learn them using ILASP~\cite{law2015ilasp}, a state-of-the-art ILP system. We briefly introduce ASP and ILASP's learning framework below.

\subsubsection{Answer Set Programming~(ASP)}
ASP is a declarative programming paradigm for knowledge representation and reasoning. In ASP, a problem is encoded as a logic program, and the solutions correspond to the models of that program, called answer sets. 

Our ASP programs consist of normal rules, constraints, and facts. 
\emph{Normal rules} have the form $\mathtt{h \codeif b_1, \ldots, b_n}$, where the \emph{head} $\mathtt{h}$ is an atom and the \emph{body} $\mathtt{b_1,\ldots, b_n}$ consists of literals. 
A literal is either an atom $\mathtt{a}$ or its negation $\mathtt{not\ a}$. \emph{Constraints} are rules of the form $\mathtt{\codeif b_1,\ldots, b_n}$.
\emph{Facts} are rules with an empty body, written as $\mathtt{h.}$ 

Given a set of ground atoms (or interpretation) $I$, a ground normal rule is satisfied if the head is satisfied by $I$ whenever all body literals are satisfied by $I$. A ground constraint is satisfied if its body is not satisfied by $I$. Facts are always satisfied.

An \emph{answer set} is defined via the reduct. The \emph{reduct} $P^{I}$ of a ground program $P$ with respect to an interpretation $I$ is constructed by removing from $P$ all rules containing $\mathtt{not\ a}$ where $\mathtt{a} \in I$, and removing all $\mathtt{not\ a}$ literals where $\mathtt{a} \notin I$. 
Then, $I$ is an \emph{answer set} of $P$ if and only if $I$ satisfies the rules in $P^I$ and no proper subset of $I$ does. 

\begin{example}
\label{ex:asp}
Consider the following ASP program:
\begin{align*}
    P=\left\lbrace
    \begin{array}{l}
    \mathtt{a \codeif b, not \; c.} \\
    \mathtt{b.}
    \end{array}
    \right\rbrace.
\end{align*}
For the interpretation $I = \set{\mathtt{a, b}}$, the reduct $P^I$ is:
\begin{align*}
\left\lbrace
\begin{array}{l}
\mathtt{a \codeif b.} \\
\mathtt{b.}
\end{array}
\right\rbrace.
\end{align*}
Since $I$ satisfies all rules in $P^I$ and no proper subset of $I$ (e.g., $\set{\mathtt{a}}, \set{\mathtt{b}}$, and $\emptyset$) does, $I$ is the unique answer set of $P$.
\end{example}

\subsubsection{Learning from Noisy Answer Sets}
\label{sec:learning_las_with_noise}
\emph{Learning from answer sets}~\cite[LAS;][]{law2020ilasp} is a paradigm for learning ASP programs from examples. We focus on learning from \emph{noisy examples}, where each example is weighted with a penalty. The goal is to find an ASP program (called a \emph{hypothesis}) that minimises a score that accounts for program complexity (e.g.,~number of literals) and the penalties incurred by uncovered examples.

Examples are \emph{weighted context-dependent partial interpretations}~\cite[WCDPIs;][]{law2018inductive}, each a tuple $\langle \exid, \expen, \langle \exinc,\exexc\rangle,\exctx\rangle$, where $\exid$ is a unique identifier; $\expen$ is a penalty; $\langle \exinc,\exexc \rangle$ is a pair of atom sets known as a \emph{partial interpretation}; and $\exctx$ is an ASP program known as the \emph{context}, providing example-specific information. An ASP program $P$ \emph{covers} a WCDPI $\langle \exid, \expen, \langle \exinc,\exexc\rangle,\exctx\rangle$ if and only if there is an answer set $A$ of $P \cup \exctx$ such that $\exinc\subseteq A$ and $\exexc \cap A = \emptyset$.

These WCDPI examples are used within a LAS task that also specifies background knowledge and the space of learnable rules. We formalise LAS tasks~\cite{law2018inductive} as follows.

\begin{definition} [LAS task] \label{def:lnas}
  A LAS task $T$ is a tuple of the form $\langle B, S_M, E\rangle$, where $B$ is an ASP program known as background knowledge, $S_M$ is a hypothesis space defining the set of learnable ASP rules, and $E$ is a set of WCDPI examples. Given a hypothesis $H \subseteq S_M$,
  \begin{enumerate}
    \item $uncov(H, T)$ is the set of all examples $e \in E$ such that $H$ does not cover $e$;
    \item the penalty of $H$, denoted as $pen(H, T)$, is the sum $\sum_{e \in uncov(H, T)} e_{pen}$;
    \item the length of $H$, denoted as $length(H)$, is the sum $\sum_{r \in H} length(r)$, where $length(r)\in\mathbb{N}_{>0}$;
    \item the score of $H$, denoted as $\mathcal{S}(H, T)$, is the sum $length(H) + pen(H, T)$;
    \item $H$ is an inductive solution of $T$ iff $\mathcal{S}(H, T)$ is finite; and
    \item $H$ is an \emph{optimal inductive solution} of $T$ iff $\mathcal{S}(H, T)$ is finite and $\nexists H' \subseteq S_M$ such that $\mathcal{S}(H, T) > \mathcal{S}(H', T)$.
  \end{enumerate}
  \label{def:ilasp_scoring}
\end{definition}

Examples with infinite penalty ($\expen = \infty$) are non-noisy and must be covered; finite penalties allow trading off coverage for simpler hypotheses. We use ILASP~\cite{law2015ilasp}, a state-of-the-art system for learning ASP programs from noisy examples.

\begin{example}
\label{ilasp_example}
Consider the LAS tasks $T_1=\langle B, S_M, E_1 \rangle$ and $T_2=\langle B, S_M, E_2 \rangle$, where:
\begin{align*}
B &= \left\{\mathtt{b \codeif c.}\right\}, 
&
S_M &= \left\{
\begin{array}{l}
\mathtt{a.} \\
\mathtt{\codeif b.} \\
\mathtt{\codeif not \; b.} \\
\mathtt{a \codeif b.} \\
\mathtt{a \codeif not \; b.}
\end{array}
\right\},
&
\begin{array}{r@{\;}l}
E_1 &= \left\{
\begin{array}{l}
\langle \text{ex1}, \infty, \langle \{\mathtt{a}\}, \emptyset \rangle, \{\mathtt{c.}\} \rangle, \\
\langle \text{ex2}, \infty, \langle \emptyset, \{\mathtt{a}\} \rangle, \emptyset \rangle
\end{array}
\right\}, \\[3ex]
E_2 &= \left\{
\begin{array}{l}
\langle \text{ex1}, 1, \langle \{\mathtt{a}\}, \emptyset \rangle, \{\mathtt{c.}\} \rangle, \\
\langle \text{ex2}, 1, \langle \emptyset, \{\mathtt{a}\} \rangle, \emptyset \rangle
\end{array}
\right\}.
\end{array}
\end{align*}
The optimal inductive solution for $T_1$ is $H_1 = \{ \mathtt{a} \codeif \mathtt{b}. \}$, with score $\mathcal{S}(H_1, T_1)=2$ since $pen(H_1, T_1)=0$ (all examples are covered) and $length(H_1)=2$.

For $T_2$, the examples are identical but have finite penalties. The empty hypothesis $H_2=\emptyset$ achieves a better score: $\mathcal{S}(H_2, T_2) = 1$ since $pen(H_2, T_2)=1$ (ex1 is uncovered) and $length(H_2)=0$. While this example yields a trivial solution, it illustrates the key principle: when examples have finite penalties, the learner can sacrifice coverage of low-confidence examples to find simpler and more general hypotheses.
\end{example}

\section{Noisy Labelled Markov Decision Processes}
\label{sec:noisy_labelled_mdps}

The labelled MDP defined in \Cref{sec:background} assumes a perfect labelling function $\labelingfunction$. In practice, however, event detection through sensors or learned models is inherently noisy. We formalise this setting through a \emph{noisy labelled MDP}, where the key change is replacing the perfect labelling function with a probabilistic one.

\begin{definition}[Noisy labelled Markov decision process]
\label{def:noisy-labeled-mdp}
A noisy labelled Markov decision process is a tuple $\langle\states, \initialstate,\actions,\transitionprob,\rewardfunction,\termfunction,\discountfactor,\propositions,\noisylabelingfunction \rangle$ where $\states,\initialstate,\actions,\transitionprob,\rewardfunction,\termfunction,\discountfactor,\propositions$ are as in \Cref{def:labeled-mdp}, and $\noisylabelingfunction$ is a noisy labelling function.
\end{definition}

The noisy labelling function maps each transition to a probability distribution over propositions, capturing sensor uncertainty. 
We use $\event{\singleproposition}$ and $\notevent{\singleproposition}$ to denote the actual presence and absence of proposition $\singleproposition$, and $\eventdetected{\singleproposition}$ and $\noteventdetected{\singleproposition}$ to denote sensor detection and non-detection, respectively.

\begin{definition}[Noisy labelling function]
	\label{def:noisylabelfun}
	The noisy labelling function $\noisylabelingfunction$ maps each transition to the probability of each proposition:
	\begin{align*}
		\noisylabelingfunction(s_t,a_t,s_{t+1}) = \set{\langle\singleproposition, [\noisylabelingfunction(\statee_{t}, \action_{t}, \statee_{t+1})]_\singleproposition \rangle \mid \singleproposition \in \propositions},
	\end{align*}
	where $[\noisylabelingfunction(\statee_{t}, \action_{t}, \statee_{t+1})]_\singleproposition=\probability{\event{\singleproposition} \mid s_t,a_t,s_{t+1}}$ is the probability of observing $\singleproposition$ on transition $\langle s_t,a_t,s_{t+1}\rangle$.
\end{definition}

For compactness, we write pairs as $\singleproposition: p, \ldots$ rather than $\langle\singleproposition, p\rangle$, omitting propositions with probability zero. For instance, if $[\noisylabelingfunction(\statee_0, \action_0, \statee_1)]_\owcoffee = 0.65$ and $[\noisylabelingfunction(\statee_0, \action_0, \statee_1)]_\owoffice = 0.01$, with all other propositions having probability zero, we write $\noisylabelingfunction(\statee_0, \action_0, \statee_1) = \{\owcoffee: 0.65, \owoffice: 0.01\}$.

A \emph{noisy trace} is the probabilistic analogue of a trace, where each label is replaced by its probability distribution.

\begin{definition}[Noisy trace]
	Given a state-action history $\history_t = \langle s_0, a_0, s_1, a_1, \ldots, s_t \rangle$, the noisy trace $\noisytrace_{t}$ is:
	\begin{align*}
		\noisytrace_{t} &= \langle \noisylabelingfunction(\emptyset, \emptyset, \statee_0), 
		\noisylabelingfunction(\statee_0, \action_0, \statee_1),
		\ldots,
		\noisylabelingfunction(\statee_{\step - 1}, \action_{\step - 1}, \statee_\step) \rangle.
	\end{align*}
\end{definition}

We make the following assumptions about this setting:

\begin{assumption}[Conditional independence]
	Given a transition $\langle \statee_t, \action_t, \statee_{t+1} \rangle$, let $\noisysensordetections_{t+1} \in \powerset{\propositions}$ denote the set of propositions detected by the sensors at this transition. We assume that propositions are conditionally independent given their respective sensor detections. Thus, the probability of each proposition depends only on its own sensor output:
	\begin{align*}
		P(l=\top\mid \statee_t, \action_t, \statee_{t+1}) = P(l=\top \mid \noisysensordetections_{t+1}) = P(l=\top \mid\tilde{l}_{t+1}).
	\end{align*}
	\label{assumption:independence}
\end{assumption}

\begin{assumption}[Deterministic termination]
	The termination function $\termfunction$ remains deterministic, so trace termination is determined with certainty.
	\label{assumption:termination}
\end{assumption}

\begin{assumption}[Reward structure]
	The reward function $\rewardfunction$ assigns a reward of $1$ to goal traces, $-1$ to dead-end traces, and $0$ to incomplete traces.
	\label{assumption:reward}
\end{assumption}

\Cref{assumption:independence} is common in noisy labelling models~\cite{DingSBR11,li2024reward,verginis2024joint}. \Cref{assumption:termination} ensures that, despite noisy labels, the agent knows with certainty when an episode terminates. This is reasonable when termination is determined by a supervising agent (e.g., a human) who can accurately assess task completion. \Cref{assumption:reward}, adopted from prior work on RM learning~\cite[e.g.,][]{ArdonFR23,FurelosBlancoLJBR23}, allows us to focus on learning RM structure rather than rewards.

We now describe one concrete way to obtain the noisy labelling function using a binary sensor model. The agent is equipped with a binary sensor for each proposition $\singleproposition \in \propositions$, characterised by its sensitivity and specificity of detection. The \emph{sensitivity} of the sensor is the probability of detection given actual presence: $\probability{\eventdetected{\singleproposition}\mid\event{\singleproposition}}$. The \emph{specificity} is the probability of non-detection given actual absence:  $\probability{\noteventdetected{\singleproposition}\mid\notevent{\singleproposition}}$. Since the sensor output may not always match the true environment state, the agent computes a posterior belief over proposition occurrence using Bayes' rule:
\begin{align}
\probability{l=\top\mid\noisy{l}=y} =
\frac{\probability{\noisy{l}=y\mid l=\top}\probability{l=\top}}
{\sum_{x \in \set{\top, \bot}}
\probability{\noisy{l}=y\mid l=x} \probability{l=x}},
\label{eq:bayes_rule}
\end{align}
where $y \in \set{\top, \bot}$ is the sensor outcome, $\probability{\event{\singleproposition}}$ is the prior probability of proposition $\singleproposition$ being present, and the probabilities $\probability{\noteventdetected{\singleproposition}\mid\event{\singleproposition}}$, $\probability{\eventdetected{\singleproposition}\mid\notevent{\singleproposition}}$ and $\probability{\notevent{\singleproposition}}$ follow from sensitivity, specificity and the prior. We experimentally determine priors using the relative occurrence of the propositions in the environment.

\begin{remarkcustom}
	The choice of sensor model is orthogonal to the rest of the method. Any model that provides $\probability{\event{\singleproposition} \mid \statee_t, \action_t, \statee_{t+1}}$ for every $\singleproposition \in \propositions$ is a suitable candidate. The binary sensor model with fixed sensitivity and specificity presented here is closely related to that of \citet{ghasemi2020task}, with the key distinction that we do not update the sensor model as new information becomes available.
\end{remarkcustom}

\begin{example}
\label{ex:noisy_labeling_fn}
Consider the \owcoffeetask task from \Cref{example-domain} with $\propositions = \set{\owcoffee, \owoffice}$ and the history $\history=\langle\langle4,6\rangle,\allowbreak\textnormal{left},\allowbreak\langle3,6\rangle\rangle$. The priors are $\probability{\event{\owcoffee}} = \frac{2}{9 \cdot 12}$ and $\probability{\event{\owoffice}} = \frac{1}{9 \cdot 12}$ (two \owcoffee and one \owoffice in a $9\times 12$ grid). Both sensors have sensitivity and specificity $0.99$. We compute the posteriors for each step using \Cref{eq:bayes_rule}:
\begin{itemize}
    \item Step $t=0$. Suppose no propositions are detected ($\noisysensordetections_0 = \emptyset$):
    \allowdisplaybreaks
    \begin{align*}
        \probability{\owcoffee=\top\mid\noisy{\owcoffee}=\bot} &= \frac{\probability{\noisy{\owcoffee}=\bot\mid \owcoffee=\top}\probability{\owcoffee=\top}}
        {\sum\limits_{x \in \set{\top, \bot}}
        \probability{\noisy{\owcoffee}=\bot\mid \owcoffee=x} \probability{\owcoffee=x}}\\
        &= \frac{(1 - 0.99)\cdot\frac{2}{9\cdot12}}{(1 - 0.99)\cdot\frac{2}{9\cdot12} + 0.99\cdot\left(1 - \frac{2}{9\cdot12}\right)} \simeq 0.0002, \\
        \probability{\owoffice=\top\mid\noisy{\owoffice}=\bot} &= \frac{(1 - 0.99)\cdot\frac{1}{9\cdot12}}{(1 - 0.99)\cdot\frac{1}{9\cdot12} + 0.99\cdot\left(1 - \frac{1}{9\cdot12}\right)} \simeq 0.0001.
    \end{align*}
    Thus $\noisylabelingfunction(\emptyset, \emptyset, \langle4,6\rangle) = \set{\owcoffee: 0.0002,\; \owoffice: 0.0001}$.
    \item Step $t=1$. Suppose the coffee sensor fires ($\noisysensordetections_1 = \set{\owcoffee}$):
    \allowdisplaybreaks
    \begin{align*}
        \probability{\owcoffee=\top\mid\noisy{\owcoffee}=\top} &= \frac{0.99\cdot\frac{2}{9\cdot12}}{0.99\cdot\frac{2}{9\cdot12} + (1-0.99)\cdot\left(1 - \frac{2}{9\cdot12}\right)} \simeq 0.6513, \\
        \probability{\owoffice=\top\mid\noisy{\owoffice}=\bot} &\simeq 0.0001.
    \end{align*}
    Thus $\noisylabelingfunction(\langle4,6\rangle, \textnormal{left}, \langle3,6\rangle) = \set{\owcoffee: 0.6513,\; \owoffice: 0.0001}$.
\end{itemize}

The resulting noisy trace is therefore $\noisytrace_1 = \langle 
\set{\owcoffee: 0.0002,\; \owoffice: 0.0001}, 
\set{\owcoffee: 0.6513,\; \owoffice: 0.0001}
\rangle$.
\end{example}

\section{Learning Reward Machines from Noisy Traces}
\label{sec:learning_rms_from_noisy_traces}

In this section, we present our approach for learning RMs from noisy traces using ILASP. We assume a fixed dataset of noisy traces is given and later describe the interleaved collection and learning process in \Cref{sec:interleaving_learning_and_exploitation}. 

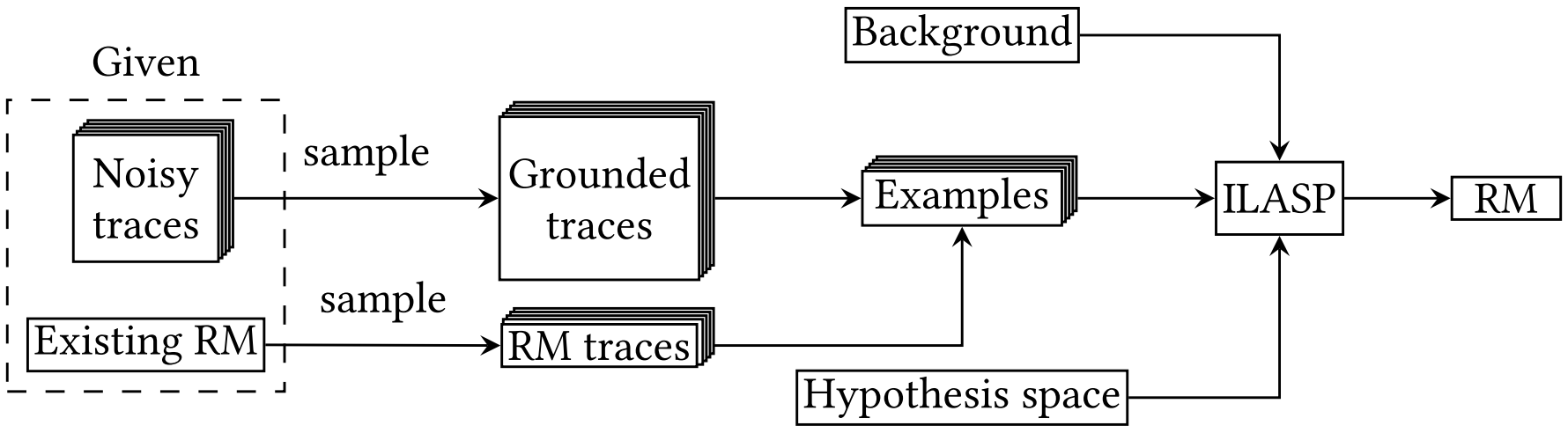
\begin{figure*}
	\centering
	\tikzstyle{none}=[inner sep=0mm]
	\tikzstyle{arrow}=[->,>=stealth]
	\tikzstyle{node}=[fill=white, draw=black, shape=rectangle, minimum width=0.4cm, minimum height=0.2cm, text centered, font=\scriptsize, inner sep=1pt]
	
	\resizebox{0.75\linewidth}{!}{
		\begin{tikzpicture}
		% ILASP box (center)
		\node [node, minimum width=0.7cm, minimum height=0.4cm] (ilasp) at (0.75, 0) {ILASP};
		
		% Background box (top input)
		\node [node, minimum width=0.6cm] (background) at (-1.0, 0.9) {Background};
		
	% Hypothesis Space box (bottom input)
	\node [node, minimum width=0.6cm] (hypothesis) at (-1.0, -1.1) {Hypothesis space};
		
	% Noisy Traces (cascaded boxes - 5 boxes, stacking to top right)
	% Background box for noisy traces
	\node [fill=white, draw=black, shape=rectangle, minimum width=0.8cm, minimum height=0.7cm, text centered, font=\scriptsize, inner sep=1pt] (noisy_traces_back4) at (-5.5+0.08, 0+0.08) {};
	\node [fill=white, draw=black, shape=rectangle, minimum width=0.8cm, minimum height=0.7cm, text centered, font=\scriptsize, inner sep=1pt] (noisy_traces_back3) at (-5.5+0.06, 0+0.06) {};
	\node [fill=white, draw=black, shape=rectangle, minimum width=0.8cm, minimum height=0.7cm, text centered, font=\scriptsize, inner sep=1pt] (noisy_traces_back2) at (-5.5+0.04, 0+0.04) {};
	\node [fill=white, draw=black, shape=rectangle, minimum width=0.8cm, minimum height=0.7cm, text centered, font=\scriptsize, inner sep=1pt] (noisy_traces_back1) at (-5.5+0.02, 0+0.02) {};
	\node [fill=white, draw=black, shape=rectangle, minimum width=0.8cm, minimum height=0.7cm, text centered, font=\scriptsize, inner sep=1pt, align=center] (noisy_traces) at (-5.5, 0) {Noisy\\traces};
	
	% Previous RM box (below noisy traces), connects to Examples
	\node [node, minimum width=0.9cm] (previous_rm) [below=0.3cm of noisy_traces] {Existing RM};
	
	% Rectangle outline around noisy traces and previous RM with "Given" label
	\node [draw=black, dashed, inner sep=3pt, fit=(noisy_traces) (noisy_traces_back4) (previous_rm), label={[font=\scriptsize]above:Given}] (given_box) {};
		
	% Grounded Traces (more cascaded boxes - 5 boxes, stacking to top right)
	% Background box for grounded traces
	\node [fill=white, draw=black, shape=rectangle, minimum width=1.1cm, minimum height=0.9cm, text centered, font=\scriptsize, inner sep=1pt] (grounded_traces_back4) at (-3.0+0.08, 0+0.08) {};
	\node [fill=white, draw=black, shape=rectangle, minimum width=1.1cm, minimum height=0.9cm, text centered, font=\scriptsize, inner sep=1pt] (grounded_traces_back3) at (-3.0+0.06, 0+0.06) {};
	\node [fill=white, draw=black, shape=rectangle, minimum width=1.1cm, minimum height=0.9cm, text centered, font=\scriptsize, inner sep=1pt] (grounded_traces_back2) at (-3.0+0.04, 0+0.04) {};
	\node [fill=white, draw=black, shape=rectangle, minimum width=1.1cm, minimum height=0.9cm, text centered, font=\scriptsize, inner sep=1pt] (grounded_traces_back1) at (-3.0+0.02, 0+0.02) {};
	\node [fill=white, draw=black, shape=rectangle, minimum width=1.1cm, minimum height=0.9cm, text centered, font=\scriptsize, inner sep=1pt, align=center] (grounded_traces) at (-3.0, 0) {Grounded\\traces};
	
	% RM traces
	\node [fill=white, draw=black, shape=rectangle, minimum width=1.1cm, minimum height=0.25cm, text centered, font=\scriptsize, inner sep=1pt] (rm_traces_back4) at (-3.0+0.08, -0.8125+0.08) {};
	\node [fill=white, draw=black, shape=rectangle, minimum width=1.1cm, minimum height=0.25cm, text centered, font=\scriptsize, inner sep=1pt] (rm_traces_back3) at (-3.0+0.06, -0.8125+0.06) {};
	\node [fill=white, draw=black, shape=rectangle, minimum width=1.1cm, minimum height=0.25cm, text centered, font=\scriptsize, inner sep=1pt] (rm_traces_back2) at (-3.0+0.04, -0.8125+0.04) {};
	\node [fill=white, draw=black, shape=rectangle, minimum width=1.1cm, minimum height=0.25cm, text centered, font=\scriptsize, inner sep=1pt] (rm_traces_back1) at (-3.0+0.02, -0.8125+0.02) {};
	\node [node, minimum width=0.9cm] (rm_traces) at (-3.0, -0.8125) {RM traces};
	
		% Examples (cascaded boxes - 5 boxes, all equal size, stacking to top right)
		% Background box for examples
		\node [fill=white, draw=black, shape=rectangle, minimum width=1.1cm, minimum height=0.3cm, text centered, font=\scriptsize, inner sep=1pt] (examples_back4) at (-1.0+0.08, 0+0.08) {};
		\node [fill=white, draw=black, shape=rectangle, minimum width=1.1cm, minimum height=0.3cm, text centered, font=\scriptsize, inner sep=1pt] (examples_back3) at (-1.0+0.06, 0+0.06) {};
		\node [fill=white, draw=black, shape=rectangle, minimum width=1.1cm, minimum height=0.3cm, text centered, font=\scriptsize, inner sep=1pt] (examples_back2) at (-1.0+0.04, 0+0.04) {};
		\node [fill=white, draw=black, shape=rectangle, minimum width=1.1cm, minimum height=0.3cm, text centered, font=\scriptsize, inner sep=1pt] (examples_back1) at (-1.0+0.02, 0+0.02) {};
		\node [fill=white, draw=black, shape=rectangle, minimum width=1.1cm, minimum height=0.3cm, text centered, font=\scriptsize, inner sep=1pt] (examples) at (-1.0, 0) {Examples};
			
		% Output box (right)
		\node [node, minimum width=0.6cm] (output) at (2.0, 0) {RM};
			
			% Arrows from inputs to ILASP
			\draw [arrow] (background.east) -| (ilasp.north);
			\draw [arrow] ([xshift=0.08cm]examples.east) -- (ilasp.west);
			\draw [arrow] (hypothesis.east) -| (ilasp.south);
			
			% Arrow from ILASP to output
			\draw [arrow] (ilasp.east) -- (output.west);
			
		% Noisy Traces -> Grounded Traces
		\draw [arrow] ([xshift=0.08cm]noisy_traces.east) -- node[above, font=\scriptsize] {sample} (grounded_traces.west);
			
			% Grounded Traces -> Examples
			\draw [arrow] ([xshift=0.08cm]grounded_traces.east) -- (examples.west);
			
			% Previous RM -> Examples
			\draw [arrow] ([xshift=0.08cm]rm_traces.east) -| (examples.south);
			
			% Existing RM -> RM traces
			\draw [arrow] (previous_rm.east) -- node[above, font=\scriptsize] {sample} (rm_traces.west);
			
		\end{tikzpicture}
	}
	\caption{Overview of the RM learning pipeline. Traces are sampled from noisy traces and from an existing RM, converted into ILASP examples, and used together with background knowledge and a hypothesis space to learn a new RM.}
	\Description{Flow diagram of the RM learning pipeline. A dashed box labelled Given contains a stack of noisy traces and an existing RM. An arrow labelled sample leads from the noisy traces to a stack of grounded traces, and another arrow labelled sample leads from the existing RM to RM traces. Grounded traces and RM traces both feed a stack of examples. The examples, a background box and a hypothesis space box all feed into ILASP, which outputs an RM.}
	\label{fig:rm_learning_diagram}
\end{figure*}

\Cref{fig:rm_learning_diagram} provides an overview of the pipeline: noisy traces are sampled to produce grounded traces, which are supplemented with traces generated from an existing RM (when available). These traces are converted into ILASP examples with associated penalties, which, together with background knowledge and a hypothesis space, are used by ILASP to learn an RM.

We organise this section as follows. \Cref{sec:example_gen} describes how we generate training examples from noisy traces and existing RMs. \Cref{subsec:rm_encoding} presents the ASP encoding of RMs and traces. Finally, \Cref{subsec:learning_task} defines the complete ILASP learning task, including background knowledge, hypothesis space, example construction, and complexity control mechanisms.

\subsection{Generating Example Traces}
\label{sec:example_gen}
We generate training examples from two sources: grounded traces sampled from noisy traces, and traces derived from an existing RM structure. These are described in the following subsections.

\subsubsection{From Noisy Traces}
\label{sec:example_gen_traces}

A key challenge in learning RMs from noisy traces lies in mapping these traces to WCDPIs. 
We use a sampling-based method: at each step in the trace, for each proposition we sample from a Bernoulli distribution parameterised by its probability. 
This generates multiple grounded traces, each representing a plausible realisation of the noisy trace.

\begin{sloppypar}
Formally, let $\noisytrace=\langle \noisylabelingfunction(\emptyset, \emptyset, \statee_0),\allowbreak \noisylabelingfunction(\statee_0, \action_0, \statee_1),\allowbreak \ldots,\allowbreak \noisylabelingfunction(\statee_{n-1}, \action_{n-1}, \statee_n) \rangle$ be a noisy trace, and let $\noisytrace_{i, \singleproposition} = [\noisylabelingfunction(\statee_{i-1}, \action_{i-1}, \statee_i)]_\singleproposition$ denote the probability of observing proposition $\singleproposition$ at step $i$. 
We generate a \emph{grounded trace} $\sampletrace = \langle \observationlabel_1, \ldots, \observationlabel_n \rangle$ from $\noisytrace$ by treating each proposition occurrence as a Bernoulli trial: $\sampletrace_{i, \singleproposition} \sim \bernoulli{\noisytrace_{i, \singleproposition}}$ for all $\singleproposition \in \propositions$, where $\observationlabel_i = \{\singleproposition \in \propositions \mid \sampletrace_{i, \singleproposition} = 1\}$. The process is performed $\numsamplespertrace$ times for each noisy trace, resulting in $\numsamplespertrace$ grounded traces.
Following prior work on learning RMs~\cite{ToroIcarteWKVCM19,FurelosBlancoLJBR21}, we compress the traces by removing the empty labels and, thereafter, removing contiguous identical labels.
\end{sloppypar}

\begin{example}
	Consider the proposition set $\propositions = \set{\owcoffee, \owoffice}$ 
	and the noisy goal trace $\noisytrace = 
	\langle  
	\set{\owcoffee\!:\! 0.01, \owoffice\!:\! 0.01},\allowbreak
	\set{\owcoffee\!:\! 0.9, \owoffice\!:\! 0.01},\allowbreak
	\set{\owcoffee\!:\! 0.9, \owoffice\!:\! 0.01},\allowbreak
	\set{\owcoffee\!:\! 0.01, \owoffice\!:\! 0.01},\allowbreak
	\set{\owcoffee\!:\! 0.01, \owoffice: 0.9}
	\rangle
	$. Sampling could yield the most likely trace: $\sampletrace= 
	\langle 
	\set{},
	\set{\owcoffee},
	\set{\owcoffee},
	\set{},
	\set{\owoffice}
	\rangle$. After compression (removing empty labels and consecutive duplicates),  $\sampletrace= 
	\langle 
	\set{\owcoffee},
	\set{\owoffice}
	\rangle$.
	\label{wcdpi-example}
\end{example}

Additionally, we generate incomplete examples from prefixes of the compressed goal and dead-end traces, similar to \citet{ArdonFR23}. This provides the learner with information about valid intermediate traces and helps prevent overgeneralisation.

	\subsubsection{From an Existing Reward Machine}
	\label{sec:example_gen_rms}
	
	The examples derived from noisy traces may be insufficient or biased for learning a good RM. This is particularly problematic when traces are collected by a policy that exploits a given RM: such traces concentrate on near-optimal paths and lack diversity. For instance, the agent may only collect goal traces along one path, missing incomplete and dead-end traces that are essential for learning intermediate states and alternative paths. This issue is present when interleaving RM learning with policy optimisation, described in \Cref{sec:interleaving_learning_and_exploitation}.
	
	To address this, we generate additional examples from the exploited RM's structure. These examples represent paths the RM considers valid, supplementing the collected traces and ensuring that the RM's knowledge is retained even when corresponding traces become rare. Given an RM, we generate traces as follows:
	\begin{enumerate}
		\item Enumerate all paths (i.e.,~sequences of edge formulas) from $\initialrmstate$ to $\acceptingrmstate$ and $\rejectingrmstate$.
		\item For each path, generate all subsequences of formulas. 
		\item Convert each subsequence to a trace by taking the minimal label satisfying each formula. The trace is a goal trace if the traversal ends with $\acceptingrmstate$, a dead-end trace if it ends with $\rejectingrmstate$, and an incomplete trace otherwise.
	\end{enumerate}

	Prior work on learning RMs without noise~\cite[e.g.,][]{FurelosBlancoLJBR21,ToroIcarteWKVCM19,HasanbeigJAMK21,XuGAMNTW20} does not require this technique, as those methods satisfy all examples exactly; hence, infrequent but valid traces are never ignored. In our setting, however, infrequent examples may correspond to noise and receive low penalties, risking the loss of valid but rare behaviours.
	
	\begin{figure}
		\centering
		\begin{tikzpicture}[shorten >=1pt,node distance=3.2cm,on grid,auto,every initial by arrow/.style ={-Latex}]
        \node[state,initial,initial text=] (u_0)   {$\initialrmstate$};
        \node[state] (u_1) [right= of u_0] {$\rmstate_1$};
        \node[state] (u_2) [right = of u_1]   {$\rmstate_2$};
        \node[state] (u_3) [right= of u_2]   {$\rmstate_3$};
        \node[state,accepting] (u_{acc}) [right = of u_3]   {$\acceptingrmstate$};
        
        \path[-Latex] (u_0) edge node[in place] {$ \langle \owcella \land \neg \owcellb, 0 \rangle$} (u_1);
        \path[-Latex] (u_1) edge node[in place] {$ \langle \owcellb, 0 \rangle$} (u_2);
        \path[-Latex] (u_2) edge node[in place] {$ \langle \owcellc \land \neg \owcelld, 0 \rangle$} (u_3);
        \path[-Latex] (u_3) edge node[in place] {$ \langle \owcelld, 1 \rangle$} (u_{acc});
        \path[-Latex] (u_0) edge [bend right] node[in place] {$ \langle \owcella \land \owcellb, 0 \rangle$} (u_2);
        \path[-Latex] (u_2) edge[bend right] node[in place] {$ \langle \owcellc \land \owcelld, 1 \rangle$} (u_{acc});
\end{tikzpicture}
		\caption{Reward machine for the \owvisit task used in \Cref{example:traces_derivation_from_rm}. The agent must go to locations $\owcella$, $\owcellb$, $\owcellc$ and $\owcelld$ in order. This RM accounts for co-location: $\owcella$ may appear with $\owcellb$, and $\owcellc$ may appear with $\owcelld$.}
		\Description{A reward machine drawn as a left-to-right chain of five states u0, u1, u2, u3 and the accepting state uA. The edges along the chain are labelled A and not B with reward 0, B with reward 0, C and not D with reward 0, and D with reward 1. Two curved shortcut edges skip a state: from u0 to u2 labelled A and B with reward 0, and from u2 to uA labelled C and D with reward 1.}
		\label{fig:rm_progression_limitation_visit_abcd}
	\end{figure}
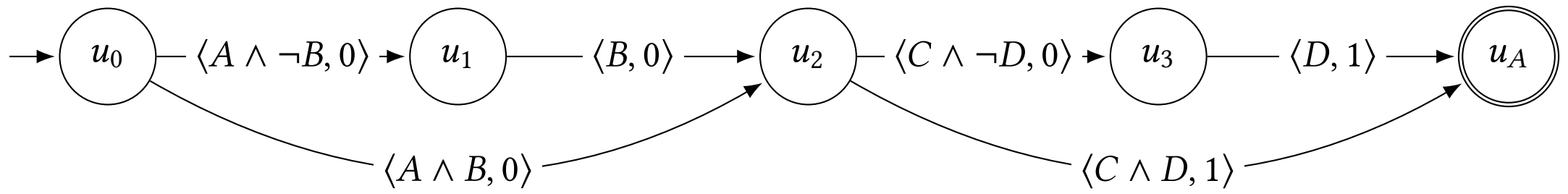
	
	\begin{example}
		\label{example:traces_derivation_from_rm}	
		Consider the RM in \Cref{fig:rm_progression_limitation_visit_abcd} for the \owvisit task in \ow:
		\begin{enumerate}
			\item There are four paths from $\initialrmstate$ to $\acceptingrmstate$:
			\begin{center}
				\begin{tabular}{cccc}
					$\langle \owcella \land \neg\owcellb, \owcellb, \owcellc\land\neg\owcelld, \owcelld\rangle$, & $\langle \owcella\land\owcellb, \owcellc\land\neg\owcelld, \owcelld\rangle$, &
					$\langle \owcella\land\owcellb, \owcellc\land\owcelld\rangle$, & $\langle \owcella\land\neg\owcellb, \owcellb, \owcellc \land \owcelld\rangle$.
				\end{tabular}
			\end{center}
			\item For the path $\langle \owcella\land\owcellb, \owcellc\land\neg\owcelld, \owcelld\rangle$, the subsequences are:
			\begin{center}
				\begin{tabular}{ccccccc}
                    $\langle \owcella\land\owcellb, \owcellc\land\neg\owcelld, \owcelld\rangle$, &
					$\langle \owcella\land\owcellb, \owcellc\land\neg\owcelld\rangle$, & $\langle \owcella\land\owcellb,\owcelld\rangle$,& $\langle\owcellc\land\neg\owcelld, \owcelld\rangle$,& $\langle \owcella\land\owcellb\rangle$, & $\langle \owcellc\land\neg\owcelld\rangle$, & $\langle\owcelld\rangle$.
				\end{tabular}
			\end{center}
			\item Each subsequence becomes a trace using minimal labels. For example, $\langle \owcella \land \owcellb, \owcellc\land\neg\owcelld\rangle$ yields the incomplete trace $\trace=\langle\{\owcella,\owcellb\}, \{\owcellc\}\rangle$.
		\end{enumerate}
	\end{example}

\subsection{ASP Representation of Reward Machines and Traces}
\label{subsec:rm_encoding}
To learn RMs using ILASP, we encode them as ASP programs. The RMs we learn have the same structure as those defined in \Cref{sec:background}, and we learn them from grounded (deterministic) traces rather than noisy traces. Consequently, we build upon the encoding of \citet{FurelosBlancoLJBR21}, which was designed for learning from perfect traces. We first introduce notation for expressing RM transitions as propositional logic formulas, then present the ASP encoding.

The state-transition function of an RM can be expressed using a logical transition function $\logicaltransitionfn:\rmstates\times\rmstates\to\textnormal{DNF}_\propositions$~\cite{FurelosBlancoLJBR21} that maps state pairs to disjunctive normal form (DNF) formulas over $\propositions$, where each disjunct corresponds to an edge. The number of disjuncts in the formula between a pair of states $\rmstate,\rmstate'\in\rmstates$ is denoted $\numdisjuncts{(\rmstate, \rmstate')}$.

\begin{example}
	\label{ex:logical_transition}
	For the RM in \Cref{fig:ow_rm}, we have $\logicaltransitionfn(\rmstate_0, \rmstate_1)=\owcoffee \land \neg \owoffice \land \neg \owdecor$ with $\numdisjuncts{(\rmstate_0, \rmstate_1)}=1$.
\end{example}

\begin{definition}[ASP representation of reward machines]
	\label{def:asp_rm}
Given an RM $\rewardmachine=\langle\rmstates,\propositions,\rmstatetransitionfun,\rmrewardtransitionfun, \initialrmstate,\acceptingrmstate,\rejectingrmstate\rangle$, we encode it as $\asp{\rewardmachine}=\aspp{\rmstates}{\rewardmachine} \cup \aspp{\logicaltransitionfn}{\rewardmachine}$, where:
\begin{equation*}
	\aspp{\rmstates}{\rewardmachine} = \left\lbrace \mathtt{used\_state}(\rmstate). \mid \rmstate \in \rmstates \right\rbrace
\end{equation*}
encodes the RM states with $\mathtt{used\_state}(\rmstate)$ indicating $\rmstate$ is an RM state, and
\begin{equation*}
	M_\logicaltransitionfn(\rewardmachine)=\left\lbrace\begin{array}{@{}l|c@{}}
		\mathtt{ed}(\rmstate,\rmstate',i).                                                                    &  \\
		\notlogicaltransitionfn(\rmstate,\rmstate',i,\mathtt{T) \codeif not~prop(}\singleproposition_1\mathtt{,T), step(T).} & \rmstate \in U\setminus \{\acceptingrmstate,\rejectingrmstate\}, \\
		\multicolumn{1}{c|}{\vdots}                                                                          & \rmstate' \in U \setminus \{\rmstate\}, \\
		\notlogicaltransitionfn(\rmstate,\rmstate',i,\mathtt{T) \codeif not~prop(}\singleproposition_n\mathtt{,T), step(T).} & 1 \leq i \leq |\logicaltransitionfn(\rmstate,\rmstate')|, \\
		\notlogicaltransitionfn(\rmstate,\rmstate',i,\mathtt{T) \codeif prop(}\singleproposition_{n+1}\mathtt{,T), step(T).} & \textnormal{disj}_i \in \logicaltransitionfn(\rmstate,\rmstate'), \\
		\multicolumn{1}{c|}{\vdots}                                                                          & \begin{aligned}\textnormal{disj}_i &= \singleproposition_1 \land \cdots \land \singleproposition_n\\ &\land \neg \singleproposition_{n+1} \land \cdots \land \neg \singleproposition_m\end{aligned} \\
		\notlogicaltransitionfn(\rmstate,\rmstate',i,\mathtt{T) \codeif prop(}\singleproposition_m,\mathtt{T), step(T).}     & 
	\end{array}\right\rbrace
\end{equation*}
encodes transitions with $\logicaltransitiondisjunct$ denoting the $i$-th disjunct of $\logicaltransitionfn(u,u')$, $\mathtt{ed}(\rmstate,\rmstate',i)$ representing the $i$-th edge from $\rmstate$ to $\rmstate'$, $\mathtt{prop}(\singleproposition, \mathtt{T})$ indicating proposition $\singleproposition$ holds at time $\mathtt{T}$, and $\mathtt{step(T)}$ denoting that $\mathtt{T}$ is a timestep.
\end{definition}

The encoding uses $\notlogicaltransitionfn$ (negation of $\logicaltransitionfn$) rather than $\logicaltransitionfn$ directly, as this reduces the search space and speeds up learning~\cite[p.~1046]{FurelosBlancoLJBR21}.

\begin{example}
	\label{ex:asp_rm_encoding}
	For the RM in \Cref{fig:ow_rm}, the transition $\logicaltransitionfn(\rmstate_0, \rmstate_1)=\owcoffee \land \neg \owoffice \land \neg \owdecor$ is encoded as
	\begin{align*}
		\begin{Bmatrix*}[l]
			\mathtt{used\_state}(u_0). \;
			\mathtt{used\_state}(u_1). \; 
			\mathtt{ed}(u_0, u_1, 1). \\
			\notlogicaltransitionfn(u_0, u_1, \mathtt{1, T) \codeif not\;prop(\owcoffee, T), step(T).} \\
			\notlogicaltransitionfn(u_0, u_1, \mathtt{1, T) \codeif prop}(\owoffice, \mathtt{T), step(T).} \\
			\notlogicaltransitionfn(u_0, u_1, \mathtt{1, T) \codeif prop}(\owdecor\mathtt{, T), step(T).}
		\end{Bmatrix*}.
	\end{align*}
\end{example}

We now define the ASP encoding for \emph{traces}, also following \citet{FurelosBlancoLJBR21}.

\begin{definition}[ASP representation of a label trace]
	\label{def:asp_trace_representation}
	Given a label trace $\trace=\langle \observationlabel_0,\ldots,\observationlabel_n\rangle$, $\asp{\trace}$ denotes the set of ASP facts
	\begin{equation*}
		\asp{\trace} = \begin{matrix*}[l]
			\{\mathtt{prop(}\singleproposition, t). \mid 0 \leq t \leq n, \singleproposition \in \observationlabel_t\}~\cup \\
			\{\mathtt{step}(t). \mid 0 \leq t \leq n \}~\cup \\
			\{\mathtt{last(}n\mathtt{).}\},
		\end{matrix*}
	\end{equation*}
	where $\mathtt{last}(n)$ denotes that $n$ is the final timestep.
\end{definition}

Given the RM and trace encodings (\Cref{def:asp_rm,def:asp_trace_representation}), we define rules for RM traversal:
\begin{equation*}
	R_\mathtt{st} = \begin{Bmatrix*}[l]
		\mathtt{st(0},u_0). \\
		\mathtt{st(T\mathord{+}1,Y)} \codeif \mathtt{st(T,X), \delta(X,Y,T).} \\
		\mathtt{accept} \codeif \mathtt{last(T), st(T\mathord{+}1},\acceptingrmstate).\\
		\mathtt{reject} \codeif \mathtt{last(T), st(T\mathord{+}1},\rejectingrmstate).
	\end{Bmatrix*},
\end{equation*}
where $\mathtt{st(T, X)}$ denotes that the state at timestep $\mathtt{T}$ is $\mathtt{X}$; $\mathtt{\delta(X,Y,T)}$ indicates that the transition from $\mathtt{X}$ to $\mathtt{Y}$ is satisfied at timestep $\mathtt{T}$; and $\mathtt{accept}$ and $\mathtt{reject}$ respectively indicate whether the trace is accepted or rejected. The rules encode that: traversals start from the initial state; the next state is determined based on the currently satisfied transition; and that the trace is accepted (resp.~rejected) if the last state is $\acceptingrmstate$ (resp.~$\rejectingrmstate$). The $\mathtt{\delta(X,Y,T)}$ atoms are defined via:
\begin{equation*}
	R_\delta = \begin{Bmatrix*}[l]
		\logicaltransitionfn\mathtt{(X,Y,E,T)} \codeif \mathtt{not~}\notlogicaltransitionfn\mathtt{(X,Y,E,T), ed(X,Y,E), step(T).}\\
		\mathtt{out\_\logicaltransitionfn(X,T)} \codeif \mathtt{\logicaltransitionfn(X,\_,\_,T).}\\
		\mathtt{\delta(X,Y,T)} \codeif \mathtt{\logicaltransitionfn(X,Y,\_,T).}\\
		\mathtt{\delta(X, X, T)} \codeif \mathtt{not~out\_\logicaltransitionfn(X,T), used\_state(X), step(T).}
	\end{Bmatrix*},
\end{equation*}
which encode (i)~$\logicaltransitionfn$ as the negation of $\notlogicaltransitionfn$; (ii)~whether an outgoing transition from state $\mathtt{X}$ is satisfied at timestep $\mathtt{T}$; (iii)~$\mathtt{\delta(X,Y,T)}$ holds when the formula on an edge from $\mathtt{X}$ to $\mathtt{Y}$ is satisfied at timestep $\mathtt{T}$; and (iv)~$\mathtt{\delta(X,X,T)}$ holds (self-loop) when no outgoing transition from $\mathtt{X}$ is satisfied at timestep $\mathtt{T}$. Note that the $\notlogicaltransitionfn$ rules~(\Cref{def:asp_rm}) are evaluated using the trace encoding~(\Cref{def:asp_trace_representation}).

\subsection{Learning from Answer Sets Task}
\label{subsec:learning_task}

We learn RMs using ILASP~\cite{law2015ilasp}, a state-of-the-art LAS system. A LAS task consists of background knowledge $\ilaspbackground$, a hypothesis space $\modebias$, and a set of WCDPI examples $\exampleset$~(\Cref{sec:background_ilasp}). To apply LAS to noisy traces, we sample grounded traces from each noisy trace, creating training examples that represent the underlying trace while accounting for uncertainty. This allows us to extend the learning approach of \citet{FurelosBlancoLJBR21} to the noisy setting.

\subsubsection{Background Knowledge}
The background knowledge $\ilaspbackground = R_\mathtt{st} \cup R_\delta \cup R_\rmstates$ consists of the traversal rules from the previous subsection ($R_\mathtt{st}$ and $R_\delta$, \Cref{subsec:rm_encoding}), which determine whether an RM accepts or rejects a trace, together with rules defining the state space:
\begin{equation*}
	\begin{aligned}
		R_\rmstates &= \{\mathtt{state}(\rmstate). \mid \rmstate \in \maxrmstates\}
		\cup \{\mathtt{state}(\acceptingrmstate).\}
		\cup \{\mathtt{state}(\rejectingrmstate).\} \\
		&\quad \cup
		\begin{Bmatrix*}[l]
			\mathtt{used\_state(\rmstates)} \codeif \mathtt{state(\rmstates), ed(\rmstates, \_, \_).} \\
			\mathtt{used\_state(\rmstates)} \codeif \mathtt{state(\rmstates), ed(\_, \rmstates, \_).}
		\end{Bmatrix*}.
	\end{aligned}
\end{equation*}
Here $\maxrmstates$ is the maximum set of states available to ILASP; hence, the learned RM may use a subset. When $|\maxrmstates|$ is large, the search space contains many equivalent RMs differing only in state identifiers. To reduce this redundancy, we extend the symmetry-breaking constraints from \citet{FurelosBlancoLJBR21} by requiring that used states have the lowest possible identifiers. The set of used states is determined via $\mathtt{used\_state(\rmstates)}$ atoms indicating that state $\mathtt{\rmstates}$ has incoming or outgoing edges.

\subsubsection{Hypothesis Space}
\label{sec:hypothesis_space}
The hypothesis space $\modebias$ defines the set of all possible RMs that can be learned. Based on the RM encoding from \Cref{subsec:rm_encoding}, we define:
\begin{equation*}
	\modebias=\left\lbrace\begin{array}{@{}l|c@{}}
		\mathtt{ed}(\rmstate,\rmstate',i).                                                  & \rmstate \in \maxrmstates\setminus \{\acceptingrmstate,\rejectingrmstate\}, \\
		\notlogicaltransitionfn(\rmstate,\rmstate',i,\mathtt{T) \codeif prop}(\singleproposition, \mathtt{T), step(T).} & \rmstate' \in \maxrmstates \setminus \{\rmstate\}, \\
		\notlogicaltransitionfn(\rmstate,\rmstate',i,\mathtt{T) \codeif not~prop}(\singleproposition, \mathtt{T), step(T).}  & i \in [1, \kappa], \singleproposition \in \propositions
	\end{array}\right\rbrace,
\end{equation*}
where $\kappa$ is the maximum number of edges between any two states.

\subsubsection{Example Set} 
\label{subsec:example_set}
The example set $E$ consists of WCDPIs generated from grounded traces~(\Cref{sec:example_gen_traces,sec:example_gen_rms}). 
Since trace termination is deterministic~(\Cref{assumption:termination}), we know with certainty whether each trace is a goal, dead-end, or incomplete trace. 
Below we describe the components of the WCDPI $\langle \exid, \expen, \langle \exinc,\exexc\rangle,\exctx \rangle$ resulting from each \emph{unique} sampled grounded trace $\sampletrace$.

\paragraph{Identifier.}
The identifier $\exid$ is unique for each WCDPI.

\paragraph{Penalty.}
The penalty $\expen$ balances class frequency, trace likelihood, and trace source through three components:
\begin{enumerate}
	\item \textbf{Class-based weight.} To address class imbalance, we assign a base weight inversely proportional to the trace class frequency~\cite{huang2016learning}: $\classweight = \frac{1}{n_c}$, where $n_c$ is the number of examples of class $c \in \{\text{goal}, \text{dead-end}, \text{incomplete}\}$.
	
	\item \textbf{Deduplication count.} If $k$ identical grounded traces exist, we retain one WCDPI with penalty multiplied by $k$. 
	This reflects that frequently sampled grounded traces are more likely under the noisy labelling function and should be prioritised. We track counts separately for traces from noisy labels ($k_{\noisytrace}$) and from the existing RM ($k_{\rewardmachine}$).

    \item  \textbf{Trace origin weight.} We introduce a hyperparameter $\traceoriginmultiplier \in \reals^+$ that controls the importance of RM-derived examples. Setting $\traceoriginmultiplier = 0$ discards RM-derived examples entirely, while $\traceoriginmultiplier = 1$ weights both sources equally.
\end{enumerate}
Thus, the final penalty is $\expen = \classweight (k_{\noisytrace} + \traceoriginmultiplier k_{\rewardmachine})$ for a WCDPI of class $c$.

\paragraph{Inclusions and Exclusions.}
The inclusion-exclusion pair $\langle\exinc,\exexc\rangle$ depends on the trace type:
\begin{itemize}
	\item $\langle\{\mathtt{accept}\},\{\mathtt{reject}\} \rangle$ for \emph{goal} traces,
	\item $\langle\{\mathtt{reject}\},\{\mathtt{accept}\} \rangle$ for \emph{dead-end} traces, and
	\item $\langle\{\},\{\mathtt{accept}, \mathtt{reject}\} \rangle$ for \emph{incomplete} traces.
\end{itemize}
These pairs encode that goal traces must be accepted and not rejected, dead-end traces must be rejected and not accepted, and incomplete traces must be neither accepted nor rejected, consistent with the RM validity definition in \Cref{sec:background_rms}.

\paragraph{Context.}
The context $\exctx$ consists of the ASP representation $\asp{\sampletrace}$ of the grounded trace. Additionally, to speed up learning, we include a constraint ensuring the terminal state is reached after the final label:
\begin{align*}
	&\mathtt{\codeif last(T), T' \leq T, st(T', \acceptingrmstate).} \quad \text{(goal traces)} \\
	&\mathtt{\codeif last(T), T' \leq T, st(T', \rejectingrmstate).} \quad \text{(dead-end traces)}
\end{align*}

\subsubsection{Scoring and Complexity Control}
\label{sec:scoring_complexity}
ILASP scores hypotheses by balancing rule length and unsatisfied example penalties (see \Cref{sec:learning_las_with_noise}). We control this trade-off through two mechanisms:
\begin{enumerate}
	\item \textbf{Penalty budget.} Adjusting the total penalty sum $\sum_{e \in E} \expen$ controls example coverage. A higher budget requires the learned RM to satisfy more examples, potentially increasing complexity.
	
	\item \textbf{Length function.} Customising the rule length function penalises complex RMs. We adjust the weight given to hypothesis length relative to the unsatisfied penalty.
\end{enumerate}
This allows us to find RMs that balance simplicity (fewer states, simpler formulas) with example coverage.
To simplify the learning task, we additionally discard any WCDPIs whose penalty falls below a threshold $\penaltythreshold$.

\section{Exploiting Reward Machines from Noisy Traces}
\label{section:rm_exploitation}

In this section, we describe how \thesystem exploits the structure of an RM when labels are noisy. Akin to \citet{li2024reward}, \thesystem learns policies with PPO~\cite{schulman2017proximal} using a neural network that takes as input the current environment state and the RM state belief, and outputs both the action distribution and value function. 

The key challenge is handling uncertainty in the RM state: since labels are noisy, the agent maintains a belief distribution over RM states rather than a single deterministic state. We introduce the RM state belief in \Cref{subsec:rm_state_belief},
extend the reward function to beliefs in \Cref{sec:reward_over_rm_state_beliefs}, and describe probabilistic reward shaping in \Cref{subsec:probabilistic_reward_shaping}.

\subsection{Reward Machine State Belief}
\label{subsec:rm_state_belief}
Due to the noisy labelling function, the current RM state cannot be known with certainty; therefore, we determine the \emph{RM state belief}~\cite{LiCVKTM22,li2024reward}.

To compute the belief, we need the probability of observing a specific label $\observationlabel \in \powerset{\propositions}$ at a transition. Under \Cref{assumption:independence}, let $[\noisylabelingfunction(\statee_t, \action_t, \statee_{t+1})]_\singleproposition$ denote the probability assigned by the noisy labelling function to proposition $\singleproposition$ at transition $\langle \statee_t, \action_t, \statee_{t+1} \rangle$. The label probability is:
\begin{align*}
	\probability{\observationlabel \mid \statee_t, \action_t, \statee_{t+1}}
	= \prod_{\singleproposition \in \propositions} p_\singleproposition,
\end{align*}
where
\begin{equation*}
	p_\singleproposition = 
	\begin{cases}
		[\noisylabelingfunction(\statee_t, \action_t, \statee_{t+1})]_\singleproposition &
		\text{if } \singleproposition \in \observationlabel; \\
		1 - [\noisylabelingfunction(\statee_t, \action_t, \statee_{t+1})]_\singleproposition &
		\text{if } \singleproposition \notin \observationlabel.
	\end{cases}
\end{equation*}

\begin{definition}[RM state belief] 
	\label{def:rm_state_belief}
	The RM state belief $\rmstatebelief_t \in \simplex{\rmstates}$ is a probability distribution over RM states at timestep $t$. The initial belief is:
\begin{align}
\label{eq:rm_state_belief_init}
\rmstatebelief_0(\rmstate)
=
\sum_{\observationlabel_0 \in 2^{\propositions}}
P(\observationlabel_0 \mid \emptyset, \emptyset, \statee_0)
\mathbf{1}\big[
    \rmstatetransitionfun(\initialrmstate, \observationlabel_0)
    =
    \rmstate
\big].
\end{align}
	Belief updates are performed as follows:
	\begin{align}
		\label{eq:rm_state_belief}
		\rmstatebelief_{t+1}(\rmstate) = \sum_{\substack{\rmstate_t \in \rmstates,\\ \observationlabel_t \in 2^{\propositions}}} P(\observationlabel_t \mid \statee_t, \action_t, \statee_{t+1}) \rmstatebelief_t(\rmstate_t) \mathbf{1}[\rmstatetransitionfun(\rmstate_t, \observationlabel_t) = \rmstate],
	\end{align}
	where $\rmstatetransitionfun$ is the RM state-transition function and $\mathbf{1}$ is the indicator function.
\end{definition}

\begin{example}
	\label{ex:rm_belief_computation}
	Consider the RM from \Cref{fig:ow_rm} and the noisy trace $\noisytrace = \langle 
	\set{\owcoffee: 0.0002,\; \owoffice: 0.0001}, 
	\set{\owcoffee: 0.6513,\; \owoffice: 0.0001}
	\rangle$ resulting from the history $\history=\langle\langle4,6\rangle,\allowbreak\textnormal{left},\allowbreak\langle3,6\rangle\allowbreak\rangle$ in \Cref{ex:noisy_labeling_fn}. We compute the belief evolution using \Cref{def:rm_state_belief}, simplifying summations to include only non-zero terms:
	\begin{itemize}
		\item Step $t=0$. The agent starts in environment state $\statee_0 = \langle 4, 6\rangle$. We compute the initial belief using the noisy label $\noisylabelingfunction(\emptyset, \emptyset, \statee_0) = \set{\owcoffee: 0.0002, \owoffice: 0.0001}$:
		\allowdisplaybreaks 
		\begin{align*}
			\rmstatebelief_0(\initialrmstate)
			&= \sum_{\substack{L_0\not\models\owcoffee\land\neg\owoffice\land\neg\owdecor,\\L_0\not\models\owcoffee\land\owoffice\land\neg\owdecor,\\L_0\not\models\owdecor}} 
            P(L_0\mid \emptyset, \emptyset, \statee_0) \mathbf{1}[\rmstatetransitionfun(\initialrmstate, L_0) = \initialrmstate] \\
			&= P(\emptyset\mid \emptyset, \emptyset, \statee_0) + P(\{\owoffice\}\mid \emptyset, \emptyset, \statee_0) \\
			&= \left[(1 - [\noisylabelingfunction(\emptyset, \emptyset, \statee_0)]_\owcoffee)(1 - [\noisylabelingfunction(\emptyset, \emptyset, \statee_0)]_\owoffice)\right] \\
			&\quad + \left[(1 - [\noisylabelingfunction(\emptyset, \emptyset, \statee_0)]_\owcoffee)[\noisylabelingfunction(\emptyset, \emptyset, \statee_0)]_\owoffice\right] \\
			&= [(1 - 0.0002)(1 - 0.0001)] + [(1 - 0.0002)(0.0001)] \\
			&= 0.9998,\\[2ex]
			\rmstatebelief_0(\rmstate_1)
			&= \sum_{L_0\models\owcoffee\land\neg\owoffice\land\neg\owdecor} 
            P(L_0\mid \emptyset, \emptyset, \statee_0) \mathbf{1}[\rmstatetransitionfun(\initialrmstate, L_0) = \rmstate_1] \\
			&= P(\{\owcoffee\}\mid \emptyset, \emptyset, \statee_0) \\
			&= [\noisylabelingfunction(\emptyset, \emptyset, \statee_0)]_\owcoffee(1 - [\noisylabelingfunction(\emptyset, \emptyset, \statee_0)]_\owoffice) \\
			&= 0.0002(1-0.0001) \\
			&\approx 0.0002,\\[2ex]
			\rmstatebelief_0(\acceptingrmstate)
			&=  \sum_{L_0\models\owcoffee\land\owoffice\land\neg\owdecor} 
            P(L_0\mid \emptyset, \emptyset, \statee_0) \mathbf{1}[\rmstatetransitionfun(\initialrmstate, L_0) = \acceptingrmstate] \\
			&= P(\{\owcoffee,\owoffice\}\mid \emptyset, \emptyset, \statee_0) \\
			&= [\noisylabelingfunction(\emptyset, \emptyset, \statee_0)]_\owcoffee [\noisylabelingfunction(\emptyset, \emptyset, \statee_0)]_\owoffice \\
			&= 0.0002 \cdot 0.0001 \\
			&\approx 0.0000,\\[2ex]
			\rmstatebelief_0(\rejectingrmstate)
			&=\sum_{L_0\models\owdecor} 
            P(L_0\mid \emptyset, \emptyset, \statee_0) \mathbf{1}[\rmstatetransitionfun(\initialrmstate, L_0) = \rejectingrmstate] \\
			&= 0.0 \quad \text{(since $[\noisylabelingfunction(\emptyset, \emptyset, \statee_0)]_\owdecor = 0$)}.
		\end{align*}
		\item Step $t=1$. After the transition $\langle \statee_0, \action_0, \statee_1\rangle$, where $\statee_0=\langle 4,6\rangle, \action_0=\textnormal{left}$,  $\statee_1=\langle 3, 6\rangle$, and $\noisylabelingfunction(\statee_0, \action_0, \statee_1) = \set{\owcoffee: 0.6513,\; \owoffice: 0.0001}$:
		\allowdisplaybreaks 
		\begin{align*}
			\rmstatebelief_1(\initialrmstate) 
			&= \sum_{\substack{L_1\not\models\owcoffee\land\neg\owoffice\land\neg\owdecor,\\L_1\not\models\owcoffee\land\owoffice\land\neg\owdecor,\\L_1\not\models\owdecor}} P(L_1\mid \statee_0, \action_0, \statee_1)\rmstatebelief_0(\initialrmstate) \\
			&= \rmstatebelief_0(\initialrmstate) \left[P(\emptyset\mid \statee_0, \action_0, \statee_1) + P(\{\owoffice\}\mid \statee_0, \action_0, \statee_1)\right] \\
			&= \rmstatebelief_0(\initialrmstate) \left[(1 - [\noisylabelingfunction(\statee_0, \action_0, \statee_1)]_\owcoffee)(1 - [\noisylabelingfunction(\statee_0, \action_0, \statee_1)]_\owoffice)\right. \\
			&\quad \left. + (1 - [\noisylabelingfunction(\statee_0, \action_0, \statee_1)]_\owcoffee)[\noisylabelingfunction(\statee_0, \action_0, \statee_1)]_\owoffice\right] \\
			&= 0.9998 \left[(1 - 0.6513)(1 - 0.0001) + (1 - 0.6513)0.0001\right] \\
			&\approx 0.3486,\\[2ex]
			\rmstatebelief_1(\rmstate_1) 
			&=  \sum_{L_1\models\owcoffee\land\neg\owoffice\land\neg\owdecor} P(L_1\mid \statee_0, \action_0, \statee_1)\rmstatebelief_0(\initialrmstate) + \sum_{\substack{L_1\not\models\owoffice\land\neg\owdecor,\\L_1\not\models\owdecor}} P(L_1\mid \statee_0, \action_0, \statee_1)\rmstatebelief_0(\rmstate_1) \\
			&= \rmstatebelief_0(\initialrmstate)P(\{\owcoffee\}\mid \statee_0, \action_0, \statee_1) + \rmstatebelief_0(\rmstate_1)\left[P(\emptyset\mid \statee_0, \action_0, \statee_1) + P(\{\owcoffee\}\mid \statee_0, \action_0, \statee_1)\right]\\
			&= \rmstatebelief_0(\initialrmstate)\left[[\noisylabelingfunction(\statee_0, \action_0, \statee_1)]_\owcoffee(1 - [\noisylabelingfunction(\statee_0, \action_0, \statee_1)]_\owoffice)\right]\\
			&\quad + \rmstatebelief_0(\rmstate_1) \left[ (1-[\noisylabelingfunction(\statee_0, \action_0, \statee_1)]_\owcoffee)(1 - [\noisylabelingfunction(\statee_0, \action_0, \statee_1)]_\owoffice) \right.\\
			&\quad\quad+ \left. [\noisylabelingfunction(\statee_0, \action_0, \statee_1)]_\owcoffee(1 - [\noisylabelingfunction(\statee_0, \action_0, \statee_1)]_\owoffice)  \right]\\
			&=0.9998[0.6513(1-0.0001)] \\
			&\quad+ 0.0002[(1-0.6513)(1-0.0001)+0.6513(1-0.0001)]\\
			&\approx0.6513,\\[2ex]
			\rmstatebelief_1(\acceptingrmstate) 
			&= \sum_{L_1\models\owcoffee\land\owoffice\land\neg\owdecor} P(L_1\mid \statee_0, \action_0, \statee_1)\rmstatebelief_0(\initialrmstate) + \sum_{L_1\models\owoffice\land\neg\owdecor} P(L_1\mid \statee_0, \action_0, \statee_1)\rmstatebelief_0(\rmstate_1) \\
			&= \rmstatebelief_0(\initialrmstate) P(\{\owcoffee,\owoffice\}\mid \statee_0, \action_0, \statee_1) + \rmstatebelief_0(\rmstate_1)[P(\{\owoffice\}\mid \statee_0, \action_0, \statee_1) + P(\{\owcoffee,\owoffice\}\mid \statee_0, \action_0, \statee_1)]\\
			&=  \rmstatebelief_0(\initialrmstate) \left[ [\noisylabelingfunction(\statee_0, \action_0, \statee_1)]_\owcoffee [\noisylabelingfunction(\statee_0, \action_0, \statee_1)]_\owoffice \right]\\
			&\quad +  \rmstatebelief_0(\rmstate_1) \left[ (1-[\noisylabelingfunction(\statee_0, \action_0, \statee_1)]_\owcoffee) [\noisylabelingfunction(\statee_0, \action_0, \statee_1)]_\owoffice \right.\\
			&\quad\quad\left.+ [\noisylabelingfunction(\statee_0, \action_0, \statee_1)]_\owcoffee [\noisylabelingfunction(\statee_0, \action_0, \statee_1)]_\owoffice \right]\\
			&=0.9998[0.6513\cdot 0.0001] + 0.0002[(1-0.6513)\cdot0.0001 + 0.6513\cdot 0.0001]\\
			&\approx 0.0001,\\[2ex]
			\rmstatebelief_1(\rejectingrmstate)
			&=\sum_{L_1\models\owdecor} P(L_1\mid \statee_0, \action_0, \statee_1)\rmstatebelief_0(\initialrmstate) + \sum_{L_1\models\owdecor} P(L_1\mid \statee_0, \action_0, \statee_1)\rmstatebelief_0(\rmstate_1) \\
			&= 0.0 \quad \text{(since $[\noisylabelingfunction(\statee_0, \action_0, \statee_1)]_\owdecor = 0$)}.
		\end{align*}
	\end{itemize}
	In summary, the resulting beliefs are:
	\begin{align*}
		\rmstatebelief_0 &\approx \begin{bmatrix} 0.9998 & 0.0002 & 0.0000 & 0.0000 \end{bmatrix}^\top, \\
		\rmstatebelief_1 &\approx \begin{bmatrix} 0.3486 & 0.6513 & 0.0001 & 0.0000 \end{bmatrix}^\top.
	\end{align*}
	The belief evolution shows the agent's increasing certainty about having observed \owcoffee: starting with near certainty of being in $\initialrmstate$ at $t=0$, the belief shifts primarily to $\rmstate_1$ by $t=1$, reflecting the strong \owcoffee detection upon moving left. However, substantial probability mass remains on $\initialrmstate$, reflecting uncertainty due to the noisy labels.
\end{example}

\subsection{Reward over Reward Machine State Beliefs}
\label{sec:reward_over_rm_state_beliefs}
In the deterministic case, the agent receives reward $\rmrewardtransitionfun(\rmstate_t, \rmstate_{t+1})$ when transitioning between RM states. However, with noisy labels, the agent maintains a belief distribution over RM states rather than knowing the current state with certainty. We extend the reward-transition function to beliefs by computing the expected reward over all possible state transitions weighted by their probabilities.

\begin{definition}[Reward over RM state beliefs]
	\label{def:rmrewardtransitionfun}
	Given a reward machine $\rewardmachine$, a pair of state beliefs $\rmstatebelief_t, \rmstatebelief_{t+1} \in \simplex{\rmstates}$, and a transition $\langle \statee_t, \action_t, \statee_{t+1} \rangle \in \states\times\actions\times\states$, the reward-transition over beliefs $\rmrewardtransitionfunbelief: \simplex{\rmstates} \times \simplex{\rmstates} \rightarrow \mathbb{R}$ is
	\begin{align}
		\label{eq:rm_reward_beliefs}
		\rmrewardtransitionfunbelief(\rmstatebelief_t, \rmstatebelief_{t+1}) = \sum_{\substack{\rmstate_t, \rmstate_{t+1} \in \rmstates,\\ \observationlabel_t \in 2^{\propositions}}} P(\observationlabel_t \mid \statee_t, \action_t, \statee_{t+1}) \rmstatebelief_t(\rmstate_t) \mathbf{1}[\rmstatetransitionfun(\rmstate_t, \observationlabel_t) = \rmstate_{t+1}] \rmrewardtransitionfun(\rmstate_t, \rmstate_{t+1}).
	\end{align}
\end{definition}

While \Cref{eq:rm_reward_beliefs} holds for any reward-transition function, under \Cref{assumption:reward}, it simplifies to a closed form that depends only on changes in belief at the terminal states:

\begin{proposition}
	\label{proposition:rmrewardtransitionfun}
	Under \Cref{assumption:reward}, the reward-transition over beliefs is:
	\begin{align}
		\label{eq:rm_reward_beliefs_simplified}
		\rmrewardtransitionfunbelief(\rmstatebelief_t, \rmstatebelief_{t+1}) 
		= (\rmstatebelief_{t+1}(\acceptingrmstate) - \rmstatebelief_t(\acceptingrmstate)) - (\rmstatebelief_{t+1}(\rejectingrmstate) - \rmstatebelief_t(\rejectingrmstate)).
	\end{align}
\end{proposition}

\begin{proof}
	See \Cref{derivation:simplification_with_ass_1}.
\end{proof}

\subsection{Probabilistic Reward Shaping}
\label{subsec:probabilistic_reward_shaping}
Reward shaping provides intermediate rewards to guide agent learning. \emph{Potential-based reward shaping}~\cite{NgHR99} defines these rewards by evaluating a \emph{potential function} $\Phi:\states\to\mathbb{R}$ at consecutive states and taking their difference. Formally, the shaped reward is
\begin{align*}
	\shapingreward(\statee_\step, \statee_{\step+1}) = \discountfactor\Phi(\statee_{\step+1}) - \Phi(\statee_\step),
\end{align*}
where $\discountfactor$ is the MDP's discount factor. Crucially, this approach preserves the set of optimal policies.

In the context of reward machines, several works~\cite{CamachoCSM17,CamachoIKVM19,FurelosBlancoLJBR21} leverage the RM structure for reward shaping by defining the potential function over RM states, $\Phi: \rmstates \rightarrow \mathbb{R}$, yielding the shaped reward
\begin{align*}
	\shapingreward(\rmstate_\step, \rmstate_{\step+1}) = \discountfactor\Phi(\rmstate_{\step + 1}) - \Phi(\rmstate_\step).
\end{align*}

In our setting with noisy labels, the agent cannot directly observe the true RM state and instead maintains a belief distribution $\rmstatebelief_\step \in \simplex{\rmstates}$ over RM states. We therefore extend the potential function to operate over beliefs: $\Tilde{\Phi} : \simplex{\rmstates} \rightarrow \mathbb{R}$. 
We define $\Tilde{\Phi}$ as the expected potential under the belief distribution:
\[
\Tilde{\Phi}(\rmstatebelief_t) = \sum_{\rmstate \in \rmstates} \rmstatebelief_t(\rmstate)\Phi(\rmstate).
\]
The shaped reward then follows the same potential-based form:
\begin{align*}
	\shapingreward(\rmstatebelief_t, \rmstatebelief_{t+1}) &= \discountfactor \Tilde{\Phi}(\rmstatebelief_{t+1}) - \Tilde{\Phi}(\rmstatebelief_t) \\
	&= \sum_{\rmstate \in \rmstates} \left(\discountfactor \rmstatebelief_{t+1}(\rmstate) - \rmstatebelief_{t}(\rmstate)\right) \Phi(\rmstate).
\end{align*}
This approach is analogous to potential-based shaping over state beliefs in partially observable MDPs~\cite{EckSDK13}.

To reward progress toward the accepting state, we define the potential function following \citet{CamachoCSM17,FurelosBlancoLJBR21}:
\begin{align*}
\Phi(\rmstate) = |\rmstates| - d_{\min}(\rmstate, \acceptingrmstate),
\end{align*}
where $d_{\min}(\rmstate, \acceptingrmstate)$ is the shortest path length from $\rmstate$ to $\acceptingrmstate$. This assigns higher potential to states closer to $\acceptingrmstate$, thereby rewarding transitions that reduce the distance to the goal. If $\acceptingrmstate$ is unreachable from $\rmstate$, we set $d_{\min}(\rmstate, \acceptingrmstate) = \infty$ (in practice, a sufficiently large number).

Following standard practice in reward shaping, we augment the reward over RM state beliefs $\rmrewardtransitionfunbelief$ (see \Cref{sec:reward_over_rm_state_beliefs}) with the shaped reward:
\begin{align*}
\rmrewardtransitionfunRS(\rmstatebelief_{\step}, \rmstatebelief_{\step+1}) = \rmrewardtransitionfunbelief(\rmstatebelief_\step, \rmstatebelief_{\step+1}) +  \shapingreward(\rmstatebelief_\step, \rmstatebelief_{\step+1}).
\end{align*}

\begin{example}
	\allowdisplaybreaks
	Consider the reward machine in \Cref{fig:ow_rm} with consecutive RM state beliefs $\rmstatebelief_\step = [\begin{matrix}1 & 0 & 0 & 0\end{matrix}]^\top$ and $\rmstatebelief_{\step+1} = [\begin{matrix}0 & 0.5 & 0.5 & 0\end{matrix}]^\top$ (ordered as $\rmstate_0, \rmstate_1, \acceptingrmstate, \rejectingrmstate$), representing a transition from certainty in $\rmstate_0$ to equal uncertainty between $\rmstate_1$ and $\acceptingrmstate$. First, we compute the potential function $\Phi$:
	\begin{equation*}
		\begin{split}
			\Phi(\rmstate_0) &= |\rmstates| - d_{\min}(\rmstate_0, \acceptingrmstate) = 4 - 1 = 3,\\
			\Phi(\acceptingrmstate) &= |\rmstates| - d_{\min}(\acceptingrmstate, \acceptingrmstate) = 4 - 0 = 4,
		\end{split}
		\quad\quad
		\begin{split}
			\Phi(\rmstate_1) &= |\rmstates| - d_{\min}(\rmstate_1, \acceptingrmstate) = 4 - 1 = 3,\\
			\Phi(\rejectingrmstate) &= |\rmstates| - d_{\min}(\rejectingrmstate, \acceptingrmstate) = 4 - \infty = -\infty.
		\end{split}
	\end{equation*}
	With discount factor $\discountfactor=0.9$, the shaped reward is: 
	\begin{align*}
		\shapingreward(\rmstatebelief_\step, \rmstatebelief_{\step+1})
		&= \sum_{\rmstate \in \rmstates} (\discountfactor \rmstatebelief_{t+1}(\rmstate) - \rmstatebelief_{t}(\rmstate)) \Phi(\rmstate) \\
		&= (\discountfactor \rmstatebelief_{t+1}(\rmstate_0) - \rmstatebelief_{t}(\rmstate_0)) \Phi(\rmstate_0) + (\discountfactor \rmstatebelief_{t+1}(\rmstate_1) - \rmstatebelief_{t}(\rmstate_1)) \Phi(\rmstate_1) \\
		&\quad + (\discountfactor \rmstatebelief_{t+1}(\acceptingrmstate) - \rmstatebelief_{t}(\acceptingrmstate)) \Phi(\acceptingrmstate) + (\discountfactor \rmstatebelief_{t+1}(\rejectingrmstate) - \rmstatebelief_{t}(\rejectingrmstate)) \Phi(\rejectingrmstate)\\
		&= (0.9 \cdot 0 - 1) \cdot 3 + (0.9 \cdot 0.5 - 0) \cdot 3 \\
		&\quad + (0.9 \cdot 0.5 - 0) \cdot 4 + (0.9 \cdot 0 - 0) \cdot (-\infty) \\
		&= 0.15.
	\end{align*}
	The positive shaped reward reflects the increased probability of being in $\acceptingrmstate$.
\end{example}

\section{Interleaving Exploitation and Learning of Reward Machines from Noisy Traces}
\label{sec:interleaving_learning_and_exploitation}

In the noiseless setting, prior work~\cite{FurelosBlancoLJBR21,HasanbeigJAMK21,ToroIcarteWKVCM19,XuGAMNTW20} triggers RM relearning when the current candidate RM fails to explain an observed trace. For instance, if the agent successfully completes the task but the corresponding trace does not reach the accepting state, this counterexample indicates the RM must be revised. With noisy labels, however, such a counterexample-based approach becomes unreliable: we cannot determine with certainty whether a trace genuinely contradicts the RM. Moreover, attempting to explain every observed noisy trace would cause the learned RM to overfit to sensor errors.

\begin{figure*}
	\centering
	\tikzstyle{none}=[inner sep=0mm]
	\tikzstyle{arrow}=[->,>=stealth]
	\tikzstyle{node}=[fill=white, draw=black, shape=rectangle, minimum width=1.2cm, minimum height=0.6cm, text centered, font=\footnotesize]
	\tikzstyle{decision}=[diamond, fill=white, draw=black, minimum width=1.0cm, minimum height=0.6cm, text centered, font=\footnotesize]
	\tikzstyle{stack}=[fill=white, draw=black, shape=rectangle, minimum width=0.8cm, minimum height=0.4cm, text centered, font=\footnotesize]
	
	\resizebox{0.9\linewidth}{!}{
		\begin{tikzpicture}
			% Environment (left, bottom row)
			\node [node] (env) at (0, -3) {Environment};
			
			% Agent (center, bottom row)
			\node [node] (agent) at (3.5, -3) {Agent};
			
			% RM (top middle)
			\node [node] (rm) at (3.5, 2) {RM};
			
			% ILASP (top right)
			\node [node] (ilasp) at (7, 2) {ILASP};
			
			% Trace (right, bottom row)
			\node [node] (trace) at (10.5, -3) {$\noisytrace_{t+1}$};
			\node [below=0.05 of trace, font=\footnotesize] {Noisy trace};
			
			% Noisy Traces (cascaded boxes above trace - 3 boxes, stacking to top right)
			\node [fill=white, draw=black, shape=rectangle, minimum width=1.2cm, minimum height=0.6cm, text centered, font=\footnotesize] (noisy_traces_back2) at (10.5+0.1, 0.3+0.1) {};
			\node [fill=white, draw=black, shape=rectangle, minimum width=1.2cm, minimum height=0.6cm, text centered, font=\footnotesize] (noisy_traces_back) at (10.5+0.05, 0.3+0.05) {};
			\node [fill=white, draw=black, shape=rectangle, minimum width=1.2cm, minimum height=0.6cm, text centered, font=\footnotesize] (noisy_traces) at (10.5, 0.3) {Noisy traces};
			
			% Grounded Traces (more cascaded boxes in line with ILASP - 5 boxes, stacking to top right)
			\node [fill=white, draw=black, shape=rectangle, minimum width=1.2cm, minimum height=0.6cm, text centered, font=\footnotesize] (grounded_traces_back4) at (10.5+0.2, 2+0.2) {};
			\node [fill=white, draw=black, shape=rectangle, minimum width=1.2cm, minimum height=0.6cm, text centered, font=\footnotesize] (grounded_traces_back3) at (10.5+0.15, 2+0.15) {};
			\node [fill=white, draw=black, shape=rectangle, minimum width=1.2cm, minimum height=0.6cm, text centered, font=\footnotesize] (grounded_traces_back2) at (10.5+0.1, 2+0.1) {};
			\node [fill=white, draw=black, shape=rectangle, minimum width=1.2cm, minimum height=0.6cm, text centered, font=\footnotesize] (grounded_traces_back1) at (10.5+0.05, 2+0.05) {};
			\node [fill=white, draw=black, shape=rectangle, minimum width=1.2cm, minimum height=0.6cm, text centered, font=\footnotesize] (grounded_traces) at (10.5, 2) {Traces};
			
			% Should Relearn decision hexagon (aligned with noisy traces)
			\node[regular polygon, regular polygon sides=6, minimum width=1cm, xscale=3, draw=black, fill=white, label=center:\footnotesize Should relearn RM?] (decision) at (7, 0.3) {};
			
			% Term function decision hexagon (just above noisy trace)
			\node[regular polygon, regular polygon sides=6, minimum width=1cm, xscale=3, draw=black, fill=white, label={center:\footnotesize Episode terminated?}] (term_decision) at (10.5, -2.0) {};
			
			% Main interaction arrows
			% Env <-> Agent (two curved lines)
			\draw [arrow] (env.east) .. controls +(0.5,0.3) and +(-0.5,0.3) .. node[midway, fill=white, font=\scriptsize] {$s_{t+1}, \termfunction_{t+1}$} (agent.west);
			\draw [arrow] (agent.west) .. controls +(-0.5,-0.3) and +(0.5,-0.3) .. node[midway, fill=white, font=\scriptsize] {$a_t$} (env.east);
			
			% Agent <-> RM (two curved lines)
			\draw [arrow] (agent.north) .. controls +(-0.8,1.5) and +(-0.8,-1.5) .. node[midway, fill=white, font=\scriptsize] {$\noisylabelingfunction_{t+1}$} (rm.south);
			\draw [arrow] (rm.south) .. controls +(0.8,-1.5) and +(0.8,1.5) .. node[midway, fill=white, font=\scriptsize] {$\rmstatebelief_{t+1}, \rmrewardtransitionfunbelief_{t+1}$} (agent.north);
			
			% RM -> ILASP
			\draw [arrow] (ilasp.north) -- +(0,0.3) -| node[pos=0.28, fill=white, font=\scriptsize] {Updated RM} ([xshift=0.17cm]rm.north);
			
			% Grounded Traces -> ILASP (horizontal)
			\draw [arrow] (grounded_traces.west) -- (ilasp.east);
			
			% Noisy Traces -> Decision (straight line from left of noisy traces)
			\draw [arrow] (noisy_traces.west) -- (decision.east);
			
			% Decision -> ILASP (straight)
			\draw [arrow] (decision.north) -- node[midway, fill=white, font=\scriptsize] {Yes} (ilasp.south);
			
			% Noisy Traces -> Grounded Traces (curved)
			\draw [arrow] ([yshift=0.1cm]noisy_traces.north) .. controls +(0,0.5) and +(0,-0.5) .. node[midway, fill=white, font=\scriptsize] {Sample} (grounded_traces.south);
			
			% Agent -> Trace (horizontal)
			\draw [arrow] (agent.east) -- node[midway, fill=white, font=\scriptsize] {$\noisylabelingfunction_{t+1}, \termfunction_{t+1}, \rmstatebelief_{t+1}$} (trace.west);
			
			% Trace -> Term Decision (upward)
			\draw [arrow] (trace.north) -- (term_decision.south);
			
			% Term Decision -> split point (merged line with "Yes")
			\coordinate (split_point) at (10.5, -1.0);
			\draw (term_decision.north) -- node[midway, fill=white, font=\scriptsize] {Yes} (split_point);
			
			% Split point -> Noisy Traces
			\draw [arrow] (split_point) -- (noisy_traces.south);
			
			% Split point -> Should Relearn RM Decision
			\draw [arrow] (split_point) -| node[near end, fill=white, font=\scriptsize] {$\rmstatebelief_{t+1}$} (decision.south);
			
			% RM -> traces
			\draw [arrow] ([xshift=-0.17cm]rm.north) -- +(0.0,0.6) -| node[pos=0.28, fill=white, font=\scriptsize] {Sample} ([yshift=0.2cm]grounded_traces.north);
		\end{tikzpicture}
	}
	\caption{
		Overview of the \thesystem algorithm that interleaves RM learning and exploitation from noisy traces. At each timestep, the agent takes action $\action_t$ and observes the next state $\statee_{t+1}$ and termination information $\termfunction_{t+1}$. The agent computes the noisy label $\noisylabelingfunction_{t+1}$ from the transition $\langle \statee_t,\action_t,\statee_{t+1}\rangle$. The noisy label and current RM are used to determine the RM state belief $\rmstatebelief_{t+1}$ (\Cref{subsec:rm_state_belief}) and RM reward $\rmrewardtransitionfunbelief_{t+1}$ (\Cref{sec:reward_over_rm_state_beliefs}). The episode trace $\noisytrace_{t+1}$ is updated with the noisy label. At episode termination, the trace is stored and we determine whether to relearn the RM using the final RM state belief (\Cref{metodology:relearn_rm}). If relearning is triggered, grounded traces are sampled from the noisy traces (\Cref{sec:example_gen_traces}) and from the current RM (\Cref{sec:example_gen_rms}). These traces are provided to ILASP to learn a new RM.
	}
	\Description{Flow diagram of the Prob-IRM loop. The environment sends the next state and termination information to the agent, and the agent sends an action back. The agent exchanges the noisy label with the RM and receives the RM state belief and RM reward. The noisy label, termination information and belief are appended to the noisy trace of the episode. A decision node asks whether the episode terminated; if yes, the trace is added to a stack of noisy traces and a second decision node asks whether the RM should be relearned, based on the RM state belief. If yes, ILASP is called with traces sampled from the noisy traces and from the current RM, and the updated RM replaces the current one.}
	\label{fig:interleaving_learning_and_exploitation}
\end{figure*}
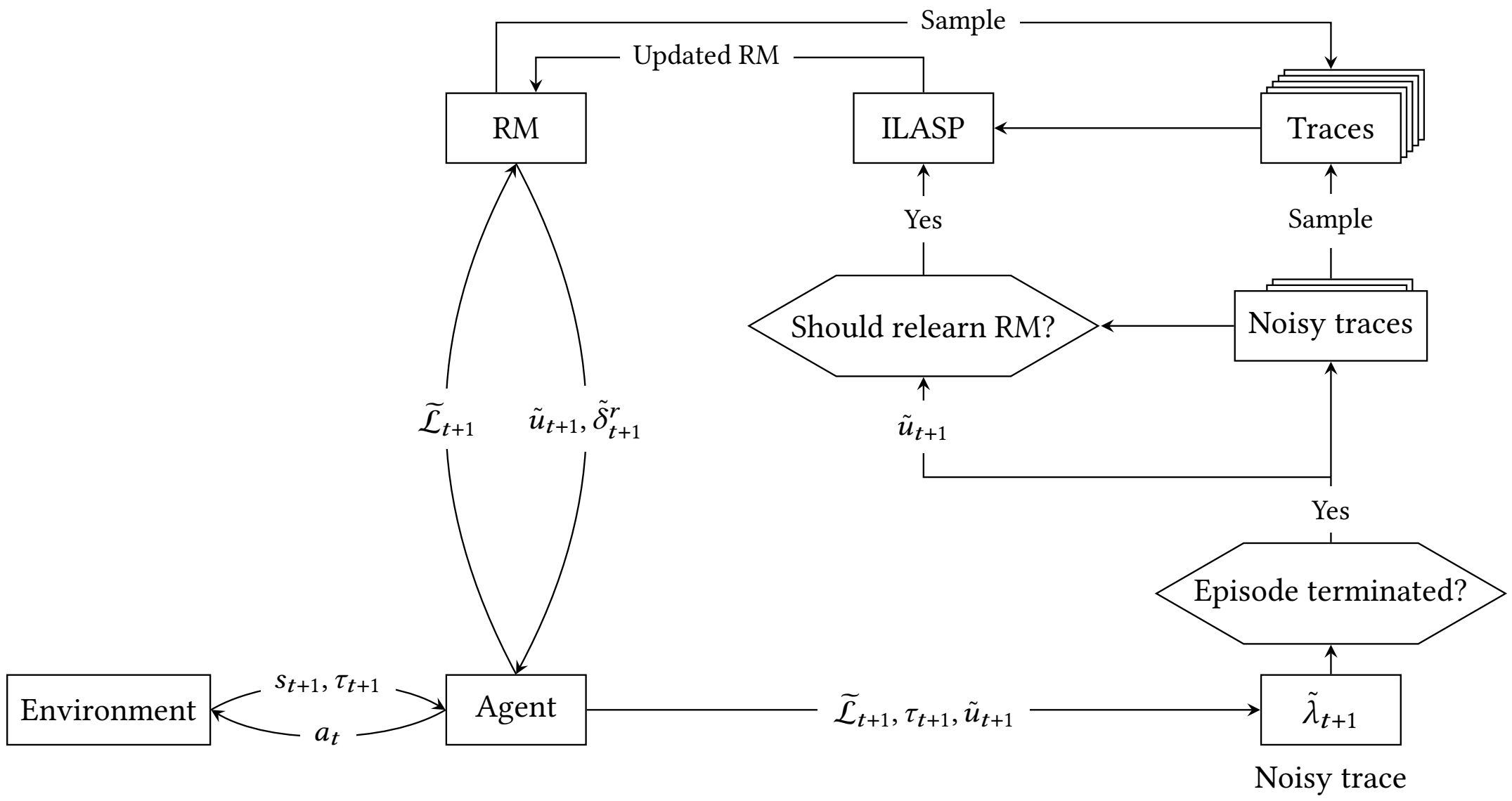

\Cref{fig:interleaving_learning_and_exploitation} provides an overview of \thesystem, our approach for learning RMs from noisy labels. To address the challenges introduced by noisy observations, we propose a criterion for determining when to relearn the RM. We describe this criterion in \Cref{metodology:relearn_rm} and present the complete algorithm in \Cref{subsec:interleaved_algorithm}.

\subsection{Reward Machine Relearning Criterion}
\label{metodology:relearn_rm}
In the noiseless setting, prior work triggers relearning when the RM is not valid with respect to a trace; namely, given a noiseless trace $\trace$ and the traversal $\rewardmachine(\trace)=\langle v_0,\ldots,v_{n+1}\rangle$, a new RM is learned if (i)~$\trace$ is a goal trace and $v_{n+1}\neq\acceptingrmstate$; (ii)~$\trace$ is a dead-end trace and $v_{n+1}\neq\rejectingrmstate$; or (iii)~$\trace$ is an incomplete trace and $v_{n+1}\in\{\acceptingrmstate,\rejectingrmstate\}$. However, with noisy labels, we cannot accurately determine the last RM state of a traversal and thus cannot directly compare it to the trace category to detect errors.

We propose a criterion for RM relearning that builds upon RM state beliefs (\Cref{subsec:rm_state_belief}) and deterministic termination information~(\Cref{assumption:termination}). The criterion is defined using an error metric that measures how well the RM state belief at termination aligns with the ground-truth termination information using cross-entropy.

\begin{definition}[RM error] 
\label{def:rm_error}
Given a noisy trace $\noisytrace=\langle \observationlabel_0,\ldots,\observationlabel_n\rangle$, the resulting RM state belief $\rmstatebelief_n\in\simplex{\rmstates}$, and termination information $\termfunction_{n}=\langle\stateterm_{n},\stategoal_{n} \rangle$ at the final step $n$, the RM error is
\begin{align*}
	\text{error}(v,\tilde{v}) = \textsc{CrossEntropy}(v,\tilde{v}) = -v^\top\log(\tilde{v}),
\end{align*}
where $v\in\{0,1\}^3$ is the ground-truth one-hot vector encoding the trace category (goal, dead-end, and incomplete) based on termination information, and $\tilde{v} \in [0,1]^3$ encodes the RM state belief over terminal ($\acceptingrmstate$ and $\rejectingrmstate$) and non-terminal states. Formally,
\begin{align*}
	v = \begin{bmatrix}
		\mathbf{1}[\stateterm_{n} = \top \land \stategoal_{n} = \top ] \\
		\mathbf{1}[\stateterm_{n} = \top \land \stategoal_{n} = \bot ] \\
		\mathbf{1}[\stateterm_{n} = \bot ]
	\end{bmatrix}, \quad
	\tilde{v} = \begin{bmatrix}
		\rmstatebelief_n(\acceptingrmstate) \\
		\rmstatebelief_n(\rejectingrmstate)  \\
		1- \rmstatebelief_n(\acceptingrmstate) - \rmstatebelief_n(\rejectingrmstate) 
	\end{bmatrix}.
\end{align*}
\end{definition}

\begin{example}
Consider the RM state belief $\rmstatebelief_1 = \begin{bmatrix}0.3486 & 0.6513 & 0.0001& 0.0000\end{bmatrix}^\top$ from \Cref{ex:rm_belief_computation} and the termination information $\termfunction_1=\langle\false,\false\rangle$ (incomplete trace). The vectors are:
\begin{align*}
	v = \begin{bmatrix}
		0 &
		0 &
		1
	\end{bmatrix}^\top, \quad
	\tilde{v} = \begin{bmatrix}
		0.0001 &
		0.0000  &
		0.9999
	\end{bmatrix}^\top.
\end{align*}
The RM error is $-v^\top\log(\tilde{v})=-\log(0.9999)\approx0.0$, indicating that the RM state belief accurately estimates that a terminal state has not been reached.
\end{example}

We trigger RM relearning based on the cumulative RM error since the last update. After each episode termination, we compute the RM error for that episode and add it to a running sum. When the average error across episodes (total error divided by the number of episodes since the last relearning) exceeds a threshold $\rmrelearningthreshold$, we trigger relearning and reset the cumulative error to zero.

To reduce sensitivity to the choice of $\rmrelearningthreshold$, we introduce two techniques:
\begin{enumerate}
	\item \textbf{Warm-up period.} After each relearning, we require a minimum number of episodes before the RM can be updated again, preventing premature relearning.
	\item \textbf{Score-based gating.} We only update the RM if the newly learned hypothesis achieves a better ILASP score (\Cref{def:ilasp_scoring}) than the current RM, preventing updates that do not improve example fit.
\end{enumerate}

Together, these techniques substantially reduce sensitivity to $\rmrelearningthreshold$, as demonstrated in \Cref{exp:cross_entropy_threshold}. The warm-up period ensures training progress even with a conservatively low threshold, while score-based gating prevents unnecessary updates when an error spike is due to temporary exploration rather than RM inadequacy.

\subsection{Interleaved Learning Algorithm}
\label{subsec:interleaved_algorithm}
We now describe \thesystem, our method for interleaving RM learning from noisy traces with RL. \Cref{alg:interleaving_pseudocode} presents the main interleaving loop, and \Cref{alg:rm_learner_pseudocode} describes the \rmlearnercls{} class that manages RM updates.

\begin{algorithm}[ht]
    \caption{\thesystem algorithm}
    \label{alg:interleaving_pseudocode}
    \begin{algorithmic}[1]
		\Require $\propositions$ (propositions), $\maxrmstates$ (RM states), $\kappa$ (max. edges between states), $\rmrelearningthreshold$ (relearning threshold), 
		\Statex \hspace{2.3em} $\mathtt{warmup\_episodes}$ (episodes before relearning),  $\mathtt{max\_ep\_len}$ (max. episode length),
		\Statex \hspace{2.3em}  $\mathtt{num\_iters}$ (iterations), $\mathtt{num\_episodes\_per\_iter}$ (episodes per iteration)

    	\State $\rewardmachine \leftarrow$ \textsc{InitRM}($\maxrmstates$)  \label{algline:init_rm}
    	\State $\pi \leftarrow \textsc{InitPolicy}(\rewardmachine)$
    	\State $\rmlearner \leftarrow \rmlearnercls()$  \label{algline:init_rm_learner}
    	\For{$i \leftarrow 1,\ldots, \mathtt{num\_iters}$}
    	\State $\mathtt{trajectories} \leftarrow \lbrace \rbrace$ \label{algline:trajectories_init}
    	\For{$ep \leftarrow 1,\ldots, \mathtt{num\_episodes\_per\_iter}$}
    	\State $\mathtt{trajectories} \leftarrow \append{\mathtt{trajectories}} {\textsc{SampleTrajectory}(\rewardmachine,\pi, \rmlearner, \mathtt{max\_ep\_len})}$ \label{algline:trajectories_update}
    	\EndFor
    	\State \textsc{UpdatePolicy}($\pi,\mathtt{trajectories}$) \Comment{See \Cref{section:rm_exploitation}} \label{algline:update_policy}
    	\If{$\rmlearner$.\textsc{ShouldRelearn}($\rmrelearningthreshold,\mathtt{warmup\_episodes}$)} \label{algline:relearning_check}
    	\State $\rewardmachine \leftarrow \rmlearner.\textsc{RelearnRM}(\propositions, \maxrmstates, \kappa)$
    	\State $\pi \leftarrow \textsc{InitPolicy}(\rewardmachine)$  \label{algline:reinit_policy}
    	\EndIf
    	\EndFor
    	
		\Statex  % Unnumbered empty line
    	
        \Function{SampleTrajectory}{$\rewardmachine,\pi,\rmlearner,\mathtt{max\_ep\_len}$} \label{algline:sample_traj_start}
            \State $\statee \leftarrow \textsc{EnvInitialState}()$
            \State $\rmstatebelief \leftarrow \textsc{InitBelief}(\rewardmachine, \noisylabelingfunction(\emptyset, \emptyset, \statee))$ \Comment{Using \Cref{eq:rm_state_belief_init}}
            \State $t \leftarrow 0$ \Comment{Initialise episode step counter}
            \State $\noisytrace\leftarrow \langle \rangle$ \Comment{Initialise noisy trace}
            \State $\trajectory \leftarrow \langle\rangle$ \Comment{Initialise raw trajectory}
            \State $\mathtt{done} \leftarrow \bot$ \Comment{Initialise episode termination indicator}
            \While{$\mathtt{done} = \bot$}
		      \State $\action \leftarrow \pi(\statee, \rmstatebelief)$ \label{algline:action_select}
		      \State $\statee', \termfunction \leftarrow$ \textsc{EnvStep}($\statee, \action$) \label{algline:env_step}
            \State $\rmstatebelief' \leftarrow$ \textsc{UpdateBelief}($\rewardmachine,\noisylabelingfunction(\statee, \action, \statee'),\rmstatebelief$) \Comment{Using \Cref{eq:rm_state_belief}} \label{algline:belief_update}
		      \State $\noisytrace \leftarrow \noisytrace \oplus \noisylabelingfunction(\statee, \action, \statee')$
                \State $\trajectory \leftarrow \trajectory \oplus \langle \statee, \action, \statee', \rmrewardtransitionfunbelief(\rmstatebelief, \rmstatebelief'), \rmstatebelief'\rangle$ \label{algline:traj_update}
                \State $\statee \leftarrow \statee'; \rmstatebelief \leftarrow \rmstatebelief'; t \leftarrow t + 1$
                \State $\texttt{is\_terminal\_rm\_state} \leftarrow (\rmstatebelief(\acceptingrmstate) + \rmstatebelief(\rejectingrmstate)) \geq 0.5$
            \State $\mathtt{done} \leftarrow \stateterm =\top$ \textbf{or}  \texttt{is\_terminal\_rm\_state} \textbf{or} $t>  \mathtt{max\_ep\_len}$
            \EndWhile
            \State \textsc{$\rmlearner$.UpdateExamples}($\noisytrace,\rewardmachine,\termfunction$) \label{algline:update_examples_call}
            \State \textsc{$\rmlearner$.UpdateRMError}($\termfunction, \rmstatebelief$) \label{algline:update_rm_error_call}
            \State \Return $\trajectory$  \label{algline:sample_traj_end}
        \EndFunction
    \end{algorithmic}
\end{algorithm}

\begin{algorithm}[ht]
    \caption{The \rmlearnercls{} class}
    \label{alg:rm_learner_pseudocode}
    \begin{algorithmic}[1]
         \Function{\rmlearnercls()}{} \label{algline:constructor_start}
            \State $\texttt{self}.\exampleset \leftarrow \lbrace \rbrace$ \Comment{Set of examples}
            \State $\texttt{self}.\mathtt{num\_episodes} \leftarrow 0$\Comment{Episodes since last RM learning}
            \State $\texttt{self}.\mathtt{rm\_error\_sum} \leftarrow 0$ \Comment{RM error sum since last RM learning} \label{algline:constructor_end}
        \EndFunction
        \Function{ShouldRelearn}{$\rmrelearningthreshold,\mathtt{warmup\_episodes}$} \label{algline:should_relearn_start}
        \If{$\mathtt{self.num\_episodes < warmup\_episodes}$}
        \State \Return $\bot$
        \EndIf
        \State \Return $\mathtt{self.rm\_error\_sum} / \mathtt{self.num\_episodes} > \rmrelearningthreshold$ \label{algline:should_relearn_end}
        \EndFunction
         \Function{RelearnRM}{$\propositions$, $\maxrmstates$, $\kappa$} \label{algline:relearn_start}
        \State $\mathtt{self.num\_episodes} \leftarrow 0$
        \State $\mathtt{self.rm\_error\_sum} \leftarrow 0$ 
        \State \Return \textsc{ILASP.LearnRM($\mathtt{self}.\exampleset$, $\propositions$, $\maxrmstates$, $\kappa$)} \Comment{See \Cref{subsec:learning_task}} \label{algline:relearn_end}
        \EndFunction
         \Function{UpdateExamples}{$\noisytrace, \rewardmachine,\termfunction$} \Comment{See \Cref{sec:example_gen}} \label{algline:update_ex_start}
        \State $\exampleset_{\noisytrace} \leftarrow$ \textsc{GenerateExamples}($\noisytrace,\termfunction$)
        \State $\exampleset_\rewardmachine \leftarrow \textsc{GenerateRMExamples}(\rewardmachine)$ 
        \State $\mathtt{self}.\exampleset \leftarrow \mathtt{self}.\exampleset\cup\exampleset_{\noisytrace}\cup\exampleset_\rewardmachine$  \label{algline:update_ex_end}
        \EndFunction
        \Function{UpdateRMError}{$\termfunction, \rmstatebelief$} \Comment{See \Cref{metodology:relearn_rm}} \label{algline:update_error_start}
        	\State $v \leftarrow \begin{bmatrix}
        		\mathbf{1}[\stateterm = \top \land \stategoal_{n} = \top ] &
        		\mathbf{1}[\stateterm = \top \land \stategoal_{n} = \bot ] &
        		\mathbf{1}[\stateterm = \bot ]
        	\end{bmatrix}^\top$
        	\State $\tilde{v} \leftarrow \begin{bmatrix} \rmstatebelief(\acceptingrmstate) & \rmstatebelief(\rejectingrmstate) & 1 -  \rmstatebelief(\acceptingrmstate) -  \rmstatebelief(\rejectingrmstate) \end{bmatrix}^\top$
        	\State $\mathtt{rm\_error}\leftarrow \textsc{CrossEntropy}(v,\tilde{v})$
	        \State $\mathtt{self.num\_episodes} \leftarrow \mathtt{self.num\_episodes} + 1$ 
        	\State $\mathtt{self.rm\_error\_sum} \leftarrow \mathtt{self.rm\_error\_sum} + \mathtt{rm\_error}$ \label{algline:update_error_end}
        \EndFunction
    \end{algorithmic}
\end{algorithm}

\Cref{alg:interleaving_pseudocode} initialises the RM, policy, and RM learner (lines~\ref{algline:init_rm}--\ref{algline:init_rm_learner}). The initial RM simply loops on the initial state and thus does not accept or reject any traces. The algorithm then runs for a fixed number of iterations. In each iteration, multiple episode trajectories are sampled using the \textsc{SampleTrajectory} function (lines~\ref{algline:trajectories_init}--\ref{algline:trajectories_update}) and used to update the policy (line~\ref{algline:update_policy}). If the relearning criterion is met, a new RM is learned and the policy is reinitialised (lines~\ref{algline:relearning_check}--\ref{algline:reinit_policy}). 

Unlike prior work~\cite{XuGAMNTW20,FurelosBlancoLJBR21} that uses iterative deepening to find a minimal RM by incrementally increasing the state count until all examples are covered, we use a fixed maximum state set $\maxrmstates$ for all RM learning iterations. Each learned RM may use only a subset of $\maxrmstates$, but the maximum remains constant. This is necessary because examples carry noise penalties: iterative deepening would terminate prematurely by ignoring noisy examples rather than increasing the state count. This strategy is also used in other RM learning work with non-noisy traces~\cite{ToroIcarteWKVCM19,HasanbeigJAMK21}.

The \textsc{SampleTrajectory} function (lines~\ref{algline:sample_traj_start}--\ref{algline:sample_traj_end}) runs an episode, collecting a trajectory for policy learning and a noisy trace with its RM error for RM learning. During an episode, the agent selects an action based on the current environment state and RM state belief, executes it in the environment (lines~\ref{algline:action_select}--\ref{algline:env_step}), and updates the RM state belief, noisy trace, and trajectory (lines~\ref{algline:belief_update}--\ref{algline:traj_update}). The episode terminates when: (i)~the environment signals termination; (ii)~the RM state belief is concentrated on terminal states; or (iii)~the maximum episode length is reached. At the end of the episode, the agent updates the RM learning examples using the noisy trace and current RM (line~\ref{algline:update_examples_call}) and increments the accumulated RM error used to determine whether to learn a new RM (line~\ref{algline:update_rm_error_call}).

The \rmlearnercls{} class (\Cref{alg:rm_learner_pseudocode}) implements the relearning criterion (\Cref{metodology:relearn_rm}) and ILASP-based RM learning (\Cref{sec:learning_rms_from_noisy_traces}). The constructor (lines~\ref{algline:constructor_start}--\ref{algline:constructor_end}) initialises an empty example set, along with an episode count and RM error accumulator that track state since the last RM update. The \textsc{ShouldRelearn} method (lines~\ref{algline:should_relearn_start}--\ref{algline:should_relearn_end}) implements the relearning criterion by checking whether the average RM error exceeds the threshold. The \textsc{RelearnRM} method (lines~\ref{algline:relearn_start}--\ref{algline:relearn_end}) resets the episode count and RM error accumulator, then calls ILASP to learn a new RM using the current example set. The \textsc{UpdateExamples} method (lines~\ref{algline:update_ex_start}--\ref{algline:update_ex_end}) updates the example set from both noisy traces and the previous RM (\Cref{sec:example_gen}). Finally, \textsc{UpdateRMError} (lines~\ref{algline:update_error_start}--\ref{algline:update_error_end}) computes the RM error (\Cref{metodology:relearn_rm}) and updates the episode count and error accumulator.

\section{Experimental Results}
\label{experiments}

To evaluate \thesystem, we design experiments across discrete and continuous domains under varying noise conditions. Our experiments address the following research questions:
\begin{enumerate}[label=\textbf{RQ\arabic*},itemsep=0.2em, parsep=0pt]%
	\item\label{rq:probirm} Does \thesystem enable agents to solve non-Markovian reward tasks under noisy labels?
	\item\label{rq:recurrent} Is learning an RM preferable to using recurrent policies?
	\item\label{rq:isa} Does noise-aware RM learning outperform noise-unaware alternatives?
	\item\label{rq:perfect_rm} How does \thesystem compare to using handcrafted RMs?
	\item\label{rq:noise_levels} How sensitive is \thesystem to different sensor noise levels?
	\item\label{rq:continuous} Does \thesystem scale to continuous domains?
\end{enumerate}

We structure this section as follows. \Cref{sec:exp_setup} describes our experimental setup, including domains, baselines, sensor configurations, and hyperparameters. \Cref{sec:officeworld_results,sec:waterworld_results} present results addressing the research questions above. \Cref{sec:ablations} analyses key design choices through ablation studies, covering reward shaping, belief updating, relearning criteria, example generation, and class rebalancing. Finally, \Cref{sec:execution_times} examines computational overhead.

Source code is available at \url{https://github.com/rparac/Prob-IRM}, building on the RLlib framework~\cite{liang2018rllib}. The raw, unaggregated data of all experiments (per-run learning curves, learned RMs, and ILASP logs) is available at \url{https://github.com/rparac/Prob-IRM/releases/tag/jair-data-v1}.

\subsection{Experimental Setup}
\label{sec:exp_setup}
We describe our experimental setup, covering policy learning, domains, sensor configurations, hyperparameters, and performance metrics. Experiments run on a CPU cluster with nodes containing Intel Xeon Platinum 8358 processors (32 cores, 2.60\,GHz) and 256\,GB RAM.

\subsubsection{Policy Learning and Baselines}
We use PPO~\cite{schulman2017proximal} to train agent policies. Unlike the tabular QRM algorithm from our prior work~\cite{parac2024learning}, PPO handles continuous state spaces and avoids discretising RM state beliefs. While the choice of RL algorithm is orthogonal to our method, we select PPO for its strong empirical performance and recent use in RM literature~\cite{li2024reward,ardon2024texttt,li2025ground}.

We compare four approaches:
\begin{itemize}
	\item \textbf{\thesystem} (ours): The network receives the environment observation and RM state belief vector (\Cref{subsec:rm_state_belief}) as input. Learns RMs from noisy traces.
	\item \textbf{\isa}~\cite{FurelosBlancoLJBR21}: A noise-unaware RM learning method. 
    The network receives the environment observation and a one-hot vector indicating the current RM state (computed directly from noisy sensor output).
	\item \textbf{Perfect RM}: The network receives an environment observation and an RM state belief vector. Unlike \thesystem and \isa, it uses a perfect handcrafted RM rather than learning one. Serves as an upper bound for \thesystem.
	\item \textbf{\recurrentbaseline}: Uses an LSTM to capture history instead of an RM. Like RM-learning approaches, has no prior task structure knowledge. Receives the noisy labelling function output concatenated with an environment observation.
\end{itemize}
Network architecture details are in \Cref{subsec:hyperparameters}.

\subsubsection{Domains}
We evaluate \thesystem in the \ow~\cite{ToroIcarteKVM18} and \ww~\cite{Karpathy15,ToroIcarteKVM18} domains.

\paragraph{\ow.} This grid-based navigation domain (introduced in \Cref{sec:background}) features deterministic movement in four directions ($\owup, \owdown, \owleft, \owright$) unless blocked by walls. The agent observes its position in the grid. We evaluate three tasks:
\begin{itemize}
	\item \owcoffeetask: collect coffee ($\owcoffee$) and deliver it to the office ($\owoffice$);
	\item \owcoffeemailtask: collect coffee and mail ($\owmail$) in any order, then reach the office ($\owoffice$);
	\item \owvisit: visit cells $\owcella$, $\owcellb$, $\owcellc$, $\owcelld$ in sequence.
\end{itemize}
Task difficulty varies with map layout. We run each experiment on five sets of ten randomly generated maps with randomised initial agent and object locations, reporting average performance.

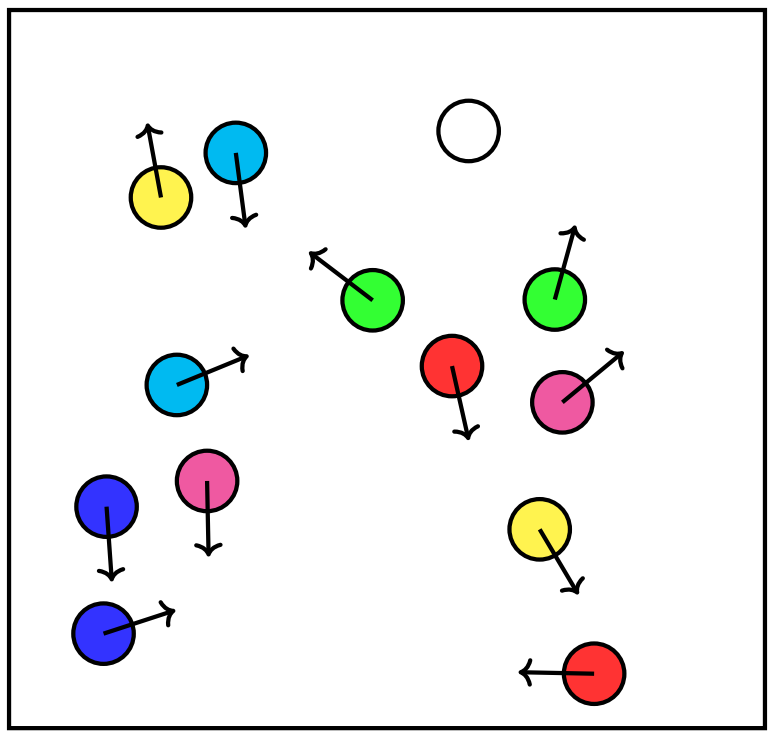
\begin{figure}[t]
	\centering
	\scalebox{0.5}{\newcommand{\ballradius}{0.4cm}

\begin{tikzpicture}
% box
\draw[ultra thick] (0,0) rectangle (10,9.5);
% agent
\filldraw[fill=white!80!white, draw=black, ultra thick] (6.08, 7.90) circle (\ballradius);
% balls
\filldraw[fill=cyan!80!white, draw=black, ultra thick] (3.00, 7.61) circle (\ballradius);
\filldraw[fill=red!80!white, draw=black, ultra thick] (5.86, 4.79) circle (\ballradius);
\filldraw[fill=blue!80!white, draw=black, ultra thick] (1.25, 1.25) circle (\ballradius);
\filldraw[fill=green!80!white, draw=black, ultra thick] (4.81, 5.66) circle (\ballradius);
\filldraw[fill=magenta!80!white, draw=black, ultra thick] (7.32, 4.31) circle (\ballradius);
\filldraw[fill=red!80!white, draw=black, ultra thick] (7.74, 0.72) circle (\ballradius);
\filldraw[fill=cyan!80!white, draw=black, ultra thick] (2.22, 4.54) circle (\ballradius);
\filldraw[fill=green!80!white, draw=black, ultra thick] (7.22, 5.67) circle (\ballradius);
\filldraw[fill=magenta!80!white, draw=black, ultra thick] (2.62, 3.27) circle (\ballradius);
\filldraw[fill=yellow!80!white, draw=black, ultra thick] (2.01, 7.02) circle (\ballradius);
\filldraw[fill=blue!80!white, draw=black, ultra thick] (1.29, 2.93) circle (\ballradius);

% Arrows
\draw [->, ultra thick] (3.00, 7.61) -- (3.13, 6.62);
\draw [->, ultra thick] (5.86, 4.79) -- (6.08, 3.81);
\draw [->, ultra thick] (1.25, 1.25) -- (2.20, 1.56);
\draw [->, ultra thick] (4.81, 5.66) -- (3.97, 6.30);
\draw [->, ultra thick] (7.32, 4.31) -- (8.13, 4.98);
\draw [->, ultra thick] (7.74, 0.72) -- (6.74, 0.74);
\draw [->, ultra thick] (2.22, 4.54) -- (3.17, 4.93);
\draw [->, ultra thick] (7.22, 5.67) -- (7.49, 6.65);
\draw [->, ultra thick] (2.62, 3.27) -- (2.64, 2.27);
\draw [->, ultra thick] (2.01, 7.02) -- (1.83, 8.00);
\draw [->, ultra thick] (1.29, 2.93) -- (1.36, 1.94);

% \filldraw[fill=cyan!80!white, draw=black, ultra thick] (6.13, 1.15) circle (\ballradius);
% \filldraw[fill=red!80!white, draw=black, ultra thick] (4.34, 6.19) circle (\ballradius);
% \filldraw[fill=blue!80!white, draw=black, ultra thick] (4.18, 3.43) circle (\ballradius);
\filldraw[fill=yellow!80!white, draw=black, ultra thick] (7.02, 2.63) circle (\ballradius);
% \filldraw[fill=red!80!white, draw=black, ultra thick] (1.87, 6.02) circle (\ballradius);
% \filldraw[fill=cyan!80!white, draw=black, ultra thick] (7.69, 6.88) circle (\ballradius);
% \filldraw[fill=green!80!white, draw=black, ultra thick] (7.86, 4.57) circle (\ballradius);
% \filldraw[fill=magenta!80!white, draw=black, ultra thick] (2.80, 8.39) circle (\ballradius);
% \filldraw[fill=yellow!80!white, draw=black, ultra thick] (6.01, 6.46) circle (\ballradius);
% \filldraw[fill=blue!80!white, draw=black, ultra thick] (5.50, 8.06) circle (\ballradius);
% % arrows
% \draw [->, ultra thick] (6.13, 1.15) -- (6.61, 0.27);
% \draw [->, ultra thick] (4.34, 6.19) -- (3.34, 6.11);
% \draw [->, ultra thick] (4.18, 3.43) -- (3.24, 3.76);
% \draw [->, ultra thick] (2.78, 3.73) -- (2.35, 2.82);
% \draw [->, ultra thick] (2.09, 1.72) -- (1.13, 2.01);
\draw [->, ultra thick] (7.02, 2.63) -- (7.53, 1.77);
% \draw [->, ultra thick] (1.87, 6.02) -- (0.96, 5.62);
% \draw [->, ultra thick] (7.69, 6.88) -- (8.06, 7.81);
% \draw [->, ultra thick] (7.86, 4.57) -- (7.57, 3.61);
% \draw [->, ultra thick] (2.80, 8.39) -- (3.10, 7.44);
% \draw [->, ultra thick] (6.01, 6.46) -- (6.92, 6.05);
% \draw [->, ultra thick] (5.50, 8.06) -- (4.54, 7.80);
\end{tikzpicture}}
	\caption{Visualisation of the \ww domain~\cite{ToroIcarteKVM18}. The agent is depicted by the white ball.}
	\Description{A square arena containing twelve coloured balls (two each of red, green, blue, magenta, cyan and yellow), each with an arrow indicating its direction of motion, and one white ball representing the agent.}
	\label{fig:water}
\end{figure}

\paragraph{\ww.} This continuous state-space domain (\Cref{fig:water}) features a ball-like agent floating among twelve moving coloured balls (two of each of six colours). All balls travel at constant speed. The agent applies force increments in four directions ($\owup$, $\owdown$, $\owleft$, $\owright$) or no force ($\ownoop$). Observations comprise the agent's absolute position and velocity, plus relative positions and velocities of all other balls. The agent observes a ball's colour upon contact; the proposition set consists of the six colours: red, green, blue, magenta, cyan, and yellow. We run each experiment with five random seeds for initial ball positions and velocities. We evaluate three tasks:
\begin{itemize}
	\item \wwrgbtask: touch red, green, and blue balls in sequence;
	\item \wwrgbctask: touch red, green, blue, then cyan balls in sequence;
	\item \wwrgminusbtask: touch red, green, and blue balls in any order, with green always after red. The blue ball can be observed before, between, or after the red and green balls.
\end{itemize}

\subsubsection{Sensor Configurations}
We experiment with multiple sensor configurations defined by:
\begin{itemize}
	\item \textbf{Noise targets.} Determines which sensors are noisy: \emph{noise-first} applies noise only to the sensor for the first required proposition, while \emph{noise-all} applies noise to all sensors. For \owcoffeemailtask, which has no single first proposition, we apply noise to both \owcoffee and \owmail sensors in the noise-first configuration.
	\item \textbf{Noise level.}  Characterised by sensor accuracy (the probability of correct detection). We test three levels: 1.0 (perfect), 0.9, and 0.8 (20\% error rate). We set sensor sensitivity and specificity to the same value to achieve these posterior probabilities, which we refer to as \emph{sensor confidence}.
\end{itemize}
While noiseless configurations (i.e., with sensor confidence set to 1.0) are solvable by methods assuming perfect labels, our implementation always uses the noisy framework, treating the noiseless setting as a special case. 

\subsubsection{Hyperparameters}
\label{subsec:hyperparameters}
We report hyperparameters in \Cref{tab:hyperparams}. We obtained these values through uniform random search on \owcoffeetask and keep them constant across all experiments. Probabilistic reward shaping is enabled by default for RM-based methods, supported by the ablation study in \Cref{exp:reward_shaping}.

\begin{table}
	\caption{Hyperparameter values. Search ranges show values tested during random search; N/A indicates fixed values.}
	\label{tab:hyperparams}
	\centering
	\renewcommand{\arraystretch}{1.2}
	\small
	\begin{tabular}{lcc}
		\toprule
		\textbf{Hyperparameter} & \textbf{Value} & \textbf{Search Range} \\
		\midrule
        \textbf{Environment} & & \\
        \quad Max. episode length & & \\
        \quad\quad \ow (\owcoffeetask, \owcoffeemailtask) & 500 & N/A\\
        \quad\quad \ow (\owvisit) & 1{,}000 & N/A\\
        \quad\quad \ww (all) & 150 & N/A\\
		\textbf{PPO} & & \\
		\quad Total batch size & 32{,}768 & N/A\\
		\quad Learning rate & 0.000153 & log-uniform [$10^{-8}, 10^{-3}$] \\
		\quad Minibatch size & 8{,}192 & \{1{,}024, 2{,}048, 4{,}096, 8{,}192\}\\
		\quad GAE lambda & 0.95 & uniform [0.9, 1] \\
		\quad PPO clip range & 0.2 & \{0.1, 0.2, 0.3\}\\
		\quad Num. epochs & 13 & \{5, \ldots, 15\}\\
		\quad Value loss coeff. & 0.5 & uniform [0.5, 1] \\
		\quad Entropy loss coeff. & 0.1 & \{0.1, 1\}\\
		\quad Discount ($\gamma$) & 0.99  & N/A\\
		\quad Value clip & 10.0 & N/A\\
		\quad Gradient clip & 0.5 & log-uniform [$10^{-1}, 10^{1}$]\\
		\quad Gradient clipping & Global norm & N/A\\
		\textbf{Network} & & \\
		\quad Num. hidden layers & 4 & \{1, 2, 4\}\\
		\quad Units per hidden layer & 64 & \{8, 16, 32, 64, 128\}\\
		\quad Activation & ReLU & \{ReLU, Tanh, ELU\}\\
		\quad Share actor-critic layers & True & \{True, False\}\\
		\quad Weight initialiser & Orthogonal & N/A\\
		\quad LSTM size (\recurrentbaseline) & 32 & \{8, 16, 32\} \\
		\textbf{\thesystem} & & \\
		\quad Max. number of RM states ($|\maxrmstates|$) & 10 & N/A \\
		\quad Max. edges per state pair ($\maxedgesconst$) & 1 & N/A \\
		\quad Optimiser & Adam & N/A \\
		\quad Cross-entropy threshold ($\rmrelearningthreshold$) & 0.6 & \set{0, 0.5, 0.6, 0.8, 1, 1.5, 2} \\
		\quad Example penalty budget & 3{,}200 & log-uniform [$10^2, 10^4$] \\
		\quad Min. example penalty threshold ($\penaltythreshold$) & 2 & N/A \\
		\quad Num. grounded traces per noisy trace ($\numsamplespertrace$) & 10 & N/A \\
		\quad Trace origin mult. ($\traceoriginmultiplier$) & 0.25 & \{0.25, 0.5, 1, 1.5\} \\
		\quad Reward shaping & True & \{True, False\} \\
		\bottomrule
	\end{tabular}
\end{table}

\subsubsection{Performance Metrics}
To assess agent performance, we record the undiscounted return collected throughout each episode, excluding reward shaping when used. We aggregate returns over all training runs within each experiment to compute mean and standard deviation.

\subsection{\ow Main Results}
\label{sec:officeworld_results}

We evaluate \thesystem and the baselines for the \owcoffeetask, \owcoffeemailtask, and \owvisit tasks under the \emph{noise-first} configuration. 

\subsubsection{Coffee Delivery}
\begin{figure}[ht]
	\centering
	\begin{subfigure}[t]{0.24\linewidth}
		\centering
		\includegraphics[width=\linewidth]{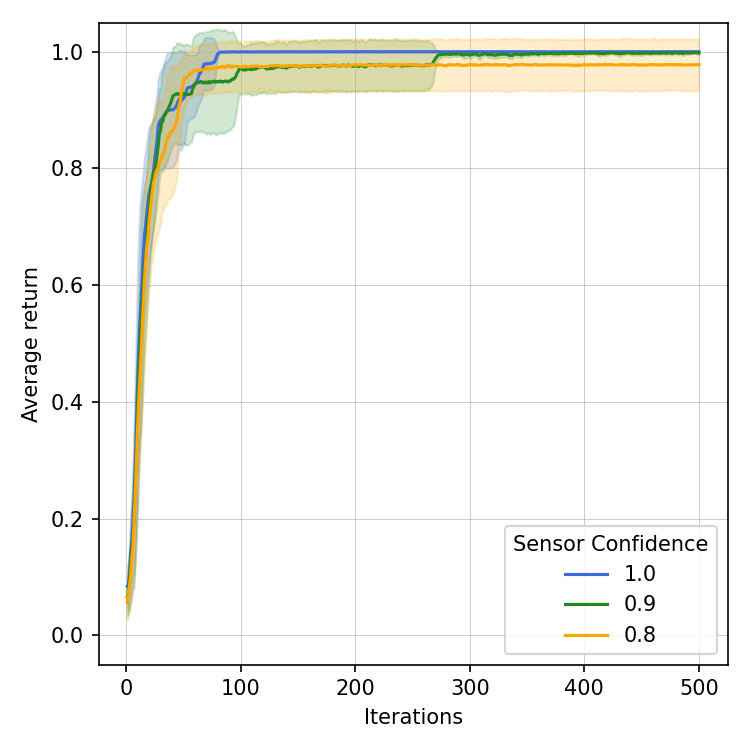}
		\caption{Perfect RM}
		\label{fig:deliver_coffee_baseline}
	\end{subfigure}
	\hfill
	\begin{subfigure}[t]{0.24\linewidth}
		\centering
		\includegraphics[width=\linewidth]{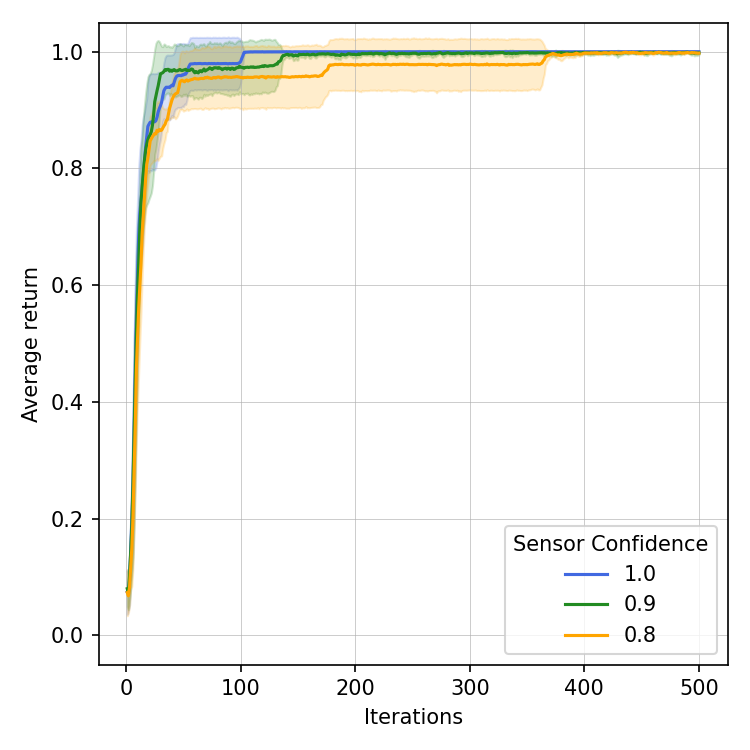}
		\caption{\thesystem}
		\label{fig:deliver_coffee_prob_irm}
	\end{subfigure}
	\hfill
	\begin{subfigure}[t]{0.24\linewidth}
		\centering
		\includegraphics[width=\linewidth]{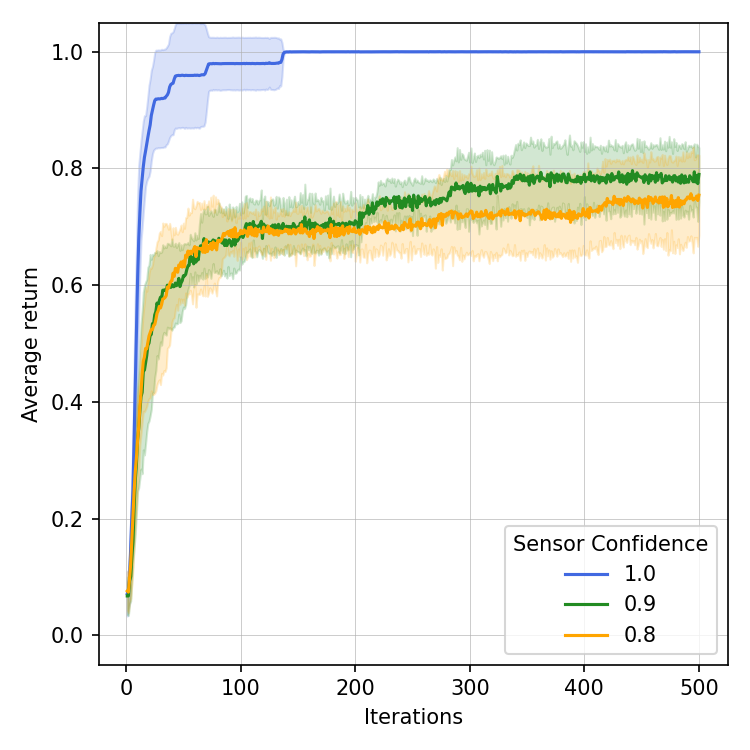}
		\caption{\isa}
		\label{fig:deliver_coffee_isa}
	\end{subfigure}
	\hfill
	\begin{subfigure}[t]{0.24\linewidth}
		\centering
		\includegraphics[width=\linewidth]{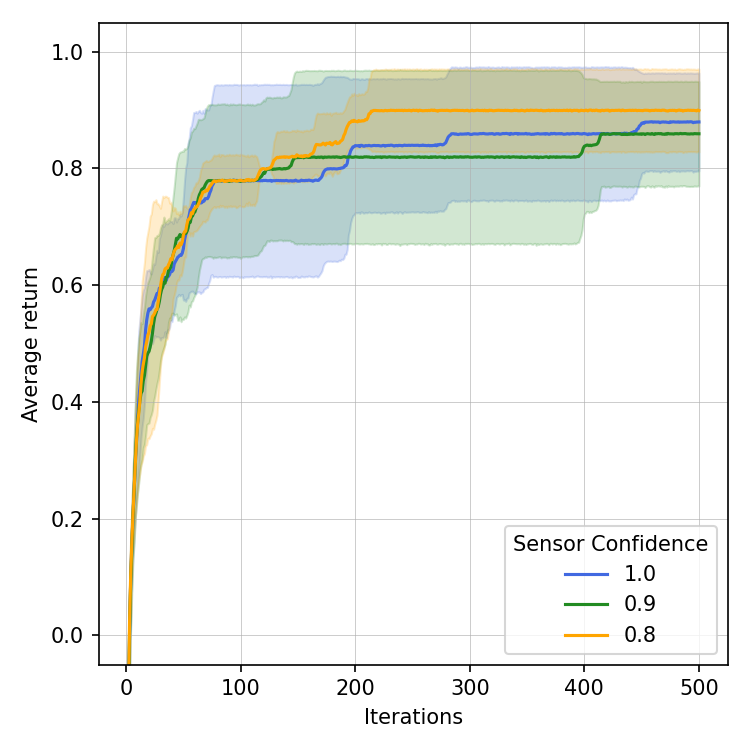}
		\caption{\recurrentbaseline}
		\label{fig:deliver_coffee_recurrent}
	\end{subfigure}
	\caption{Learning curves for \owcoffeetask across different sensor confidence levels.}
	\Description{Four panels of learning curves for the Coffee task, one per method: Perfect RM, Prob-IRM, ISA and Recurrent PPO. Each panel plots average return (vertical axis, 0 to 1) against training iterations (horizontal axis), with one curve per sensor confidence level (1.0, 0.9 and 0.8); lines show the mean and shaded bands the standard deviation over runs. Prob-IRM reaches returns close to the Perfect RM baseline at all three confidence levels, ISA performs well only without noise, and Recurrent PPO does not solve the task consistently.}
	\label{fig:deliver_coffee_comparison}
\end{figure}

\Cref{fig:deliver_coffee_comparison} shows results for \owcoffeetask, the simplest task. \thesystem consistently completes the task, matching the Perfect RM baseline performance (\ref{rq:probirm}, \ref{rq:perfect_rm}). The primary difference is convergence time: \thesystem takes longer as it must first learn an accurate RM (\ref{rq:perfect_rm}). Higher noise levels further increase convergence time, as RM learning becomes more challenging (\ref{rq:noise_levels}).

Both \recurrentbaseline and \isa fail to match this performance. The former never achieves the goal consistently (\ref{rq:recurrent}), while the latter performs well in the noiseless case but degrades substantially under noise (\ref{rq:isa}).

\subsubsection{Coffee and Mail Delivery}
\begin{figure}[ht]
	\centering
	\begin{subfigure}[t]{0.24\linewidth}
		\centering
		\includegraphics[width=\linewidth]{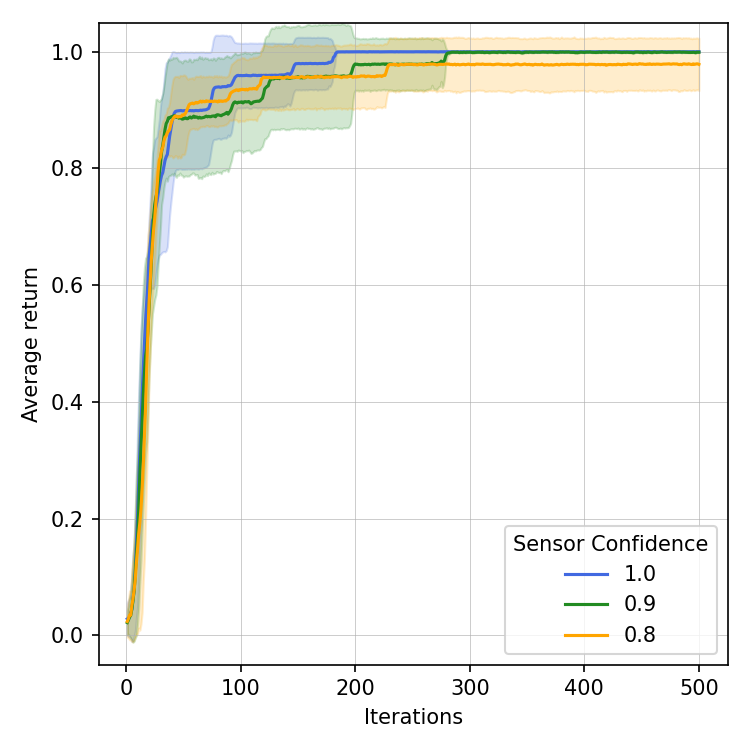}
		\caption{Perfect RM}
		\label{fig:coffee_mail_baseline}
	\end{subfigure}
	\hfill
	\begin{subfigure}[t]{0.24\linewidth}
		\centering
		\includegraphics[width=\linewidth]{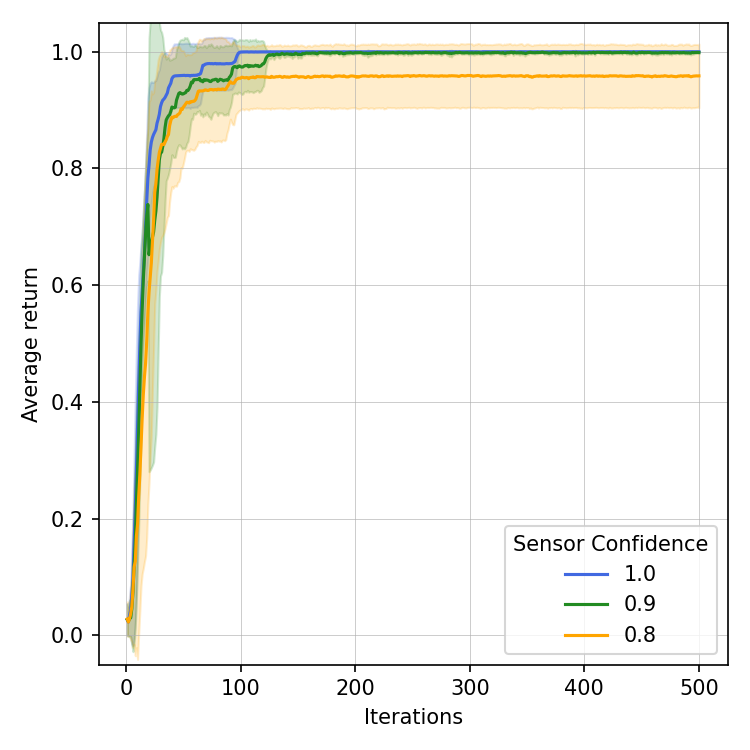}
		\caption{\thesystem}
		\label{fig:coffee_mail_prob_irm}
	\end{subfigure}
	\hfill
	\begin{subfigure}[t]{0.24\linewidth}
		\centering
		\includegraphics[width=\linewidth]{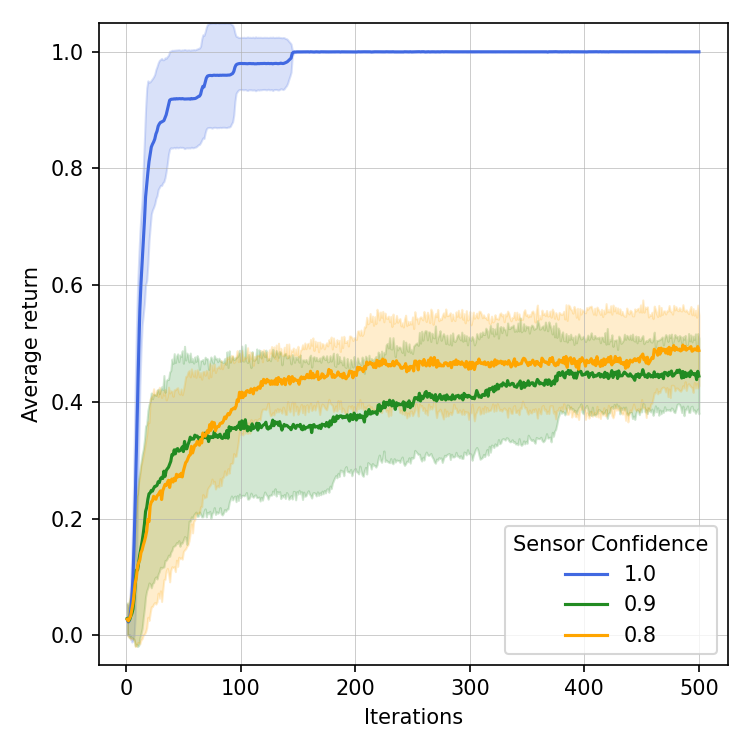}
		\caption{\isa}
		\label{fig:coffee_mail_isa}
	\end{subfigure}
	\hfill
	\begin{subfigure}[t]{0.24\linewidth}
		\centering
		\includegraphics[width=\linewidth]{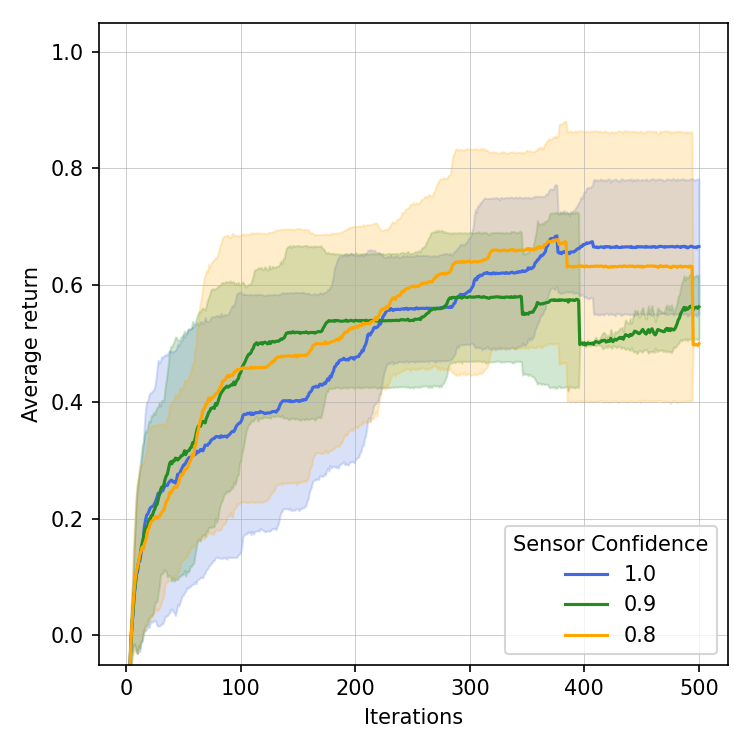}
		\caption{\recurrentbaseline}
		\label{fig:coffee_mail_recurrent}
	\end{subfigure}
	\caption{Learning curves for \owcoffeemailtask across different sensor confidence levels.}
	\Description{Four panels of learning curves for the CoffeeMail task, one per method: Perfect RM, Prob-IRM, ISA and Recurrent PPO. Each panel plots average return (vertical axis, 0 to 1) against training iterations (horizontal axis), with one curve per sensor confidence level (1.0, 0.9 and 0.8); lines show the mean and shaded bands the standard deviation over runs. Prob-IRM is comparable to the Perfect RM baseline across noise levels, ISA performs well only without noise, and Recurrent PPO does not solve the task.}
	\label{fig:coffee_mail_comparison}
\end{figure}

\Cref{fig:coffee_mail_comparison} shows results for \owcoffeemailtask. \thesystem again achieves performance comparable to the Perfect RM baseline across all noise levels (\ref{rq:probirm}, \ref{rq:perfect_rm}). 

As in \owcoffeetask, \recurrentbaseline fails to achieve the task goal (\ref{rq:recurrent}), while \isa follows the same pattern, performing well in the noiseless case but obtaining even lower performance under noise (\ref{rq:isa}).

\subsubsection{Sequential Visitation}
\begin{figure}[ht]
	\centering
	\begin{subfigure}[t]{0.24\linewidth}
		\centering
		\includegraphics[width=\linewidth]{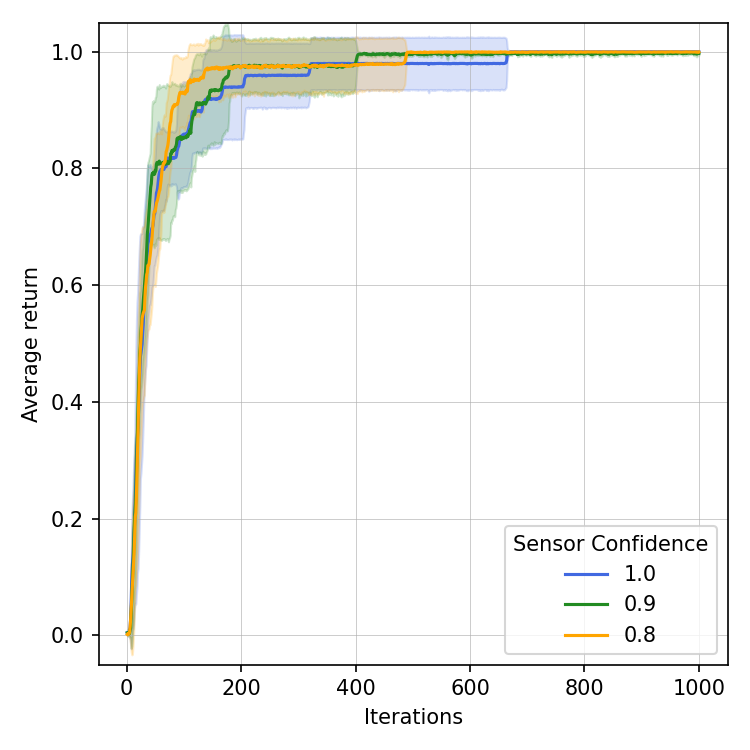}
		\caption{Perfect RM}
		\label{fig:visit_abcd_baseline}
	\end{subfigure}
	\hfill
	\begin{subfigure}[t]{0.24\linewidth}
		\centering
		\includegraphics[width=\linewidth]{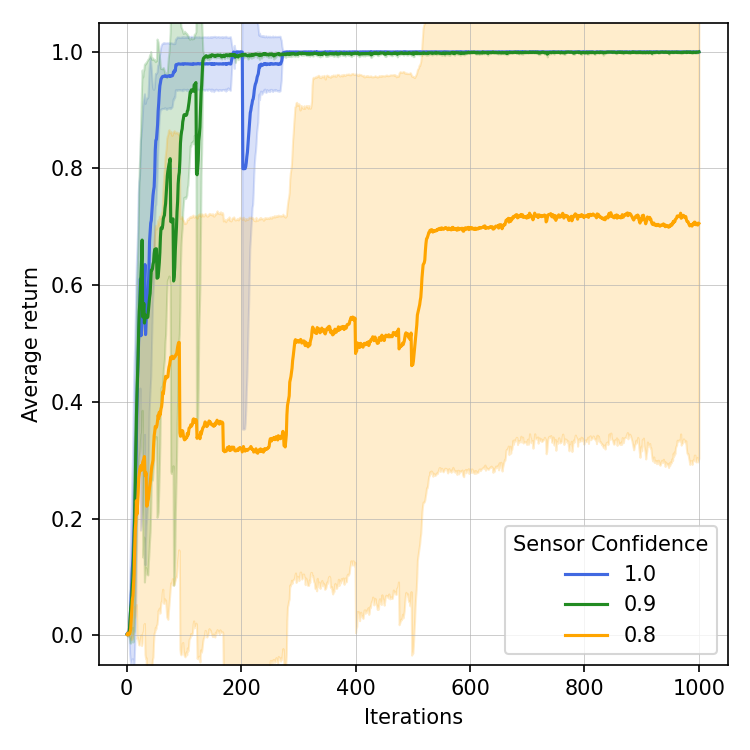}
		\caption{\thesystem}
		\label{fig:visit_abcd_prob_irm}
	\end{subfigure}
	\hfill
	\begin{subfigure}[t]{0.24\linewidth}
		\centering
		\includegraphics[width=\linewidth]{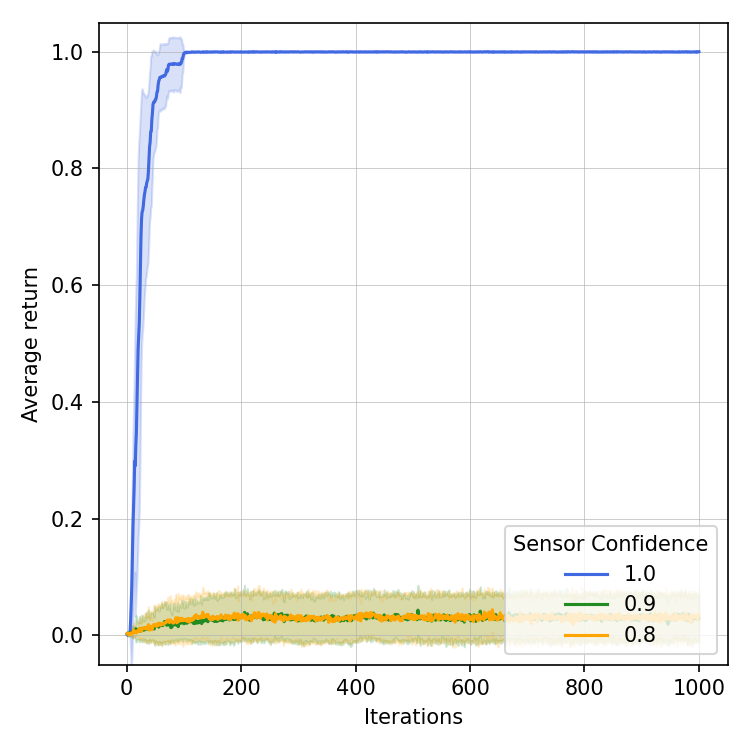}
		\caption{\isa}
		\label{fig:visit_abcd_isa}
	\end{subfigure}
	\hfill
	\begin{subfigure}[t]{0.24\linewidth}
		\centering
		\includegraphics[width=\linewidth]{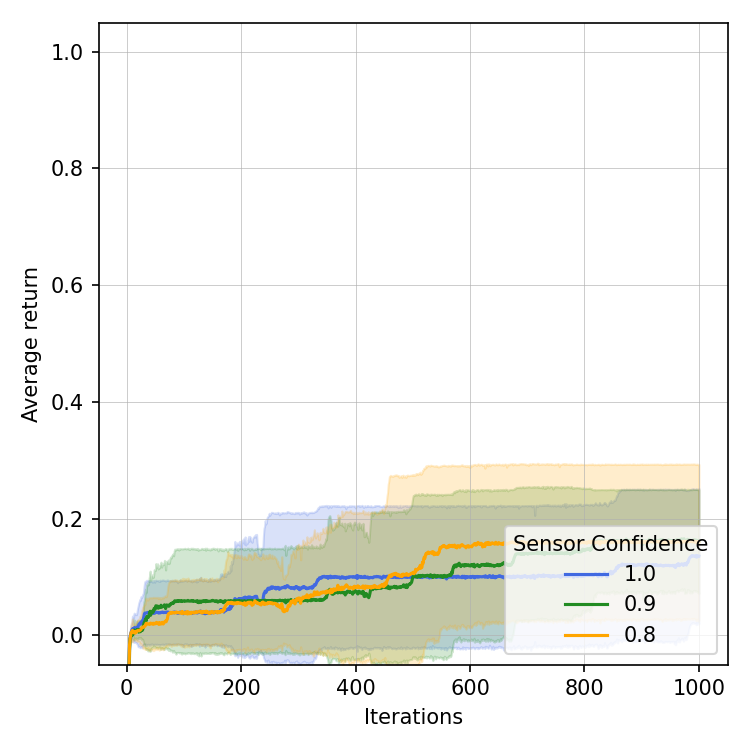}
		\caption{\recurrentbaseline}
		\label{fig:visit_abcd_recurrent}
	\end{subfigure}
	\caption{Learning curves for \owvisit across different sensor confidence levels.}
	\Description{Four panels of learning curves for the VisitABCD task, one per method: Perfect RM, Prob-IRM, ISA and Recurrent PPO. Each panel plots average return (vertical axis, 0 to 1) against training iterations (horizontal axis), with one curve per sensor confidence level (1.0, 0.9 and 0.8); lines show the mean and shaded bands the standard deviation over runs. Prob-IRM solves the task at confidence 1.0 and 0.9 and degrades at 0.8, ISA fails to converge under noise, and Recurrent PPO does not solve the task.}
	\label{fig:visit_abcd_comparison}
\end{figure}

\Cref{fig:visit_abcd_comparison} shows results for \owvisit, the most challenging task. \thesystem consistently achieves the task's goal for sensor confidences of 1.0 and 0.9, but degrades at 0.8 due to one particularly difficult seed (\ref{rq:probirm}, \ref{rq:perfect_rm}, \ref{rq:noise_levels}). 

\recurrentbaseline again fails to solve the task at any noise level (\ref{rq:recurrent}). More notably, \isa fails to converge under noise despite performing well in the noiseless case (\ref{rq:isa}). This degradation is more pronounced than in simpler tasks, highlighting the growing importance of noise-aware RM learning as task complexity increases.

\subsubsection{Summary}
\thesystem consistently solves all three \ow tasks (\ref{rq:probirm}), substantially outperforming \recurrentbaseline and \isa (\ref{rq:recurrent}, \ref{rq:isa}). The former fails to solve the tasks across all noise levels, while the latter, despite performing well in noiseless settings, degrades under noise as task complexity grows.

\thesystem is comparable to Perfect RM at lower noise levels (0.9, 1.0), with performance degrading at higher noise (0.8) on harder tasks (\ref{rq:perfect_rm}, \ref{rq:noise_levels}). The primary trade-off is increased convergence time, as the agent must first learn an accurate RM.

Unlike our previous work~\cite{parac2024learning}, we observe stable RM relearning frequencies across noise levels. We attribute this to collecting more traces between relearning checks, yielding more accurate estimates of the underlying distribution and consequently more stable RM learning.

\subsection{\ww Main Results}
\label{sec:waterworld_results}
We evaluate \thesystem and baselines on the three tasks of the \ww domain under the \emph{noise-first} configuration, addressing \ref{rq:continuous} on scalability to continuous state spaces.

\subsubsection{\wwrgbtask}
\begin{figure}
	\centering
	\begin{subfigure}[t]{0.24\linewidth}
		\centering
		\includegraphics[width=\linewidth]{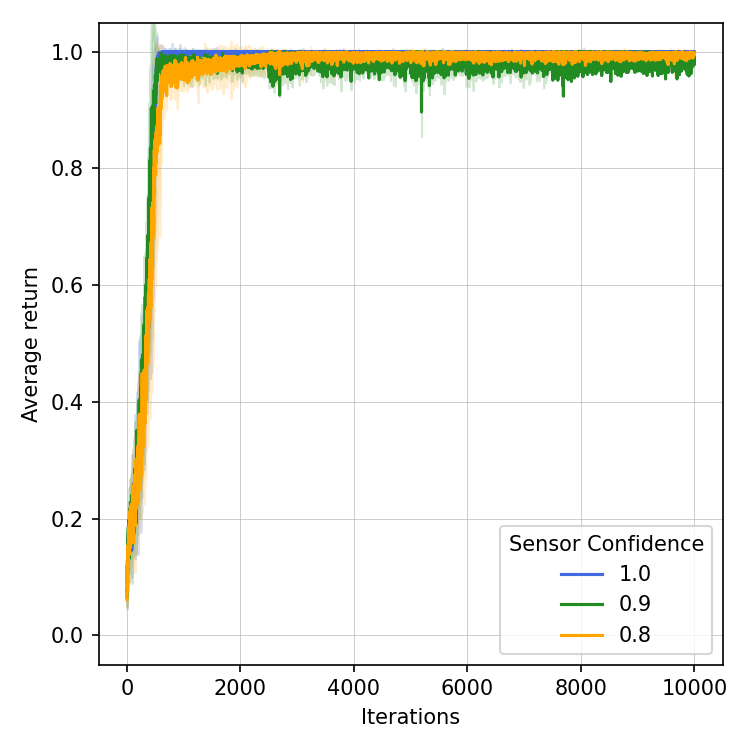}
		\caption{Perfect RM}
		\label{fig:ww_rgb_perfect_rm}
	\end{subfigure}
	\hfill
	\begin{subfigure}[t]{0.24\linewidth}
		\centering
		\includegraphics[width=\linewidth]{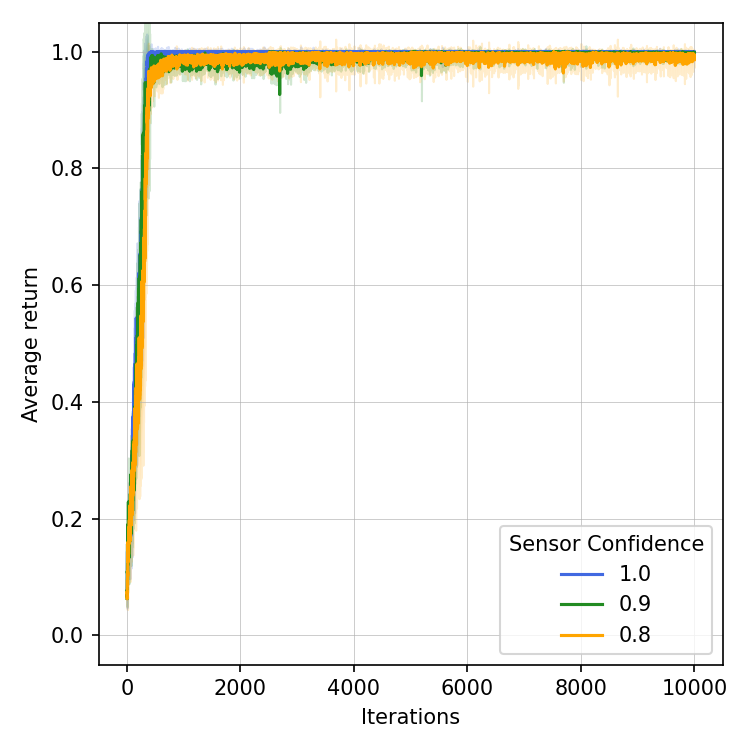}
		\caption{\thesystem}
		\label{fig:ww_rgb_prob_irm}
	\end{subfigure}
	\hfill
	\begin{subfigure}[t]{0.24\linewidth}
		\centering
		\includegraphics[width=\linewidth]{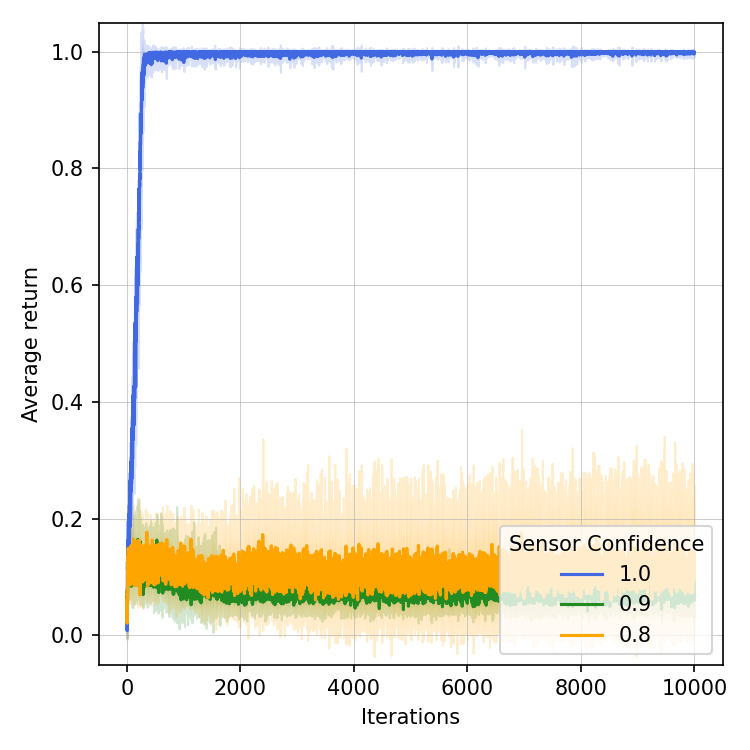}
		\caption{\isa}
		\label{fig:ww_rgb_isa}
	\end{subfigure}
	\hfill
	\begin{subfigure}[t]{0.24\linewidth}
		\centering
		\includegraphics[width=\linewidth]{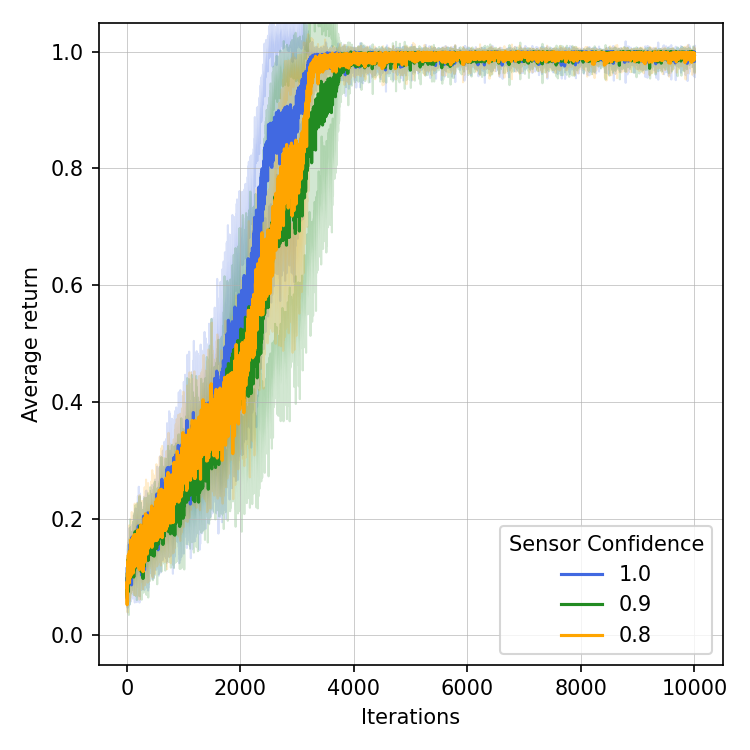}
		\caption{\recurrentbaseline}
		\label{fig:ww_rgb_recurrent}
	\end{subfigure}
	\caption{Learning curves for \wwrgbtask across different sensor confidence levels.}
	\Description{Four panels of learning curves for the RGB task, one per method: Perfect RM, Prob-IRM, ISA and Recurrent PPO. Each panel plots average return (vertical axis, 0 to 1) against training iterations (horizontal axis), with one curve per sensor confidence level (1.0, 0.9 and 0.8); lines show the mean and shaded bands the standard deviation over runs. Prob-IRM converges close to the Perfect RM baseline at all noise levels, Recurrent PPO converges more slowly, and ISA degrades under noise.}
	\label{fig:rgb_comparison}
\end{figure}

\Cref{fig:rgb_comparison} shows results for \wwrgbtask. \thesystem solves the task across all noise levels, converging close to the Perfect RM baseline (\ref{rq:probirm}, \ref{rq:perfect_rm}, \ref{rq:noise_levels}). \recurrentbaseline converges significantly more slowly (\ref{rq:recurrent}), while \isa follows the pattern observed in \ow: good performance in the noiseless case but substantial degradation under noise (\ref{rq:isa}).

\subsubsection{\wwrgminusbtask}
\begin{figure}
	\centering
	\begin{subfigure}[t]{0.24\linewidth}
		\centering
		\includegraphics[width=\linewidth]{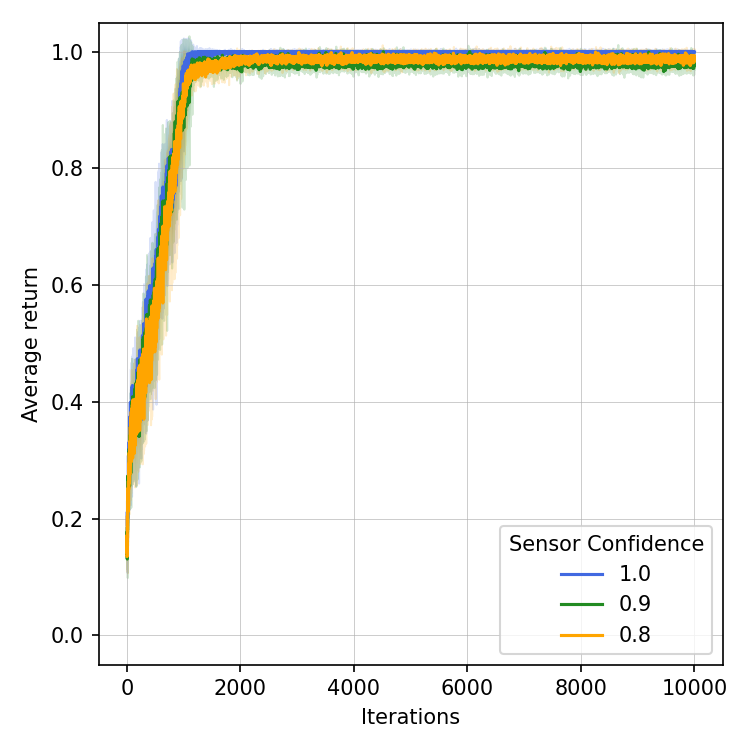}
		\caption{Perfect RM}
		\label{fig:ww_rg_b_perfect_rm}
	\end{subfigure}
	\hfill
	\begin{subfigure}[t]{0.24\linewidth}
		\centering
		\includegraphics[width=\linewidth]{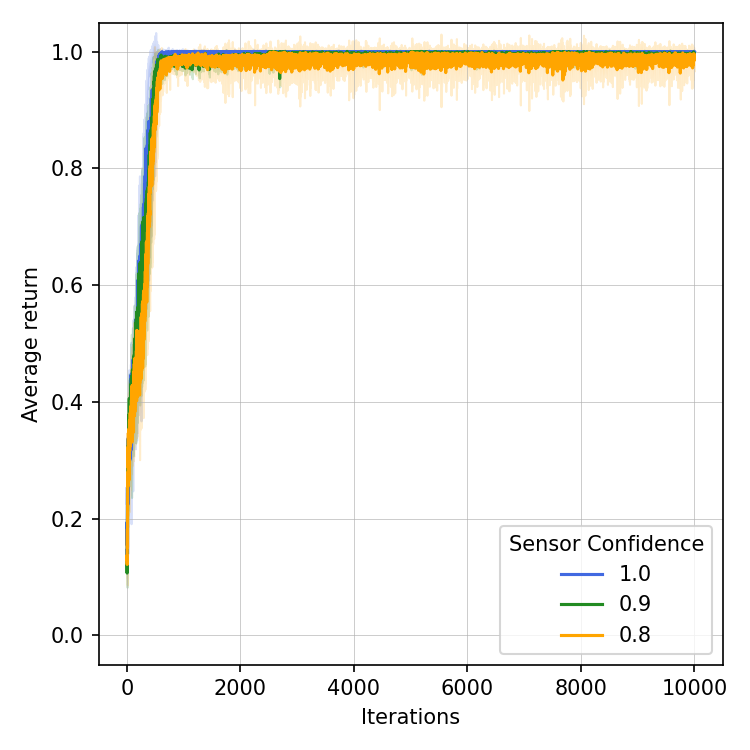}
		\caption{\thesystem}
		\label{fig:ww_rg_b_prob_irm}
	\end{subfigure}
	\hfill
	\begin{subfigure}[t]{0.24\linewidth}
		\centering
		\includegraphics[width=\linewidth]{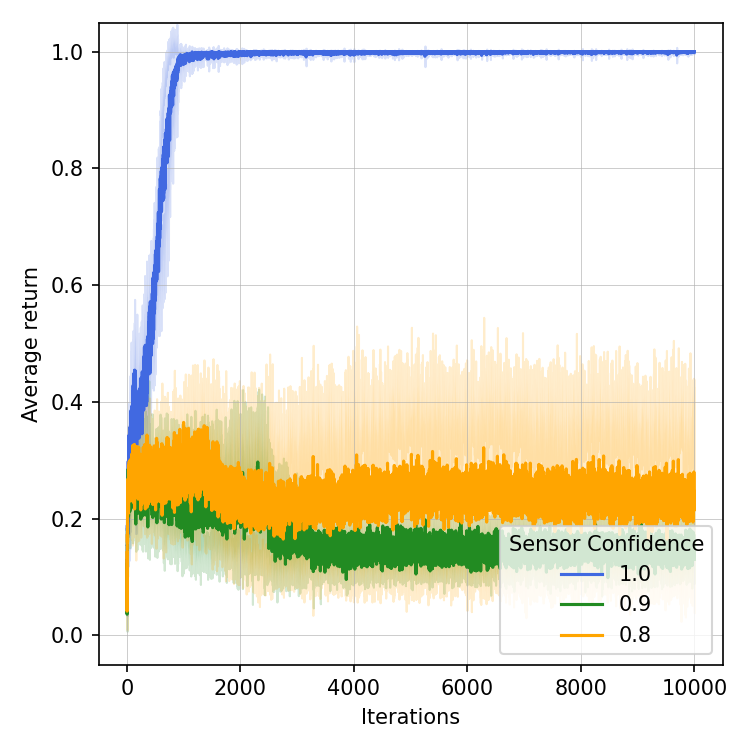}
		\caption{\isa}
		\label{fig:ww_rg_b_isa}
	\end{subfigure}
	\hfill
	\begin{subfigure}[t]{0.24\linewidth}
		\centering
		\includegraphics[width=\linewidth]{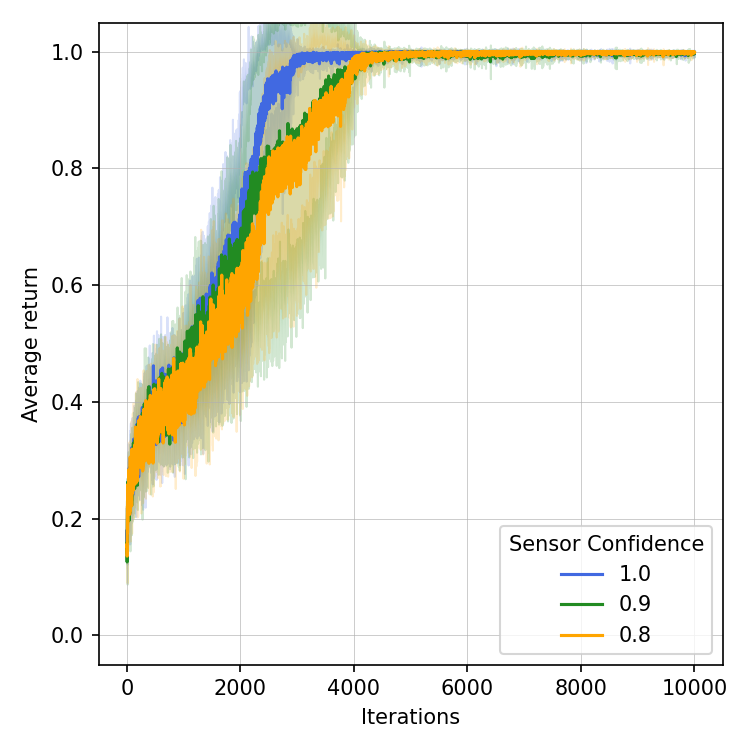}
		\caption{\recurrentbaseline}
		\label{fig:ww_r_gb_recurrent}
	\end{subfigure}
	\caption{Learning curves for \wwrgminusbtask across different sensor confidence levels.}
	\Description{Four panels of learning curves for the RG-B task, one per method: Perfect RM, Prob-IRM, ISA and Recurrent PPO. Each panel plots average return (vertical axis, 0 to 1) against training iterations (horizontal axis), with one curve per sensor confidence level (1.0, 0.9 and 0.8); lines show the mean and shaded bands the standard deviation over runs. Prob-IRM matches the Perfect RM baseline, Recurrent PPO converges more slowly, and ISA degrades under noise.}
	\label{fig:rg_b_comparison}
\end{figure}

\Cref{fig:rg_b_comparison} shows results for \wwrgminusbtask. \thesystem matches the Perfect RM baseline in both final return and convergence speed (\ref{rq:perfect_rm}). \recurrentbaseline converges more slowly but eventually solves the task (\ref{rq:recurrent}), while \isa again degrades under noise (\ref{rq:isa}).

\subsubsection{\wwrgbctask}
\begin{figure}[ht]
    \centering
    \begin{subfigure}{0.24\linewidth}
        \centering
        \includegraphics[width=\linewidth]{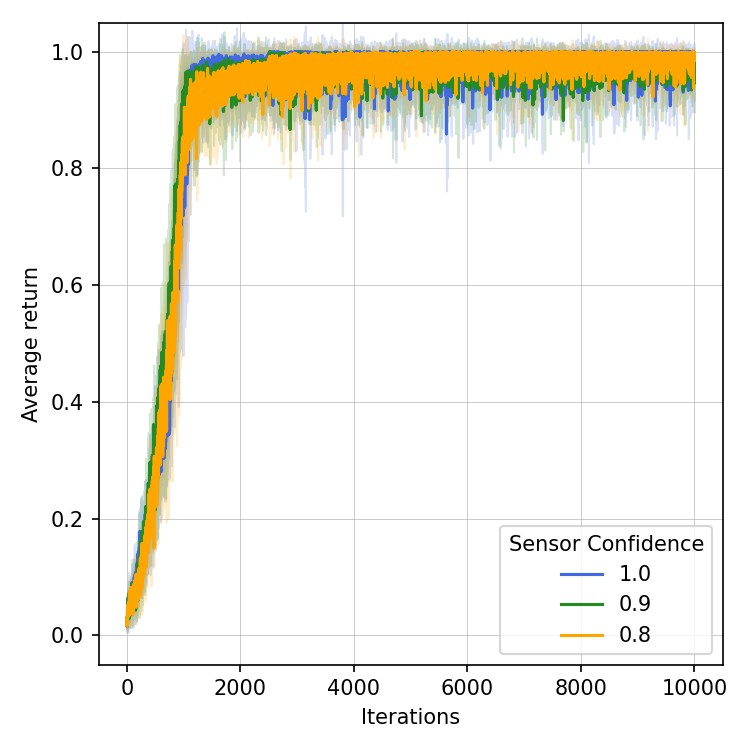}
        \caption{Perfect RM}
        \label{fig:ww_rgbc_perfect_rm}
    \end{subfigure}
    \hfill
    \begin{subfigure}{0.24\linewidth}
        \centering
        \includegraphics[width=\linewidth]{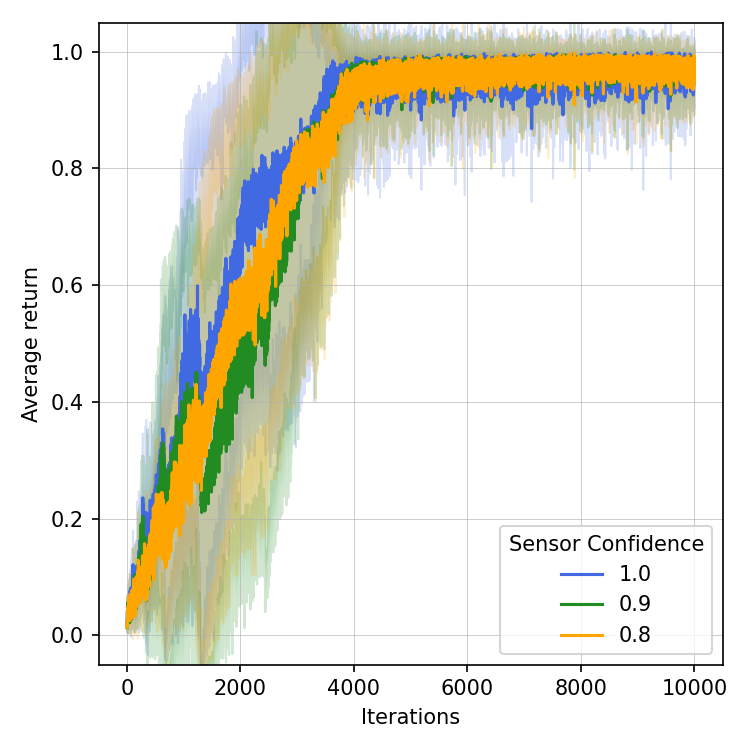}
        \caption{\thesystem}
        \label{fig:ww_rgbc_prob_irm}
    \end{subfigure}
    \hfill
    \begin{subfigure}{0.24\linewidth}
        \centering
        \includegraphics[width=\linewidth]{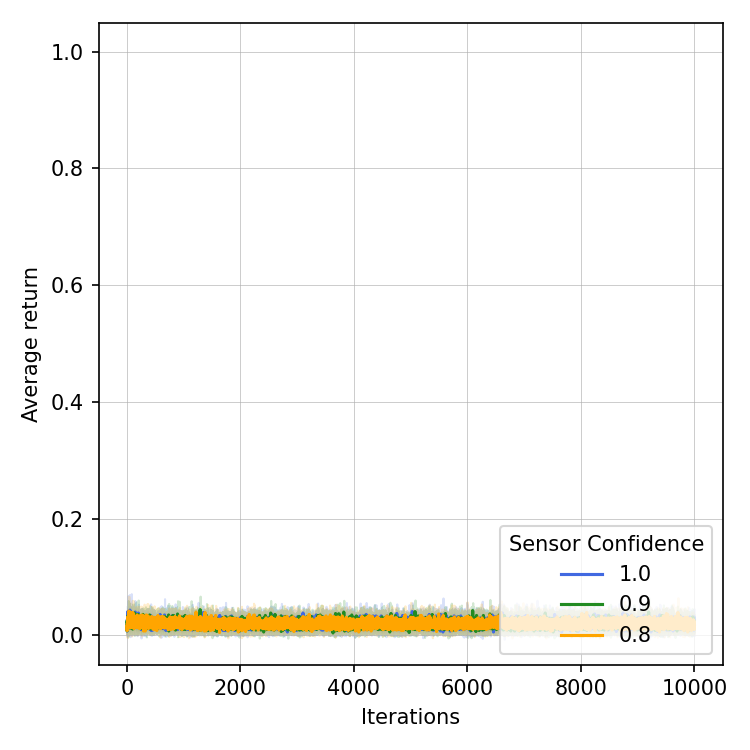}
        \caption{\isa}
        \label{fig:ww_rgbc_isa}
    \end{subfigure}
    \hfill
    \begin{subfigure}{0.24\linewidth}
        \centering
        \includegraphics[width=\linewidth]{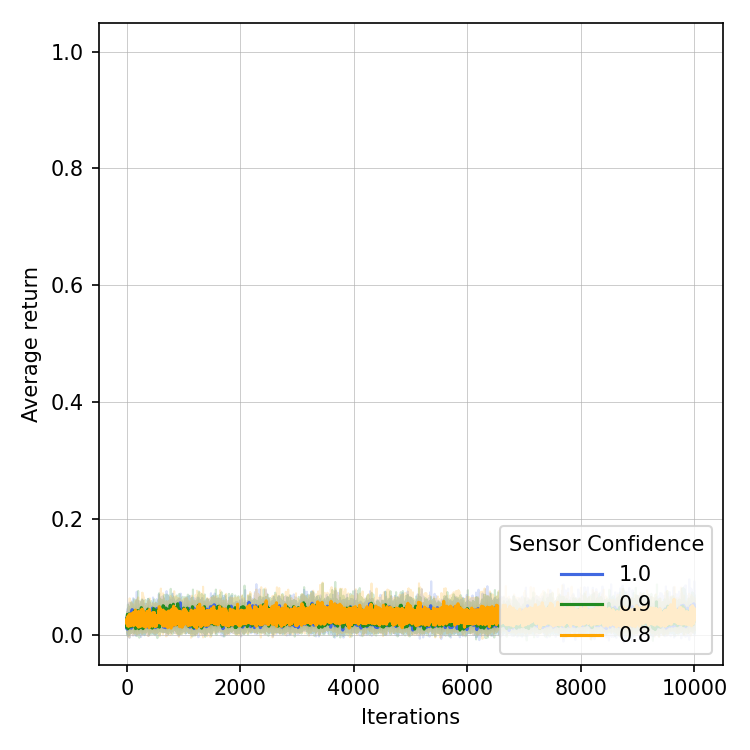}
        \caption{\recurrentbaseline}
        \label{fig:ww_rgbc_recurrent}
    \end{subfigure}
    \caption{Learning curves for \wwrgbctask across different sensor confidence levels.}
    \Description{Four panels of learning curves for the RGBC task, one per method: Perfect RM, Prob-IRM, ISA and Recurrent PPO. Each panel plots average return (vertical axis, 0 to 1) against training iterations (horizontal axis), with one curve per sensor confidence level (1.0, 0.9 and 0.8); lines show the mean and shaded bands the standard deviation over runs. Prob-IRM solves the task at all noise levels, while ISA and Recurrent PPO fail to make progress.}
    \label{fig:rgbc_comparison}
\end{figure}

\Cref{fig:rgbc_comparison} shows results for \wwrgbctask, the most challenging task. \thesystem solves it across all noise levels (\ref{rq:probirm}, \ref{rq:perfect_rm}), while \recurrentbaseline fails to make progress regardless of noise level (\ref{rq:recurrent}). 

Notably, \isa fails even in the noiseless setting (\ref{rq:isa}), contrasting with results in \citet{FurelosBlancoLJBR21}. We attribute this to our distributed training architecture: rather than checking for RM relearning after each trace, we check at iteration end, adding all conflicting examples simultaneously. The resulting large learning task causes \isa to time out.

\subsection{Ablations}
\label{sec:ablations} 
We conduct ablation studies to assess the impact of key design choices in \thesystem.

\subsubsection{Probabilistic Reward Shaping}
\label{exp:reward_shaping}

\begin{figure}[ht]
    \centering
    \begin{subfigure}[t]{0.4\linewidth}
        \centering
        \includegraphics[width=\linewidth]{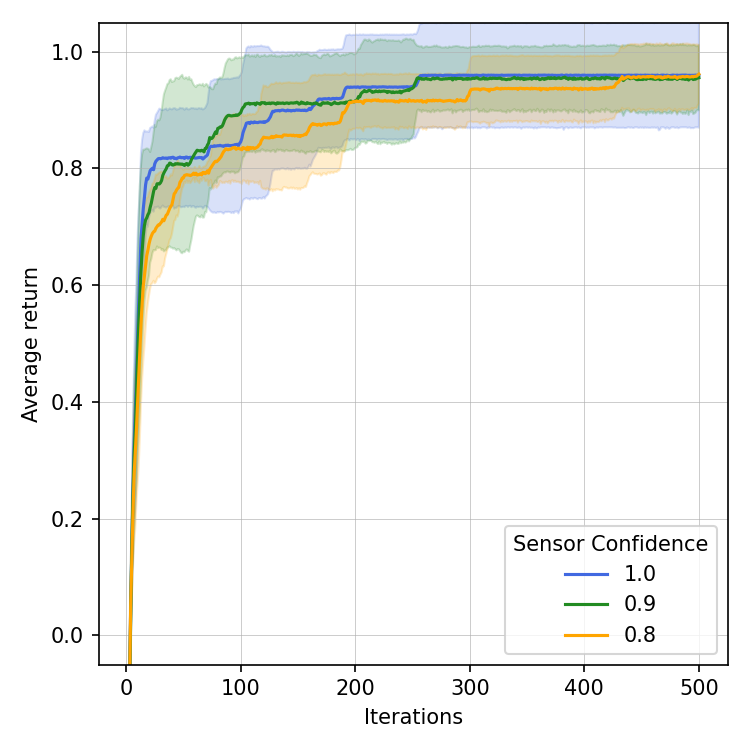}
        \caption{Without reward shaping}
    \end{subfigure}
    \begin{subfigure}[t]{0.4\linewidth}
        \centering
        \includegraphics[width=\linewidth]{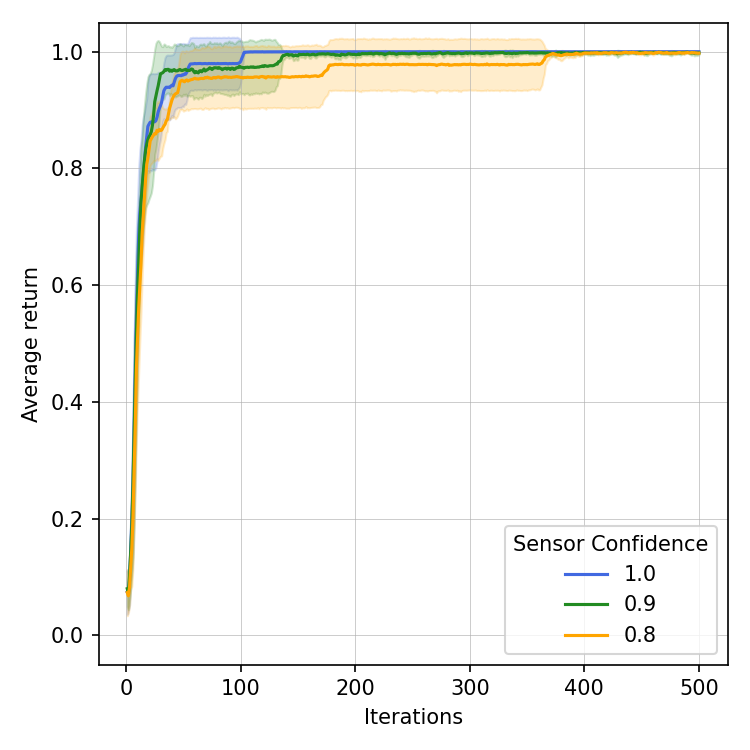}
        \caption{With reward shaping}
    \end{subfigure}
    \caption{\thesystem learning curves for \owcoffeetask with and without probabilistic reward shaping.}
    \Description{Two panels of Prob-IRM learning curves for the Coffee task, without and with probabilistic reward shaping. Each panel plots average return (vertical axis, 0 to 1) against training iterations (horizontal axis), with one curve per sensor confidence level (1.0, 0.9 and 0.8); lines show the mean and shaded bands the standard deviation over runs. Both variants reach similar final returns; with reward shaping the curves rise faster.}
    \label{fig:reward_shaping}
\end{figure}

\Cref{fig:reward_shaping} compares \thesystem with and without probabilistic reward shaping (see \Cref{subsec:probabilistic_reward_shaping}) on \owcoffeetask. Both variants reach similar final performance, but reward shaping accelerates convergence and achieves slightly higher returns. This speed-up occurs because shaped rewards provide intermediate signals upon partial task completion.

Unlike our previous work~\cite{parac2024learning}, we observe little difference in this benefit across noise levels, and in all cases, a single RM learning iteration suffices to find a robust RM. We attribute this to collecting more traces before learning the first RM, reducing reliance on reward shaping to generate informative counterexamples.

\subsubsection{Multiple Noise Sources}
\label{sec:mns}
We assess \thesystem's robustness to multiple independent noise sources using the \emph{noise-all} configuration on \owcoffeetask, where $\owcoffee$, $\owdecor$, and $\owoffice$ sensors are all noisy, tripling noise exposure compared to \emph{noise-first} (\ref{rq:perfect_rm}).

\begin{figure}[ht]
	\centering
	\begin{subfigure}[t]{\linewidth}
		\centering
		\includegraphics[width=0.35\linewidth]{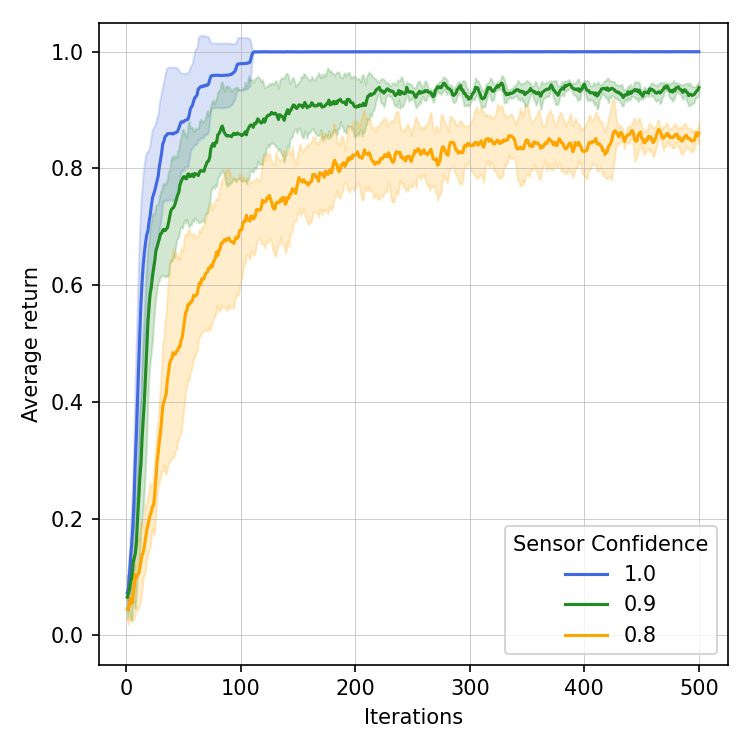}
		\quad
		\resizebox{0.495\linewidth}{!}{\begin{tikzpicture}[shorten >=1pt,node distance=2.06cm,on grid,auto,every initial by arrow/.style ={-Latex}]
    \node[state,initial,initial text=] (u_0)   {$\initialrmstate$};
    \node[state,accepting] (u_acc) [below =3cm of u_0]  {$\acceptingrmstate$};
    \node[state] (u_1) [left =3.5cm of u_acc]   {$\rmstate_1$};
    \node[state] (u_rej) [right =3.5cm of u_acc]  {$\rejectingrmstate$};
    
    \path[-Latex] (u_0) edge [loop above] node {$\langle \emph{o.w.}, 0 \rangle$} ();
    \path[-Latex] (u_1) edge [loop below] node {$\langle \emph{o.w.}, 0 \rangle$} ();
    \path[-Latex] (u_acc) edge [loop right] node {$\langle \emph{o.w.}, 0 \rangle$} ();
    \path[-Latex] (u_rej) edge [loop below] node {$\langle \emph{o.w.}, 0 \rangle$} ();
    
    \path[-Latex] (u_0) edge [bend right] node[in place] {$ \langle \owcoffee \land \neg \owoffice \land \neg \owdecor, 0 \rangle$} (u_1);
    \path[-Latex] (u_0) edge [bend left] node[in place] {$\langle \owdecor, 0 \rangle$} (u_rej);
    \path[-Latex] (u_1) edge[bend right] node[in place] {$\langle \owdecor, 0 \rangle$} (u_rej);
    \path[-Latex] (u_1) edge[bend left] node[in place] {$\langle \owoffice\land\neg\owdecor, 1 \rangle$} (u_acc);
    \path[-Latex] (u_0) edge node[in place] {$\langle \owcoffee \land \owoffice \land\neg\owdecor, 1 \rangle$} (u_acc);
\end{tikzpicture}}
		\caption{Perfect RM}
	\end{subfigure}

	\begin{subfigure}[t]{\linewidth}
		\centering
		\includegraphics[width=0.35\linewidth]{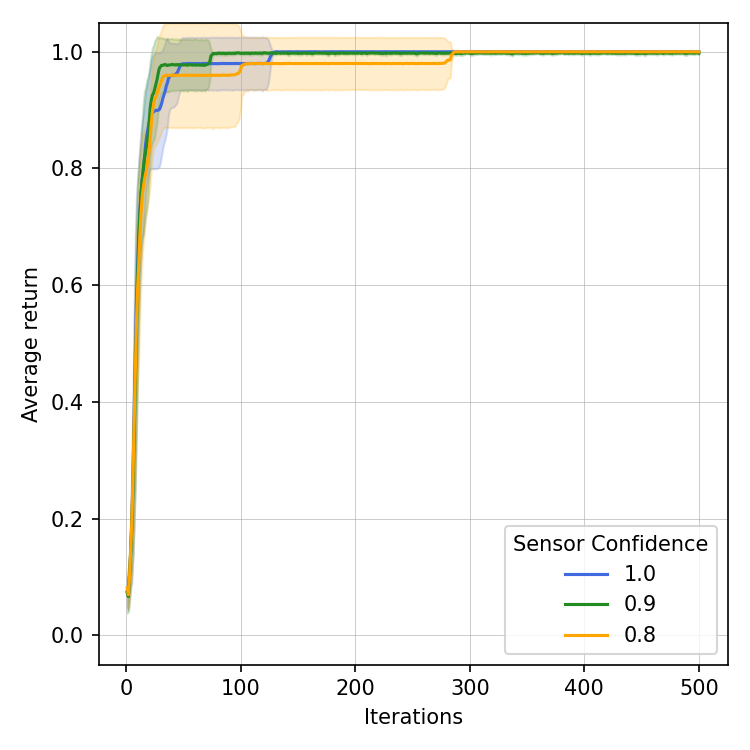}
		\quad
		\resizebox{0.495\linewidth}{!}{\begin{tikzpicture}[shorten >=1pt,node distance=2.06cm,on grid,auto,every initial by arrow/.style ={-Latex}]
    \node[state,initial,initial text=] (u_0)   {$\initialrmstate$};
    \node[state,accepting] (u_acc) [below =3cm of u_0]  {$\acceptingrmstate$};
    \node[state] (u_1) [left =3.5cm of u_acc]   {$\rmstate_1$};
    \node[state] (u_rej) [right =3.5cm of u_acc]  {$\rejectingrmstate$};
    
    \path[-Latex] (u_0) edge [loop above] node {$\langle \emph{o.w.}, 0 \rangle$} ();
    \path[-Latex] (u_1) edge [loop below] node {$\langle \emph{o.w.}, 0 \rangle$} ();
    \path[-Latex] (u_acc) edge [loop right] node {$\langle \emph{o.w.}, 0 \rangle$} ();
    \path[-Latex] (u_rej) edge [loop below] node {$\langle \emph{o.w.}, 0 \rangle$} ();
    
    \path[-Latex] (u_0) edge [bend right] node[in place] {$ \langle \owcoffee \land \neg \owdecor, 0 \rangle$} (u_1);
    \path[-Latex] (u_0) edge [bend left] node[in place] {$\langle \owdecor, 0 \rangle$} (u_rej);
    \path[-Latex] (u_1) edge[bend right] node[in place] {$\langle \owdecor, 0 \rangle$} (u_rej);
    \path[-Latex] (u_1) edge[bend left] node[in place] {$\langle \owoffice\land\neg\owdecor, 1 \rangle$} (u_acc);
\end{tikzpicture}}
		\caption{\thesystem}
	\end{subfigure}
	\caption{Learning curves and exploited RMs for \owcoffeetask in the \emph{noise-all} setting.}
	\Description{Two rows, each pairing a learning-curve plot with the reward machine that was exploited, for the Coffee task with all sensors noisy. Each panel plots average return (vertical axis, 0 to 1) against training iterations (horizontal axis), with one curve per sensor confidence level (1.0, 0.9 and 0.8); lines show the mean and shaded bands the standard deviation over runs. The top row shows the Perfect RM baseline with the handcrafted four-state reward machine for the Coffee task, where returns decrease as sensor confidence decreases. The bottom row shows Prob-IRM with its learned reward machine, which has the same four states but a simpler edge from u0 to u1 labelled coffee and not decoration and no direct edge from u0 to the accepting state; its returns are close to 1 at all confidence levels.}
	\label{fig:noise_all_comparison}
\end{figure}

\Cref{fig:noise_all_comparison} shows the results. Surprisingly, \thesystem outperforms the Perfect RM baseline. This occurs because the learned RM is more robust to noise accumulation: with both $\owcoffee$ and $\owoffice$ sensors noisy, there is a non-zero probability at each step of erroneously transitioning to $\acceptingrmstate$ from $\initialrmstate$. As this probability accumulates over the episode, it may cause premature termination. The learned RM mitigates this by slowing probability mass accumulation towards $\acceptingrmstate$, reducing spurious terminations and simplifying policy learning.

\subsubsection{Pre-trained Noise Model}
\label{subsec:pretrained_noise_model}
To evaluate \thesystem under a different noise model, we replace the probabilistic sensors with pre-trained linear classifiers on \owcoffeetask. We collect observation-label pairs from 1{,}000 random episodes as training data, with the resulting sensor F1 scores in \Cref{tab:pretrained_sensor_scores}. 

This noise model differs substantially from our main experiments: sensor confidence varies across events, noise includes both false positives and false negatives, and any event may be detected regardless of task relevance.

\begin{figure}[ht]
    \centering
    \includegraphics[width=0.45\linewidth]{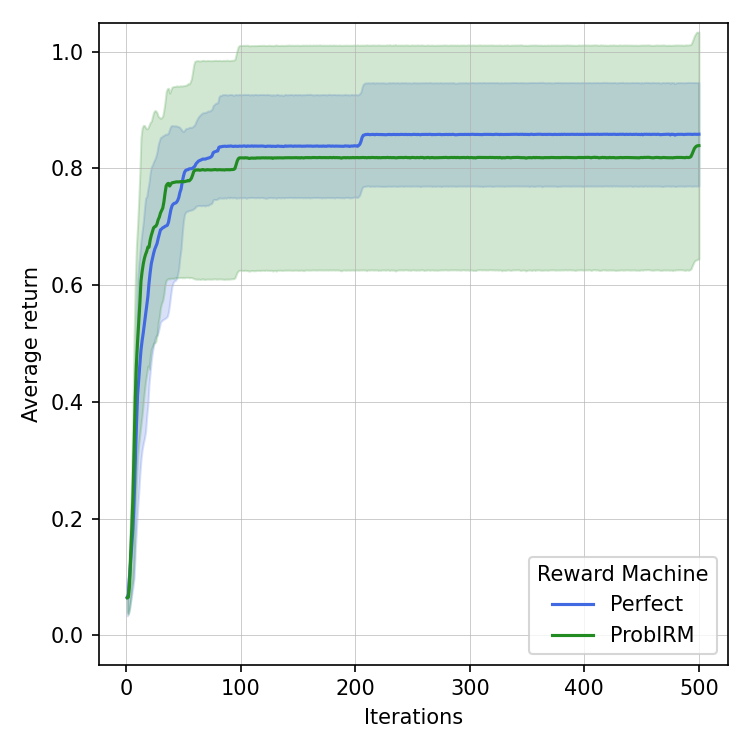}
    \caption{Learning curves for \owcoffeetask with a pre-trained noise model.}
    \Description{One plot of average return (0 to 1) against training iterations for the Coffee task with pre-trained classifiers as sensors, with one curve for the Perfect RM baseline and one for Prob-IRM; lines show the mean and shaded bands the standard deviation. Both curves level off between 0.8 and 0.9, with Prob-IRM slightly below the Perfect RM baseline.}
    \label{fig:pretrained_noise}
\end{figure}

\Cref{fig:pretrained_noise} compares learning curves for Perfect RM and \thesystem. Both converge to suboptimal performance, confirming this noise model is challenging. Despite this, \thesystem is only slightly worse than the Perfect RM baseline, demonstrating effectiveness under more complex noise models.

\begin{table}[ht]
	\caption{F1 scores for each sensor in the pre-trained sensor model ablation (10 runs).}
	\label{tab:pretrained_sensor_scores}
	\centering
	\begin{tabular}{@{}cccccccccc@{}}
		\toprule
		\multicolumn{2}{c}{\textbf{Sensor}}   & \owcella                  & \owcellb                  &\owcellc                 & \owcelld                 & \owcoffee             & \owmail              & \owoffice             & \owdecor              \\ \midrule
		\multirow{4}{*}{\rotatebox{90}{\textbf{F1 Score}}} & \textbf{Min.}      & 0.60 & 0.70 & 0.80 & 0.70 & 0.87 & 0.80 & 0.60 & 0.40 \\
		& \textbf{Max.}      & 1.00 & 1.00 & 1.00 & 1.00 & 1.00 & 1.00 & 0.90 & 0.80 \\
		& \textbf{Avg.}      & 0.84 & 0.90 & 0.89 & 0.92 & 0.94 & 0.95 & 0.72 & 0.59 \\
		& \textbf{Std. Dev.} & 0.14 & 0.10 & 0.07 & 0.10 & 0.05 & 0.07 & 0.10 & 0.14 \\ \bottomrule
	\end{tabular}%
\end{table}

\subsubsection{Task Length}
\label{exp:task_length_ablation}
To evaluate \thesystem's scalability to longer tasks, we generalise \owvisit to require visiting increasingly long sequences of locations. All agents use sensors at 80\% confidence (our lowest tested level).

\begin{figure}[ht]
    \centering
    \includegraphics[width=0.45\linewidth]{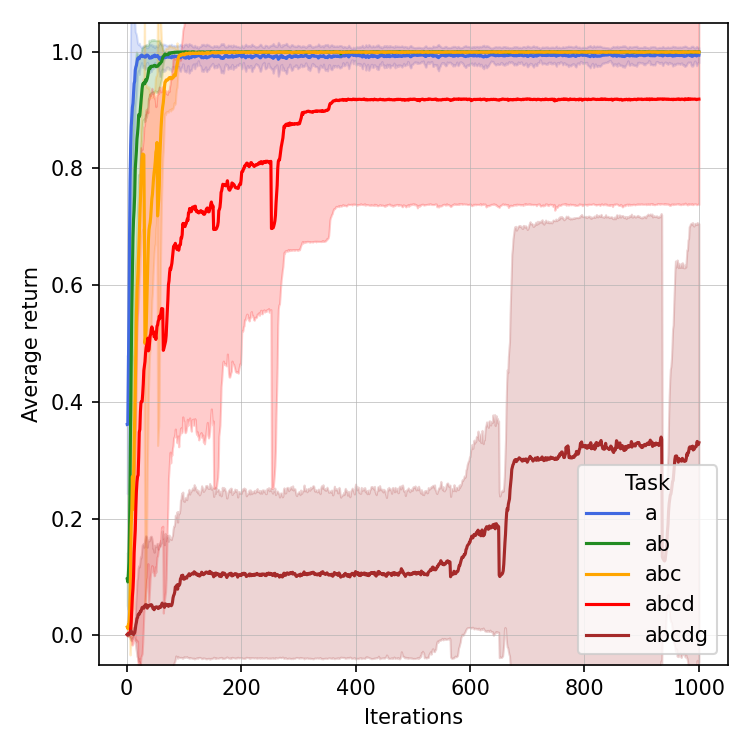}
    \caption{Learning curves for \owvisit tasks with varying sequence lengths (80\% sensor confidence).}
    \Description{One plot of average return (0 to 1) against training iterations for Prob-IRM on visitation tasks of increasing length, with one curve per task (one to five locations); lines show the mean and shaded bands the standard deviation. Tasks with up to three locations converge to returns near 1, the four-location task converges lower with high variance, and the five-location task stays far below the others.}
    \label{fig:long_tasks_ablation}
\end{figure}

\Cref{fig:long_tasks_ablation} shows learning curves. Agents converge to high performance for sequences up to length 3. Performance begins degrading at length 4 and drops drastically at length 5. This is expected: the probability of observing an error-free goal trace decreases exponentially with task length. Without sufficient clean goal traces, learning a correct RM becomes increasingly difficult.

\subsubsection{Belief Updating vs Thresholding}
\label{subsec:belief_updating_thresholding}

We provide empirical justification for our belief updating approach (\Cref{subsec:rm_state_belief}) over thresholding. Thresholding treats any proposition with probability above a fixed threshold as true, avoiding belief maintenance but introducing a critical drawback: under our sensor model, it either works perfectly or fails to learn any RM.

\begin{figure}[ht]
    \centering
    \includegraphics[width=0.45\linewidth]{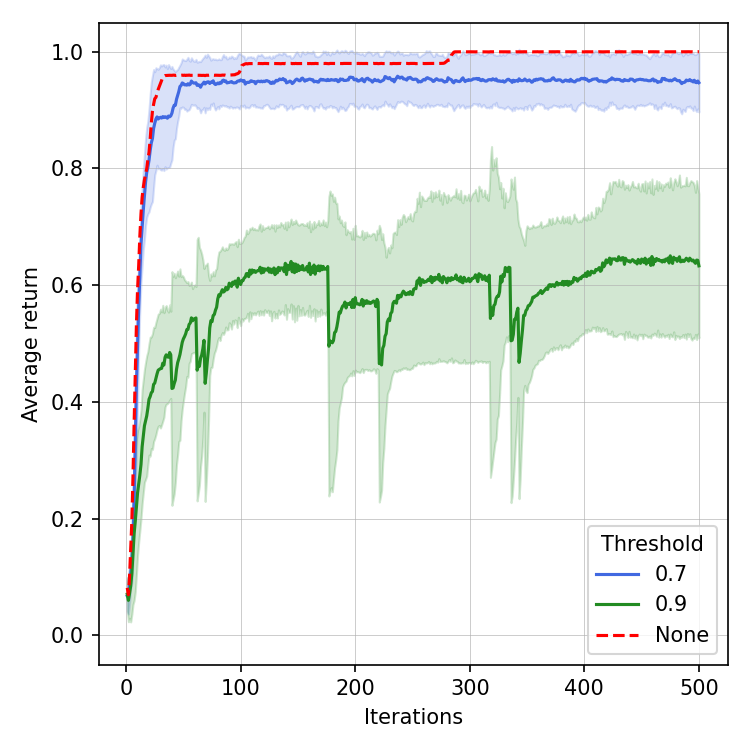}
    \caption{Learning curves for \owcoffeetask (\emph{noise-all}, 80\% confidence) comparing thresholding at 0.7 and 0.9 with belief updating (dashed line).}
    \Description{One plot of average return (0 to 1) against training iterations for the Coffee task, with a curve for a threshold of 0.7, a curve for a threshold of 0.9 and a dashed curve for belief updating without a threshold; lines show the mean and shaded bands the standard deviation. Belief updating and the 0.7 threshold reach returns above 0.9, while the 0.9 threshold remains substantially lower and unstable.}
    \label{fig:thresholding}
\end{figure}

We train \thesystem agents on \owcoffeetask under \emph{noise-all} with 80\% sensor confidence, using thresholds of 0.7 (below) and 0.9 (above). \Cref{fig:thresholding} shows the results. The 0.7 threshold achieves good performance by correctly identifying propositions. However, the 0.9 threshold fails: it incorrectly rejects \emph{every} proposition, making RM learning impossible.

This highlights thresholding's brittleness: success requires either precise knowledge of the noise posterior or careful threshold tuning, which are both restrictive in practice. In contrast, belief updating requires no such knowledge, introduces no additional hyperparameters, and achieves strong performance in the same setting (\Cref{fig:deliver_coffee_comparison}).

\subsubsection{Cross-Entropy Threshold}
\label{exp:cross_entropy_threshold}

We evaluate the impact of the cross-entropy relearning threshold $\rmrelearningthreshold$ on \owcoffeemailtask, where RM relearning is triggered multiple times. We test values from 0 (relearn every iteration) to 2 (relearn when the model assigns less than 13.5\% probability to the correct class). 

\begin{figure}[ht]
    \centering
    \begin{subfigure}[t]{0.19\linewidth}
        \centering
        \includegraphics[width=\linewidth]{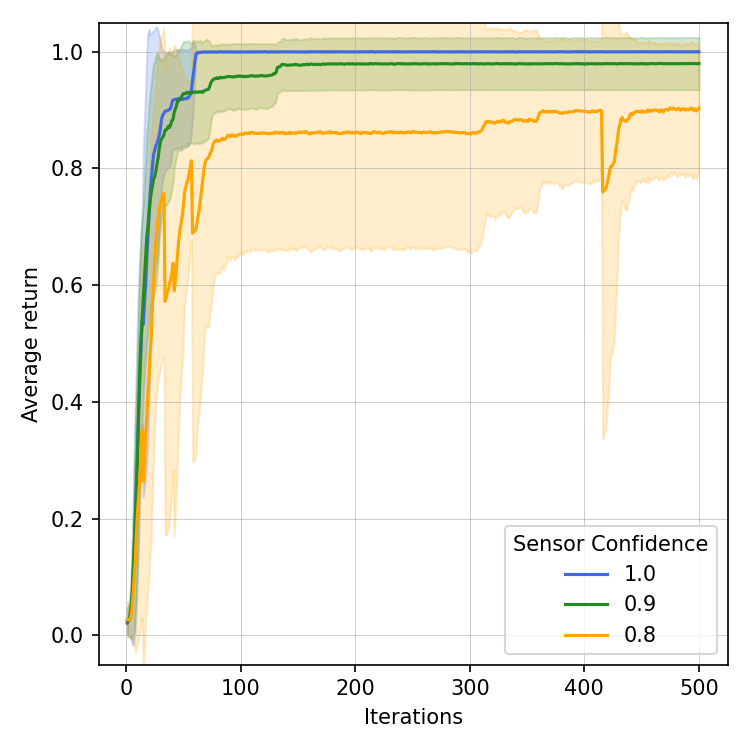}
        \caption{$\rmrelearningthreshold= 0$}
    \end{subfigure}
    \hfill
    \begin{subfigure}[t]{0.19\linewidth}
        \centering
        \includegraphics[width=\linewidth]{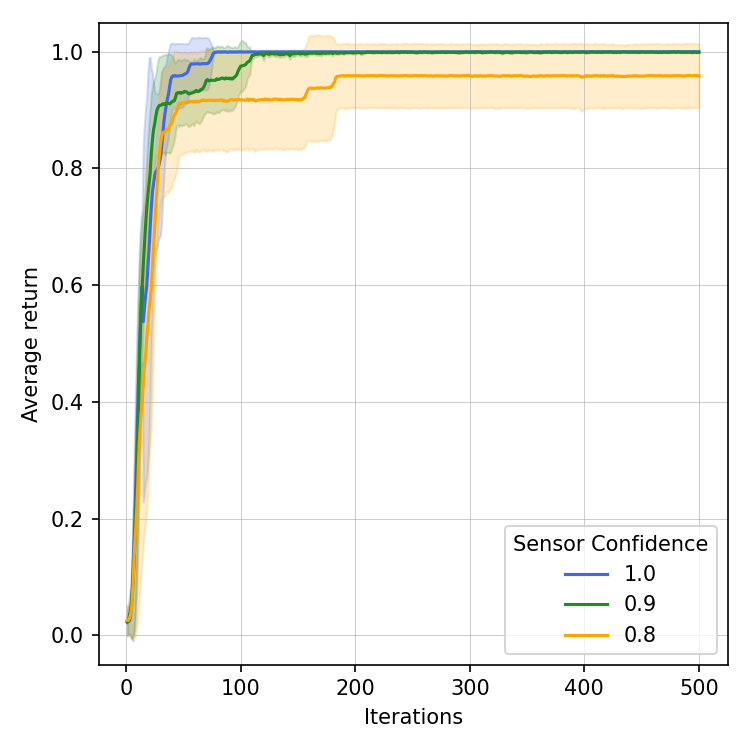}
        \caption{$\rmrelearningthreshold= 0.5$}
    \end{subfigure}
    \hfill
    \begin{subfigure}[t]{0.19\linewidth}
        \centering
        \includegraphics[width=\linewidth]{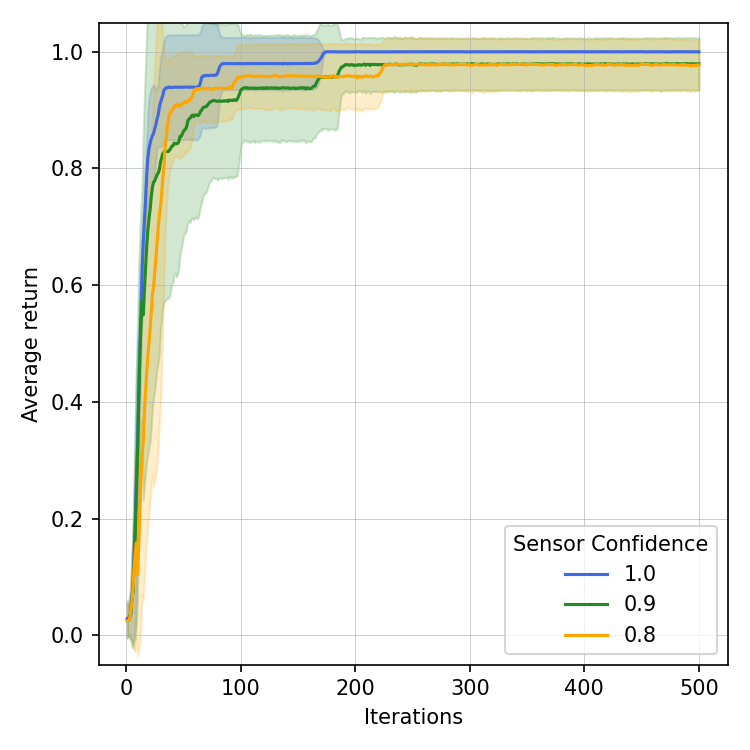}
        \caption{$\rmrelearningthreshold= 1$}
    \end{subfigure}
    \hfill
    \begin{subfigure}[t]{0.19\linewidth}
        \centering
        \includegraphics[width=\linewidth]{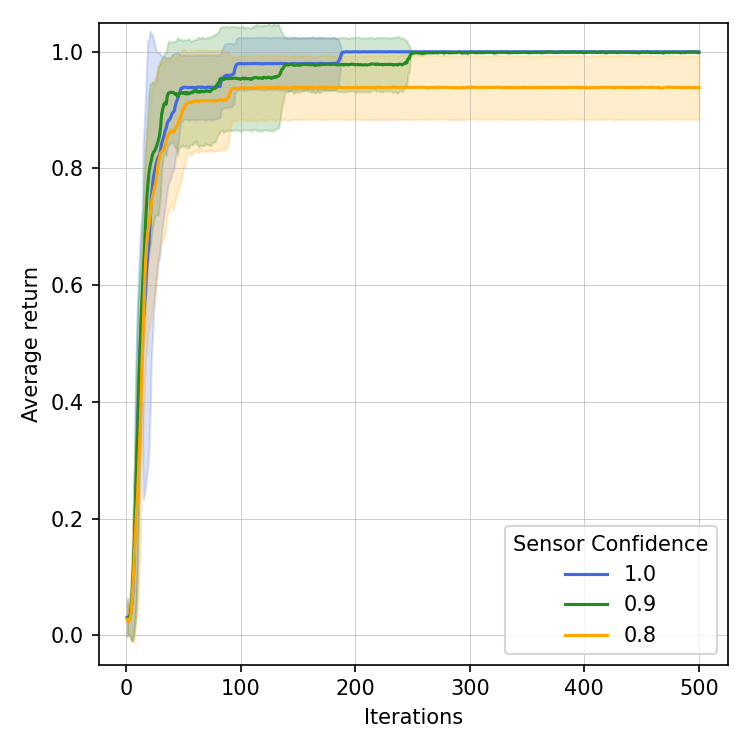}
        \caption{$\rmrelearningthreshold= 1.5$}
    \end{subfigure}
    \hfill
    \begin{subfigure}[t]{0.19\linewidth}
        \centering
        \includegraphics[width=\linewidth]{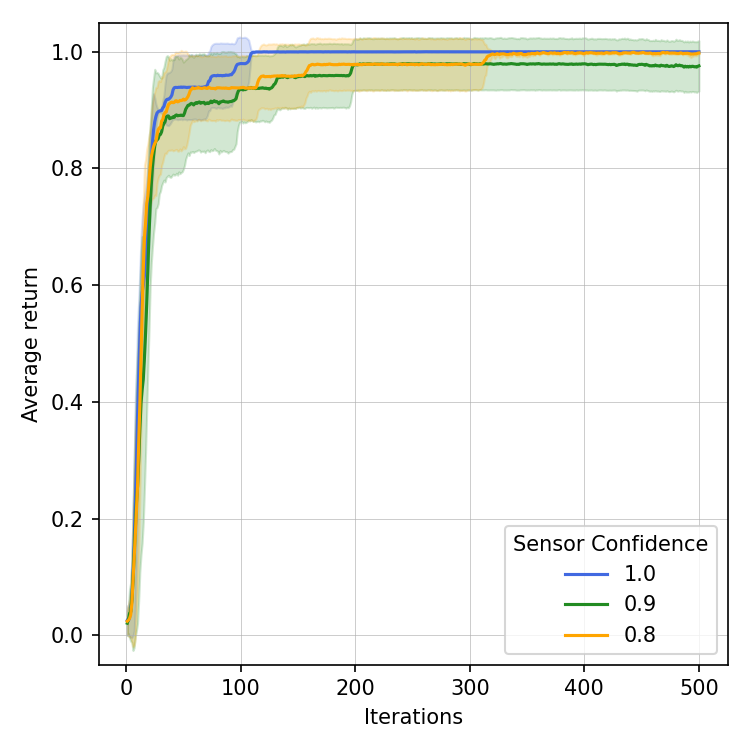}
        \caption{$\rmrelearningthreshold= 2$}
    \end{subfigure}
    \caption{\thesystem learning curves for different values of the cross-entropy threshold $\rmrelearningthreshold$.}
    \Description{Five panels of Prob-IRM learning curves, one per value of the cross-entropy threshold (0, 0.5, 1, 1.5 and 2). Each panel plots average return (vertical axis, 0 to 1) against training iterations (horizontal axis), with one curve per sensor confidence level (1.0, 0.9 and 0.8); lines show the mean and shaded bands the standard deviation over runs. The curves are similar across threshold values.}
    \label{fig:cross-entropy-threshold}
\end{figure}

\Cref{fig:cross-entropy-threshold} shows the results. Performance is generally insensitive to this parameter, benefiting from ILASP's scoring function as a verification step after RM learning (\Cref{metodology:relearn_rm}). This verification prevents updates that do not improve example coverage, making the threshold less critical.

\subsubsection{Balancing Example Coverage and RM Simplicity}
\label{exp:ex_fit_vs_rm_simplicity}
The ILASP scoring function (\Cref{def:ilasp_scoring}) balances hypothesis length against uncovered example penalties. By adjusting the total penalty budget, we control the trade-off between example coverage and RM simplicity.

To evaluate this trade-off, we adjust the length function and the example penalty budget (\Cref{sec:scoring_complexity}). For the former, the lengths of $\mathtt{ed}$ and $\notlogicaltransitionfn$ rules are set to 2 and 1, respectively. For the latter, low budgets cause low-penalty examples to be ignored since their penalty rounds to 0. To ensure a fair comparison, we scale the WCDPI example inclusion threshold $\penaltythreshold$ proportionally to the budget, preventing higher budgets from disproportionately including more examples.

\begin{figure}[ht]
    \centering
    \begin{subfigure}[t]{0.19\linewidth}
        \centering
        \includegraphics[width=\linewidth]{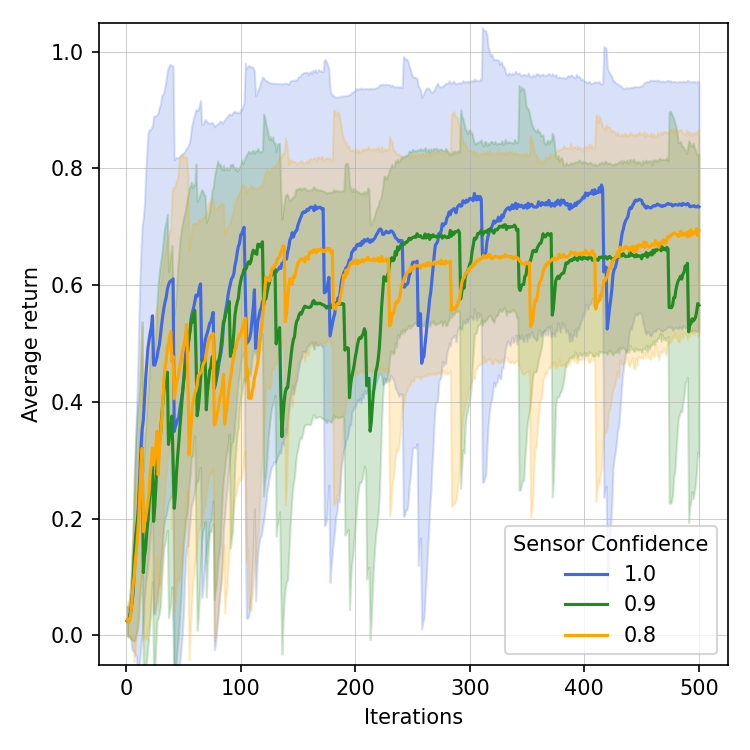}
        \caption{100}
    \end{subfigure}
    \hfill
    \begin{subfigure}[t]{0.19\linewidth}
        \centering
        \includegraphics[width=\linewidth]{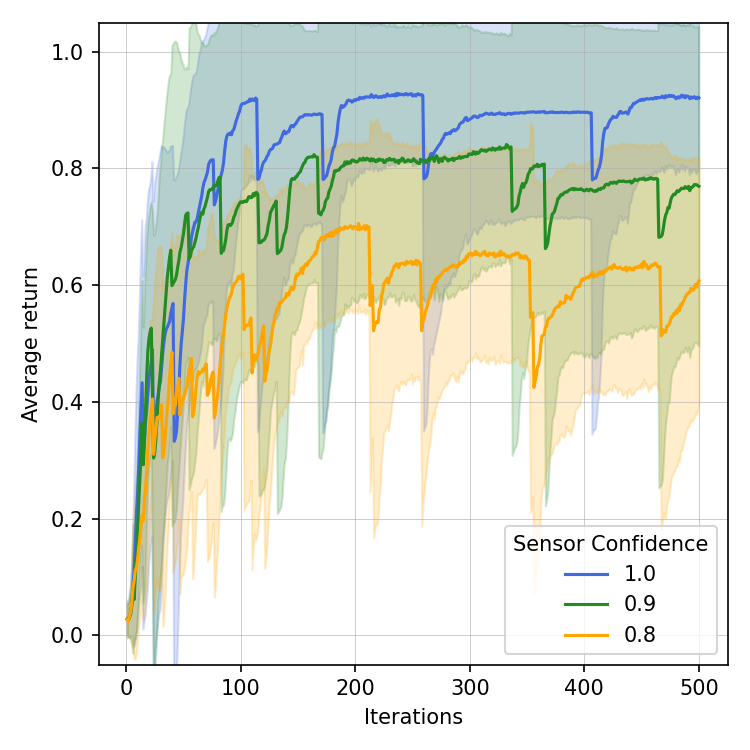}
        \caption{200}
    \end{subfigure}
    \hfill
    \begin{subfigure}[t]{0.19\linewidth}
        \centering
        \includegraphics[width=\linewidth]{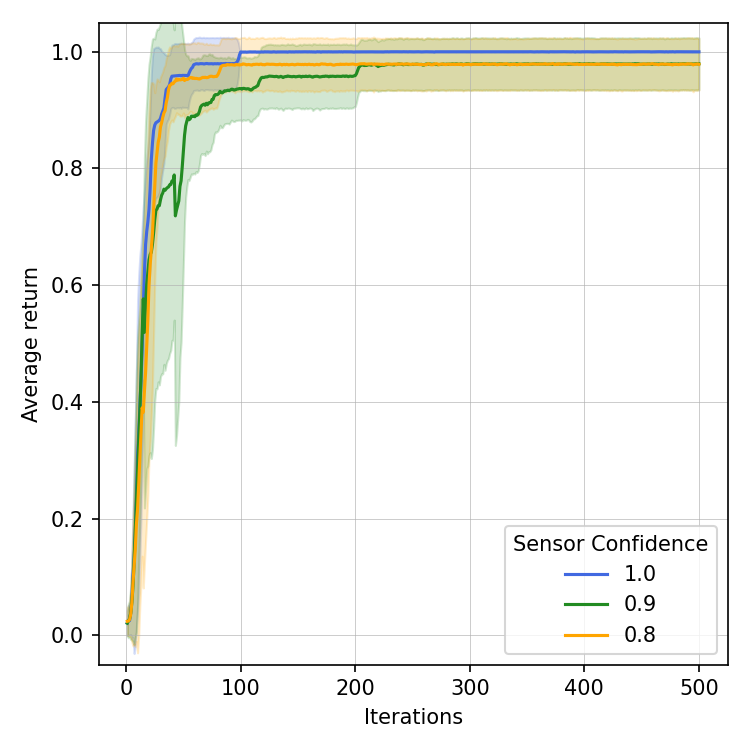}
        \caption{500}
    \end{subfigure}
    \hfill
    \begin{subfigure}[t]{0.19\linewidth}
        \centering
        \includegraphics[width=\linewidth]{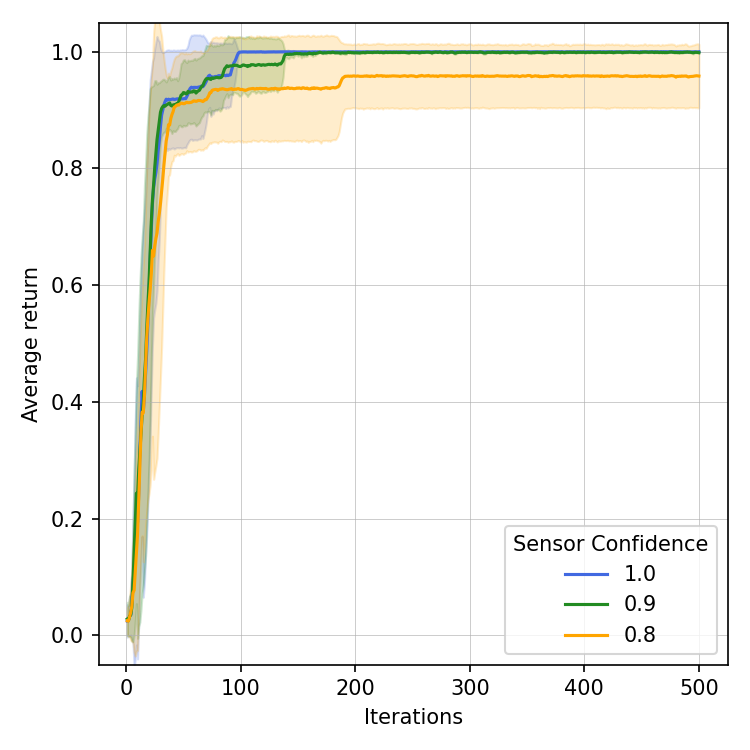}
        \caption{1{,}000}
    \end{subfigure}
    \hfill
    \begin{subfigure}[t]{0.19\linewidth}
        \centering
        \includegraphics[width=\linewidth]{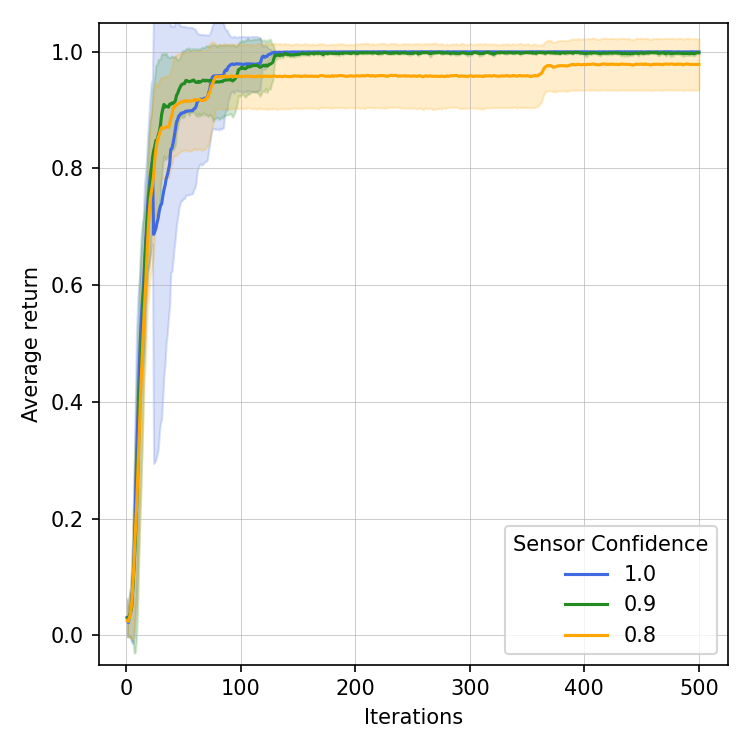}
        \caption{10{,}000}
    \end{subfigure}
    \caption{\thesystem learning curves for different example \emph{penalty budgets} in \owcoffeemailtask.}
    \Description{Five panels of Prob-IRM learning curves, one per example penalty budget (100, 200, 500, 1,000 and 10,000). Each panel plots average return (vertical axis, 0 to 1) against training iterations (horizontal axis), with one curve per sensor confidence level (1.0, 0.9 and 0.8); lines show the mean and shaded bands the standard deviation over runs. The lowest budgets give lower and noisier returns than the higher budgets.}
    \label{fig:fit_of_examples}
\end{figure}

\Cref{fig:fit_of_examples} shows results for \owcoffeemailtask. Low budgets (100, 200) over-emphasise compactness, yielding RMs that fail to solve the task reliably. Higher budgets achieve better performance but with increased variance. Our default is 3{,}200, the smallest value for which ILASP consistently finds solutions within the 60-minute timeout.

\subsubsection{Class-Based Weight Rebalancing}
\label{exp:impact_of_rebalancing}

We evaluate the impact of class-based weight rebalancing on \owcoffeetask. We compare the rebalancing procedure from \Cref{subsec:example_set}, which sets $\classweight = \frac{1}{n_c}$ where $n_c$ is the number of examples of class $c \in \{\text{goal}, \text{dead-end}, \text{incomplete}\}$, against a baseline where $\classweight = 1$ for all examples.

\begin{figure}[ht]
    \begin{subfigure}[t]{0.4\linewidth}
        \centering
        \includegraphics[width=\linewidth]{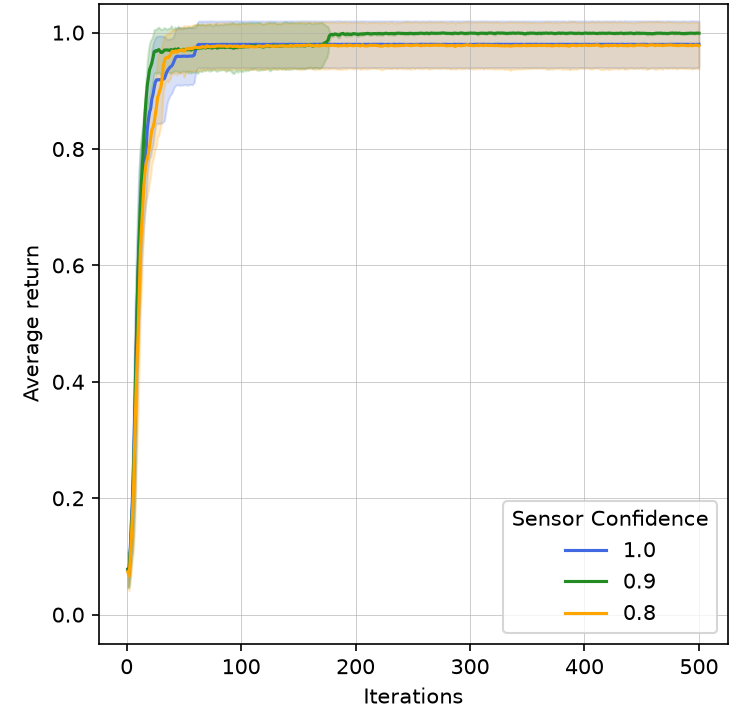}
        \caption{With class-based weight rebalancing}
    \end{subfigure}
    \begin{subfigure}[t]{0.4\linewidth}
        \centering
        \includegraphics[width=\linewidth]{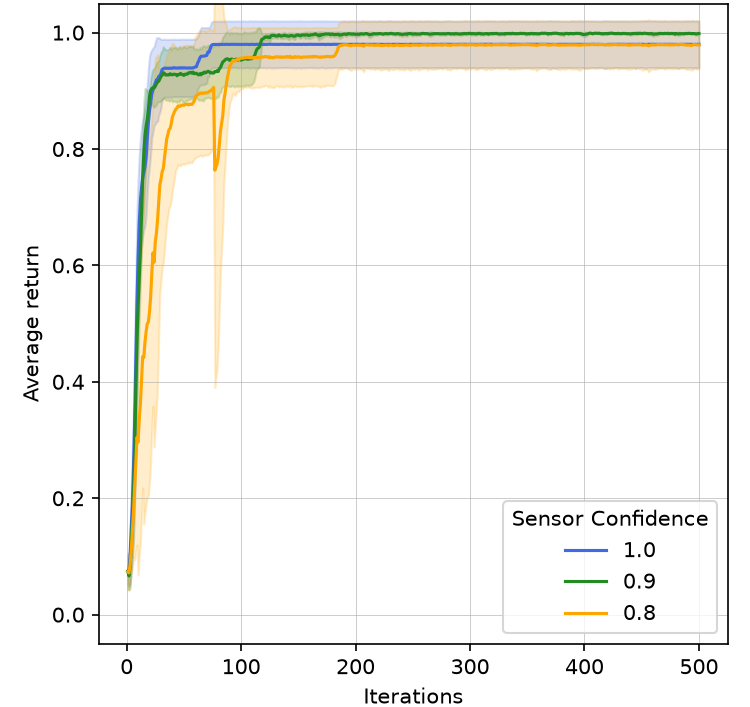}
        \caption{Without class-based weight rebalancing}
    \end{subfigure}
    \caption{\thesystem learning curves for \owcoffeetask when class-based weight rebalancing is enabled and disabled.}
    \Description{Two panels of Prob-IRM learning curves for the Coffee task, with and without class-based weight rebalancing. Each panel plots average return (vertical axis, 0 to 1) against training iterations (horizontal axis), with one curve per sensor confidence level (1.0, 0.9 and 0.8); lines show the mean and shaded bands the standard deviation over runs. With rebalancing the curves converge faster and more stably.}
    \label{fig:rebalancing_impact}
\end{figure}

\Cref{fig:rebalancing_impact} shows that rebalancing improves both convergence speed and stability. Without rebalancing, the learned RM initially captures only negative transitions (avoiding $\owdecor$), since dead-end examples are more likely than goal ones under a random policy. As the agent learns to avoid $\owdecor$, the example distribution becomes more informative, but rebalancing accelerates this by ensuring goal examples are adequately weighted.

\subsubsection{Example Generation from Previous RMs}
\label{exp:example_generation_previous_rms}
\begin{figure}[ht]
    \centering
    \includegraphics[width=0.45\linewidth]{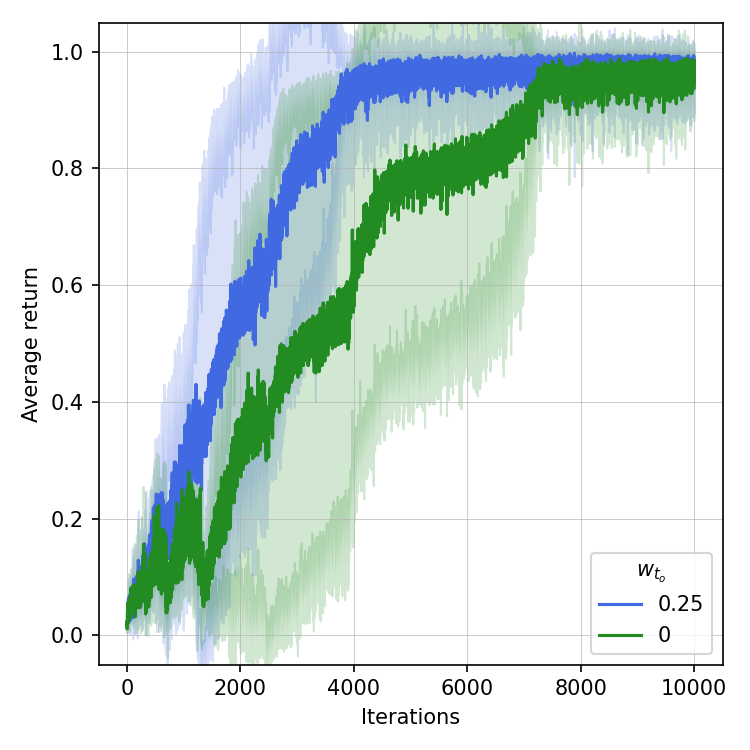}
    \caption{\thesystem learning curves for \wwrgbctask under 80\% sensor confidence with ($\traceoriginmultiplier = 0.25$) and without ($\traceoriginmultiplier = 0$) example generation from previous RMs.}
    \Description{One plot of average return (0 to 1) against training iterations for the RGBC task at 80 percent sensor confidence, with one curve for a trace origin multiplier of 0.25 and one for a multiplier of 0; lines show the mean and shaded bands the standard deviation. The curve with multiplier 0.25 rises markedly faster; both approach returns near 1 by the end of training.}
    \label{fig:previous_rm_examples}
\end{figure}

We evaluate the impact of generating examples from previously learned RMs (\Cref{sec:example_gen_rms}). \Cref{fig:previous_rm_examples} compares \thesystem with ($\traceoriginmultiplier = 0.25$) and without ($\traceoriginmultiplier = 0$) this technique on \wwrgbctask under 80\% sensor confidence. Including examples from previous RMs substantially improves convergence speed, confirming the effectiveness of this approach.

\subsection{RM Learning Computational Overhead}
\label{sec:execution_times} 
We evaluate ILASP runtime as a percentage of total execution time across all \thesystem experiments. \Cref{tab:ilasp_runtimes} summarises the results.

For \owcoffeetask and \owcoffeemailtask, ILASP cost is negligible across all noise levels, remaining below 1\% and typically below 0.01\%. For \owvisit, cost is similarly low at 90\% and 100\% sensor confidence, but increases substantially at 80\%, where ILASP accounts for roughly 25\% of total runtime on average, reaching 75\% in the most challenging instances. \ww tasks show the same trend.

In general, ILASP cost correlates positively with both task complexity and noise level. Two factors drive this: first, as training progresses, the growing number of examples makes each ILASP task harder to solve. Second, ILASP's optimality guarantee (\Cref{def:ilasp_scoring}) requires progressively increasing a score threshold until a solution is found, meaning harder tasks require more iterations.

\begin{table}[ht]
	 \caption{ILASP runtime as a percentage of total training time across tasks and sensor confidence levels.}
	\label{tab:ilasp_runtimes}
    \centering
    \renewcommand{\arraystretch}{1.25}
    \resizebox{0.6\textwidth}{!}{%
        \begin{tabular}{@{}lrrlrrrrr@{}}
            \toprule
            \textbf{Domain} & \textbf{Noise} &  &  &
            \textbf{Min.}   & \textbf{Median} & \textbf{Max.} &
            \textbf{Average} & \textbf{Std. Dev.} \\
            \midrule

            % ---------------------- deliver_coffee ----------------------
            \owcoffeetask & 80\%  & & & {\color[HTML]{c0c0c0} \textless 0.01\%} & {\color[HTML]{c0c0c0} \textless 0.01\%} & {\color[HTML]{c0c0c0} \textless 0.01\%} & {\color[HTML]{c0c0c0} \textless 0.01\%} & {\color[HTML]{c0c0c0} \textless 0.01\%} \\
            \cdashline{2-9}[0.5pt/5pt]

                           & 90\%  & & & {\color[HTML]{c0c0c0} \textless 0.01\%} & {\color[HTML]{c0c0c0} \textless 0.01\%} & {\color[HTML]{c0c0c0} \textless 0.01\%} & {\color[HTML]{c0c0c0} \textless 0.01\%} & {\color[HTML]{c0c0c0} \textless 0.01\%} \\
            \cdashline{2-9}[0.5pt/5pt]

                           & 100\% & & & {\color[HTML]{c0c0c0} \textless 0.01\%} & {\color[HTML]{c0c0c0} \textless 0.01\%} & {\color[HTML]{c0c0c0} \textless 0.01\%} & {\color[HTML]{c0c0c0} \textless 0.01\%} & {\color[HTML]{c0c0c0} \textless 0.01\%} \\
            \hd

            % ---------------------- coffee_mail ----------------------
            \owcoffeemailtask & 80\%  & & & 0.05\% & 0.10\% & 0.96\% & 0.31\% & 0.31\% \\
            \cdashline{2-9}[0.5pt/5pt]

                        & 90\%  & & & 0.01\% & 0.05\% & 0.96\% & 0.21\% & 0.34\% \\
            \cdashline{2-9}[0.5pt/5pt]

                        & 100\% & & & {\color[HTML]{c0c0c0} \textless 0.01\%} & 0.01\% & 0.05\% & 0.02\% & 0.01\% \\
            \hd

            % ---------------------- visit_abcd ----------------------
            \owvisit & 80\%  & & & 0.04\% & 20.32\% & 75.42\% & 27.14\% & 21.69\% \\
            \cdashline{2-9}[0.5pt/5pt]

                       & 90\%  & & & 0.02\% & 0.08\% & 8.54\% & 1.75\% & 3.11\% \\
            \cdashline{2-9}[0.5pt/5pt]

                       & 100\% & & & 0.05\% & 0.20\% & 0.26\% & 0.18\% & 0.07\% \\
            \midrule

            % ---------------------- waterworld_rgb ----------------------
            \wwrgbtask & 80\%  & & & {\color[HTML]{c0c0c0} \textless 0.01\%} & 2.36\% & 7.10\% & 2.40\% & 2.31\% \\
            \cdashline{2-9}[0.5pt/5pt]

                           & 90\%  & & & {\color[HTML]{c0c0c0} \textless 0.01\%} & 0.02\% & 1.09\% & 0.23\% & 0.43\% \\
            \cdashline{2-9}[0.5pt/5pt]

                           & 100\% & & & 0.02\% & 0.04\% & 0.08\% & 0.05\% & 0.02\% \\
            \hd

            % ---------------------- waterworld_rg_b ----------------------
            \wwrgminusbtask & 80\%  & & & {\color[HTML]{c0c0c0} \textless 0.01\%} & 0.29\% & 2.23\% & 0.56\% & 0.83\% \\
            \cdashline{2-9}[0.5pt/5pt]

                            & 90\%  & & & {\color[HTML]{c0c0c0} \textless 0.01\%} & 0.42\% & 1.31\% & 0.53\% & 0.52\% \\
            \cdashline{2-9}[0.5pt/5pt]

                            & 100\% & & & 0.42\% & 1.28\% & 2.40\% & 1.38\% & 0.70\% \\
            \hd

            % ---------------------- waterworld_rgbc ----------------------
            \wwrgbctask & 80\%  & & & 0.07\% & 0.48\% & 30.66\% & 5.83\% & 11.27\% \\
            \cdashline{2-9}[0.5pt/5pt]

                            & 90\%  & & & 0.28\% & 0.41\% & 15.24\% & 2.99\% & 5.32\% \\
            \cdashline{2-9}[0.5pt/5pt]

                            & 100\% & & & 0.20\% & 0.78\% & 13.53\% & 3.90\% & 4.85\% \\
            \bottomrule
        \end{tabular}%
    }
\end{table}
\section{Related Work}
\label{sec:related}

In this section, we discuss work related to \thesystem, organised into four areas: learning RMs without noise (\Cref{subsec:learning_rms_without_noise}), uncertainty-aware RM exploitation (\Cref{subsec:related_uncertain_rm_exploiation}), learning RMs under uncertainty (\Cref{subsec:learning_rm_uncertainty}), and inductive logic programming with noisy examples (\Cref{subsec:ilp_noisy}).

\subsection{Learning Reward Machines without Noise}
\label{subsec:learning_rms_without_noise}
Since \citet{ToroIcarteKVM18} introduced reward machines, numerous methods have been proposed for learning them from traces, including active learning~\cite{gaon2020reinforcement}, discrete optimisation~\cite{ToroIcarteWKVCM19,christoffersen2023learning,ToroIcarteWKVCM23}, inductive logic programming~\cite{furelos2020induction,FurelosBlancoLJBR21,FurelosBlancoLJBR23,ArdonFR23,ardon2024texttt}, program synthesis~\cite{HasanbeigJAMK21,hasanbeig2024symbolic}, SAT solving~\cite{XuGAMNTW20,CorazzaGN22}, and state-merging~\cite{gaon2020reinforcement}. However, these methods assume deterministic labelling functions and cannot handle label noise.

We build upon \citet{FurelosBlancoLJBR21}, which uses ILASP~\cite{law2015ilasp}, an inductive logic programming system whose weighted example mechanism makes it well suited for extension to noisy settings. However, this approach learns minimal RMs through iterative deepening, a property we cannot maintain under noise as discussed in \Cref{subsec:interleaved_algorithm}. Additionally, the method can suffer from overgeneralisation, which we describe in \Cref{sec:discussion}.

\subsection{Uncertainty-Aware Exploitation of Reward Machines}
\label{subsec:related_uncertain_rm_exploiation}
Traditional RM exploitation methods assume perfect labelling functions. These include QRM~\cite{ToroIcarteKVM18}, CRM~\cite{ToroIcarteKVM22}, and hierarchical RL approaches~\cite{ToroIcarteKVM18,ToroIcarteKVM22,FurelosBlancoLJBR23,FurelosBlancoLJBR21}. Recent work uses deep RL algorithms such as PPO, encoding the current RM state via one-hot vectors~\cite{ardon2024texttt} or graph embeddings~\cite{FurelosBlancoPKSRD26,YalcinkayaLVS23,YalcinkayaLVS24}. However, these algorithms suffer significant performance degradation under sensor noise due to erroneous labels causing agents to skip necessary subtasks or pursue incorrect objectives~\cite{nodari2023adversarial}.

\citet{LiCVKTM22,li2024reward} evaluate several exploitation strategies under the noisy setting formalised in \Cref{def:noisylabelfun}, including thresholding (converting probabilistic labels to deterministic ones by treating any proposition with probability above 0.5 as occurring), independent belief updating (maintaining a distribution over RM states), recurrent policies (bypassing labels entirely), and RNN-based belief estimation (which requires access to perfect labels during training). \citet{LiCVKTM22} combine these with QRM, while \citet{li2024reward} use PPO. We adopt the independent belief updating approach, conditioning PPO on the current RM state belief vector (\Cref{subsec:rm_state_belief}), and compare against thresholding (\Cref{subsec:belief_updating_thresholding}) and recurrent policies (\Cref{sec:officeworld_results,sec:waterworld_results}) in our experiments.

\citet{hasanbeig2019reinforcement} also consider label uncertainty within a broader framework for uncertain transition dynamics. While their formulation permits probabilistic labelling, it does not explicitly include labelling or termination functions.

\subsection{Learning Reward Machines under Uncertainty}
\label{subsec:learning_rm_uncertainty}

Several approaches have addressed uncertainty in different components of the RM learning problem. \citet{DBLP:conf/aips/DohmenTAB0V22} address uncertainty in RM transitions, learning RMs with probabilistic transition functions by adapting the $\textnormal{L}^\ast$ active learning algorithm~\cite{angluin1987learning}. \citet{CorazzaGN22} address uncertainty in rewards, learning RMs with stochastic reward emissions. In contrast, we address uncertainty in labels, learning deterministic RMs from noisy labels.

The work most similar to ours is due to \citet{verginis2024joint}, who also tackle noisy labels. They adapt JIRP~\cite{XuGAMNTW20}, a SAT-based approach that learns RMs by iteratively covering counterexamples, similarly to the work we build upon~\cite{FurelosBlancoLJBR21}. Since JIRP does not handle uncertainty, they use thresholding to convert noisy labels to deterministic ones while simultaneously learning an improved labelling function estimate. To handle inconsistencies between candidate RMs and newly observed traces, they maintain two labelling function estimates: one for generating learning examples and one representing the best estimate so far. When these diverge significantly, they restart the learning process. Their approach thus focuses on learning an improved labelling function to approximate the true one, whereas we directly integrate noise into the RM learning procedure through weighted examples.

While their labelling function learning could potentially be combined with our approach, their Bayesian update mechanism requires tracking repeated state visits, which complicates application to continuous state spaces without discretisation.

\subsection{Inductive Logic Programming with Noisy Examples}
\label{subsec:ilp_noisy}

Several works~\cite{DBLP:journals/ml/CunningtonLLR23,cunnington2023symbolic,dreossi2024towards,cunnington2024role} use ILASP to learn logical rules from noisy examples in domains beyond RM learning. Like our approach, \citet{DBLP:journals/ml/CunningtonLLR23} aggregate multiple sources of noise for a single example. However, they use t-norm operators~\cite{klement2013triangular} to combine probabilistic predictions, whereas we use a sampling-based approach with uniform penalty weights.

\section{Discussion}
\label{sec:discussion}

In this section, we discuss the limitations of \thesystem.

\subsection{Belief Overestimation from Temporally Correlated Noisy Labels}
\label{subsec:belief_overestimation}

Our RM state belief update (\Cref{subsec:rm_state_belief}) treats each noisy label as providing new independent evidence for RM state transitions. However, as noted by \citet{li2024reward}, when agents revisit the same state, this causes belief overestimation: noisy labels from the same state reflect the same underlying reality, yet the update rule treats them as independent samples, causing belief to spuriously accumulate.

\begin{example}
	Consider extending the history and trace from \Cref{ex:rm_belief_computation} with three additional visits to the same state:
	\begin{align*}
		\history &= \langle \langle 4,6\rangle, \textnormal{left}, \langle 3,6\rangle, \textnormal{left}, \langle 3,6\rangle, \textnormal{left}, \langle 3,6\rangle \rangle, \\
		\noisytrace &= \langle 
		\{\owcoffee\!:\! 0.0002, \owoffice\!:\! 0.0001\},
		\{\owcoffee\!:\! 0.6513, \owoffice\!:\! 0.0001\},
		\{\owcoffee\!:\! 0.6513, \owoffice\!:\! 0.0001\},
		\{\owcoffee\!:\! 0.6513, \owoffice\!:\! 0.0001\}
		\rangle.
	\end{align*}
	Computing the RM state belief evolution using \Cref{def:rm_state_belief}, we have:
	\begin{itemize}
		\item Step $t=0$. The agent starts in environment state $\statee_0 = \langle 4, 6\rangle$. We compute the initial belief using the noisy label $\noisylabelingfunction(\emptyset, \emptyset, \statee_0) = \set{\owcoffee: 0.0002, \owoffice: 0.0001}$. From \Cref{ex:rm_belief_computation}:
		 \[
		\rmstatebelief_0 \approx 
		\begin{bmatrix}
			0.9998 & 0.0002 & 0.0000 & 0
		\end{bmatrix}^\top.
		\]
		
		\item Step $t=1$. The label is $\noisylabelingfunction(\statee_0, \action_0, \statee_1) = \set{\owcoffee: 0.6513,\; \owoffice: 0.0001}$, where $\statee_0=\langle 4,6\rangle, \action_0=\textnormal{left}$,  $\statee_1=\langle 3, 6\rangle$. From \Cref{ex:rm_belief_computation}:
		\[
		\rmstatebelief_1 \approx 
		\begin{bmatrix}
			0.3486 & 0.6513 & 0.0001 & 0
		\end{bmatrix}^\top.
		\]

		\item Step $t=2$. The agent remains at $\langle 3,6\rangle$ and observes the same noisy label. As such, the resulting belief is:
		\[
		\rmstatebelief_2 \approx
		\begin{bmatrix}
			0.1215 & 0.8783 & 0.0002 & 0
		\end{bmatrix}^\top.
		\]
		
		\item Step $t=3$. The agent remains at $\langle 3,6\rangle$ again. The same noisy label yields:
		\[
		\rmstatebelief_3 \approx
		\begin{bmatrix}
			0.0424 & 0.9573 & 0.0003 & 0
		\end{bmatrix}^\top.
		\]
	\end{itemize}
	
	Despite receiving no new information (the agent remained in the same state), belief in $\rmstate_1$ increased from 0.6513 to 0.9573 due to treating temporally correlated labels as independent.
\end{example}

\citet{li2024reward} address this by training a recurrent neural network to model temporal dependencies in noisy labels. However, their approach requires access to perfect labels during training, limiting applicability to fully noisy settings. Despite not explicitly addressing this limitation, our experiments show minimal practical impact.

\subsection{Overgeneralisation from Temporally Dependent Propositions}
\label{subsec:overgeneralisation}

A second limitation arises when propositions exhibit temporal dependencies. In these cases, the learned RM may be \emph{valid} with respect to all observed traces (\Cref{sec:background_rms}) yet fail to capture causally necessary propositions, causing \emph{overgeneralisation}.

\begin{example}
	Consider a modified \owcoffeetask (\Cref{example-domain}) where the office (\owoffice) can only be reached after having been in the coffee location (\owcoffee) first. \thesystem, like methods learning minimal RMs~\cite[e.g.,][]{FurelosBlancoLJBR21,XuGAMNTW20}, would learn an RM containing only the office transition, omitting coffee entirely despite it being necessary. The learned RM accepts all observed traces but overgeneralises to traces the environment cannot produce.
\end{example}

Interestingly, noisy detections may inadvertently help by breaking strict temporal dependencies between propositions. However, a more principled solution requires generating negative examples: traces that would be accepted by the candidate RM but cannot occur in the environment. \citet{hasanbeig2024symbolic} generate such examples up to a tunable length $l$, repeating learning until no negative examples remain. A similar strategy could extend our approach, though selecting an appropriate $l$ is task-dependent and non-trivial. A more general solution may require explicit causal modelling of proposition dependencies, which we leave for future work.

\subsection{Convergence and Cyclic Relearning}
\label{sec:infinite_cycling}

Our approach builds upon ISA~\cite{FurelosBlancoLJBR21}, which provably converges to a target RM in a finite number of steps (Theorem~7.1). However, a key difference distinguishes our method: while ISA uses unweighted examples that must all be satisfied, \thesystem uses weighted examples with finite penalties. This allows the learner to trade off example coverage against RM complexity, but introduces a potential failure mode: previously covered examples can become uncovered by subsequent RMs. We demonstrate that this can cause infinite cyclic relearning, even in the noiseless setting.

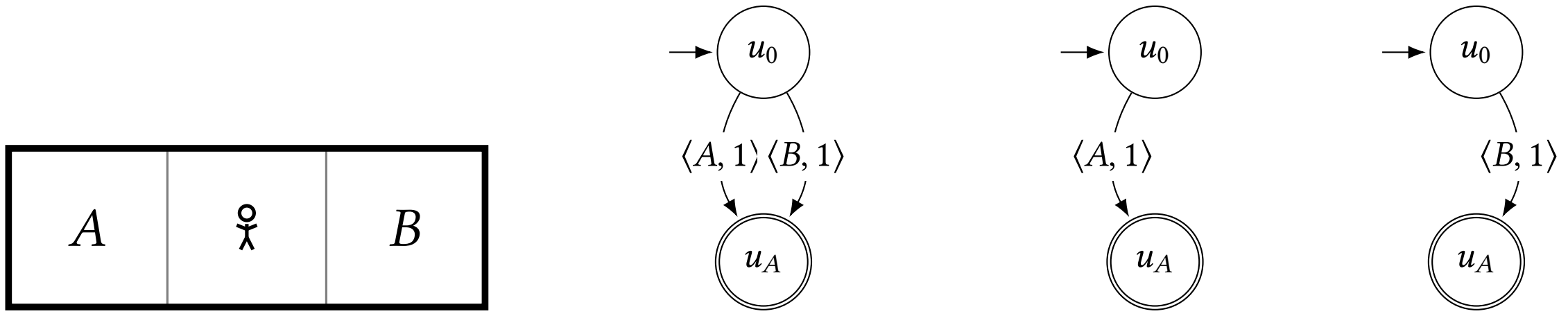
\begin{figure}
    \begin{subfigure}{0.35\linewidth}
        \centering
        \resizebox{0.9\linewidth}{!}{
        \begin{tikzpicture}[scale=1]
    \draw[gray] (0, 0) grid (3, 1);
    \draw[very thick] (0, 0) rectangle (3, 1);
    \node[] at (0.5, 0.5) {\owcella};
    \node[] at (1.5, 0.5) {\owagent};
    \node[] at (2.5, 0.5) {\owcellb};
\end{tikzpicture}
        }
        \caption{Domain}
    \end{subfigure}
    \hfill
    \begin{subfigure}{0.2\linewidth}
        \centering
        \begin{tikzpicture}[shorten >=1pt,node distance=2.06cm,on grid,auto,every initial by arrow/.style ={-Latex}]
        \node[state,initial,initial text=] (u_0)   {$\initialrmstate$};
        \node[state,accepting] (u_{acc}) [below =2cm of u_0]   {$\acceptingrmstate$};
        
        \path[-Latex] (u_0) edge [bend right] node[in place] {$\langle \owcella, 1 \rangle $} (u_{acc});
        \path[-Latex] (u_0) edge [bend left] node[in place] {$ \langle 
        \owcellb, 1 \rangle$} (u_{acc});
\end{tikzpicture}
        \caption{Perfect RM}
    \end{subfigure}
    \hfill
    \begin{subfigure}{0.2\linewidth}
        \centering
        \begin{tikzpicture}[shorten >=1pt,node distance=2.06cm,on grid,auto,every initial by arrow/.style ={-Latex}]
        \node[state,initial,initial text=] (u_0)   {$\initialrmstate$};
        \node[state,accepting] (u_{acc}) [below =2cm of u_0]   {$\acceptingrmstate$};
        
        \path[-Latex] (u_0) edge [bend right] node[in place] {$ \langle 
        \owcella, 1 \rangle$} (u_{acc});
\end{tikzpicture}
        \caption{Learned $\rewardmachine_1$}
        \label{fig:cycling_rm1}
    \end{subfigure}
    \begin{subfigure}{0.2\linewidth}
        \centering
        \begin{tikzpicture}[shorten >=1pt,node distance=2.06cm,on grid,auto,every initial by arrow/.style ={-Latex}]
        \node[state,initial,initial text=] (u_0)   {$\initialrmstate$};
        \node[state,accepting] (u_{acc}) [below =2cm of u_0]   {$\acceptingrmstate$};
        
        \path[-Latex] (u_0) edge [bend left] node[in place] {$ \langle 
        \owcellb, 1 \rangle$} (u_{acc});
\end{tikzpicture}
        \caption{Learned $\rewardmachine_2$}
        \label{fig:cycling_rm2}
    \end{subfigure}
    \caption{Simplified \ow task (Visit $\owcella$ or $\owcellb$) demonstrating cyclic relearning. The perfect RM accepts both propositions, but \thesystem alternates between $\rewardmachine_1$ (accepts only $\owcella$) and $\rewardmachine_2$ (accepts only $\owcellb$) when the edge cost exceeds the example penalty.}
    \Description{Four panels. (a) A one-row grid of three cells: A on the left, the agent in the middle and B on the right. (b) The perfect reward machine, with two edges from the initial state u0 to the accepting state uA, labelled A with reward 1 and B with reward 1. (c) The learned reward machine RM1, with a single edge from u0 to uA labelled A with reward 1. (d) The learned reward machine RM2, with a single edge from u0 to uA labelled B with reward 1.}
    \label{fig:cycling}
\end{figure}

\Cref{fig:cycling} illustrates a simplified \ow task where the agent must visit either $\owcella$ or $\owcellb$, moving only left or right. We set the ILASP example penalty sum (\Cref{exp:ex_fit_vs_rm_simplicity}) to 10 and the cost of a rule representing an edge to 6. For simplicity, we assume the agent alternates between the actions \emph{left} and \emph{right} across episodes. We show that \thesystem cycles indefinitely between $\rewardmachine_1$ (\Cref{fig:cycling_rm1}) and $\rewardmachine_2$ (\Cref{fig:cycling_rm2}):
\begin{enumerate}
	\item Initially, the agent goes left, reaches $\owcella$, and terminates. The following WCDPI is formed:
	\begin{align*}
		\left\langle 
		\mathtt{ex1}, 10, \langle \{\mathtt{accept}\}, \{\mathtt{reject}\} \rangle,
		\begin{Bmatrix*}[l]
			\mathtt{prop(}\owcella\mathtt{, 0).} \\ 
			\mathtt{step(0).} ~ \mathtt{last(0).}
		\end{Bmatrix*}
		\right\rangle.
	\end{align*}
	The example has penalty 10. Recall that penalties are proportional to the relative frequency of the example. ILASP learns $\rewardmachine_1$ from this example.
	
	\item The agent goes right, reaches $\owcellb$, and terminates. The learning task now contains two examples, each with penalty 5:
	\begin{align*}
		\left\langle 
		\mathtt{ex1}, 5, \langle \{\mathtt{accept}\}, \{\mathtt{reject}\} \rangle,
		\begin{Bmatrix*}[l]
			\mathtt{prop(}\owcella\mathtt{, 0).} \\ 
			\mathtt{step(0).} ~ \mathtt{last(0).}
		\end{Bmatrix*}
		\right\rangle,
	\end{align*}
	and
	\begin{align*}
		\left\langle 
		\mathtt{ex2}, 5, \langle \{\mathtt{accept}\}, \{\mathtt{reject}\} \rangle,
		\begin{Bmatrix*}[l]
			\mathtt{prop(}\owcellb\mathtt{, 0).} \\ 
			\mathtt{step(0).} ~ \mathtt{last(0).}
		\end{Bmatrix*}
		\right\rangle.
	\end{align*}
	
	The perfect RM (accepting both) would require adding an edge (cost 6), while ignoring $\mathtt{ex2}$ costs only 5. Since $\rewardmachine_1$ already satisfies this constraint, it is retained. Note that $\rewardmachine_2$ would be equally optimal, but RMs are only updated when the new candidate achieves a strictly better score.
	
	\item The agent goes left, reaches $\owcella$, and terminates. The current RM ($\rewardmachine_1$) captures the trace, so no relearning occurs.
	
	\item The agent goes right, reaches $\owcellb$, and terminates. Another trace for $\mathtt{ex2}$ is produced and its penalty (7) grows large enough that ignoring it becomes costlier than adding an edge. ILASP then learns $\rewardmachine_2$.
	
	\item Subsequently, the agent alternates between producing $\mathtt{ex1}$ and $\mathtt{ex2}$, with class rebalancing causing penalties to oscillate and triggering cyclic relearning as shown in \Cref{tab:cycling_examples}.
\end{enumerate}

\begin{table}[ht]
\centering
\caption{Example penalties and examples required for RM learning iterations in the cycling scenario.
Column ``Additional traces needed to trigger relearning'' captures the number of new traces collected between iterations $i-1$ and $i$;
$Y$ is a variable representing the number of $\mathtt{ex2}$-generating traces at iteration $2n-1$.}
\label{tab:cycling_examples}
\resizebox{\linewidth}{!}{%
\begin{tabular}{ccccccc}
\toprule
\# & RM learned & Penalty of $\mathtt{ex1}$ & Penalty of $\mathtt{ex2}$  & \makecell[c]{Number of traces\\generating $\mathtt{ex1}$ } & \makecell[c]{Number of traces\\generating $\mathtt{ex2}$ } & \makecell[c]{Additional traces \\ needed to \\ trigger relearning} \\
\midrule
1 & $\rewardmachine_1$ & $\frac{1}{1} \times 10 = 10$ & --- & 1 & 0 & 1 \\
2 & $\rewardmachine_2$ & $\frac{1}{3} \times 10 \approx 3$ & $\frac{2}{3} \times 10 \approx 7$ & 1 & 2 & 2 \\
3 & $\rewardmachine_1$ & $\frac{3}{5} \times 10 = 6$ & $\frac{2}{5} \times 10 = 4$ & 3 & 2 & 2 \\
4 & $\rewardmachine_2$ & $\frac{3}{7} \times 10 \approx 4$ & $\frac{4}{7} \times 10 \approx 6$ & 3 & 4 & 2 \\
5 & $\rewardmachine_1$ & $\frac{5}{9} \times 10 \approx 6$ & $\frac{4}{9} \times 10 \approx 4$ & 5 & 4 & 2  \\
6 & $\rewardmachine_2$ & $\frac{5}{12} \times 10 \approx 4$ & $\frac{7}{12} \times 10 \approx 6$ & 5 & 7 & 3 \\
\multicolumn{7}{c}{\vdots} \\
10 & $\rewardmachine_2$ & $\frac{15}{34} \times 10 \approx 4$ & $\frac{19}{34} \times 10 \approx 6$ & 15 & 19 & 7 \\
\multicolumn{7}{c}{\vdots} \\
100 & $\rewardmachine_2$ & $4$ & $6$ & $1.24 \times 10^{9}$ & $1.52 \times 10^{9}$ & $5.02 \times 10^{8}$ \\
\multicolumn{7}{c}{\vdots} \\
1000 & $\rewardmachine_2$ & $4$ & $6$ & $1.38 \times 10^{87}$ & $4.14 \times 10^{87}$ & $1.37 \times 10^{87}$ \\
\multicolumn{7}{c}{\vdots} \\
$2n-1$ & $\rewardmachine_1$ & $6$ & $4$ & $\left\lceil \frac{11}{9} \cdot Y \right\rceil$ & $Y$ & N/A\textsuperscript{a} \\
$2n$ & $\rewardmachine_2$ & $4$ & $6$ & $\left\lceil \frac{11}{9} \cdot Y \right\rceil$ & $\left\lceil \frac{11}{9} \cdot \left\lceil \frac{11}{9} \cdot Y \right\rceil \right\rceil$ &
\makecell[c]{
$\left\lceil \frac{11}{9} \cdot \left\lceil \frac{11}{9} \cdot Y \right\rceil \right\rceil - Y$ \\[0.5ex]
$\approx \left\lceil \frac{40}{81} Y \right\rceil$
} \\
\bottomrule
\end{tabular}}%
\par\smallskip
{\raggedright\small\quad\textsuperscript{a}\,Depends on the value at the previous row.\par}
\end{table}

This cycling behaviour is a direct consequence of using finite penalty budgets. With an infinite penalty budget, all examples would carry infinite weight, requiring complete coverage and reducing \thesystem to ISA, which provably converges~\cite{FurelosBlancoLJBR21}. The finite penalty budget enables trading off RM complexity against example coverage, but introduces the possibility of cyclic relearning.

Although infinite cycling is theoretically possible, it is unlikely in practice for several reasons. Even under a policy and hyperparameter configuration specifically designed to produce traces contradicting the current RM, an exponentially increasing number of traces is required to induce a meaningful shift in WCDPI penalties. Moreover, trained RL policies tend to exploit narrow trajectory distributions in later training stages, further reducing trace diversity and stabilising example distributions.

\section{Conclusions and Future Work}
\label{sec:conclusions}

We present \thesystem, a method for learning and exploiting reward machines from noisy labels. Our core contributions are: (1)~a mechanism for learning RMs from noisy traces using weighted examples; (2)~an algorithm interleaving RM and policy learning; and (3)~a comprehensive evaluation on discrete and continuous domains.

\thesystem outperforms recurrent baselines and achieves performance comparable to handcrafted RMs. Key challenges addressed include: specifying the learning task (balancing RM complexity against trace coverage via weighted examples), generating training examples (sampling from noisy traces and previous RMs), and determining when to trigger relearning (via RM error thresholding). Compared to our previous work~\cite{parac2024learning}, we introduce the complete ILASP encoding, integrate with PPO, add example generation from previous RMs, extend to continuous domains, and significantly expand the evaluation.

This work opens several promising research directions:
\begin{itemize}
	\item \emph{Hierarchical and First-Order RMs.} Extending to RM hierarchies~\cite{FurelosBlancoLJBR23} and first-order RMs~\cite{ardon2024texttt} could improve scalability and expressiveness.
	
	\item \emph{Human-AI Collaboration.} Learned RMs could facilitate understanding human strategies and interpreting AI intentions. Collecting sufficient human interaction traces for RM learning is challenging, making methods that can learn from limited human data particularly valuable for enabling practical human-AI collaboration systems.
	
	\item \emph{Instruction Following.} Recent work trains policies to generalise across instructions given as finite-state machines~\cite{YalcinkayaLVS23,YalcinkayaLVS24,FurelosBlancoPKSRD26} or linear temporal logic~\cite{vaezipoor2021ltl2action,leon2021nutshell,qiu2023instructing,jackermeier2024deepltl}. Our mechanisms (RM state beliefs, probabilistic reward shaping, and noise-aware learning) could enable instruction following under noisy labels, allowing deployment on complex benchmarks such as CALVIN~\cite{mees2022calvin} with realistic sensor noise.
	
	\item \emph{Labelling Function Learning.} Our work assumes the labelling function is given or pre-trained with a known proposition alphabet. Recent work learns the labelling function itself through environment interaction~\cite{umili2024neural}. Interleaving RM learning with labelling function learning is a key direction for future work.
\end{itemize}

% !TEX root = ../JAIR_Example_Template/JAIR_Example_Template.tex
\begin{acks}
    
Research was sponsored by the Army Research Laboratory and was accomplished under Cooperative Agreement Number W911NF-222-0243. The views and conclusions contained in this document are those of the authors and should not be interpreted as representing the official policies, either expressed or implied, of the Army Research Laboratory or the U.S. Government. The U.S. Government is authorised to reproduce and distribute reprints for Government purposes notwithstanding any copyright notation herein.

Roko Parac is supported by the UKRI Centre for Doctoral Training in Safe and Trusted AI (EPSRC Project EP/S023356/1). We thank Julian de Gortari Briseno for his help with the noisy sensor model and Andrew Li for helpful advice about his work.

\end{acks}

\printbibliography

\appendix

\section{Notation}
\label{app:notation}
\Cref{tab:notation} contains the notation used throughout the paper. 

{
\footnotesize
\begin{longtable}{p{2.5cm}p{8cm}}
	\caption{Summary of mathematical notation.} 
	\label{tab:notation} \\
	\toprule
	\textbf{Symbol} & \textbf{Description} \\
	\midrule
	\endfirsthead
	
	\multicolumn{2}{c}{\tablename\ \thetable\ -- continued} \\
	\toprule
	\textbf{Symbol} & \textbf{Description} \\
	\midrule
	\endhead
	
	\midrule
	\multicolumn{2}{r}{Continued on next page...} \\
	\endfoot
	
	\bottomrule
	\endlastfoot
	
	\multicolumn{2}{l}{\textbf{Reinforcement Learning and (Noisy) Labelled MDPs (\Cref{Reinforcementlearning,sec:noisy_labelled_mdps})}} \\
	$\states$ & Set of MDP states \\
	$\initialstate$ & Initial MDP state \\
	$\actions$ & Set of actions \\
    $\history$ & State-action history \\
	$\transitionprob$ & Probability transition function \\
	$\rewardfunction$ & Reward function \\
	$\termfunction$ & Termination function \\
	$\stateterm$ & Terminal history indicator \\
	$\stategoal$ & Goal history indicator \\
	$\discountfactor$ & Discount factor \\
    $\policy$ & Policy \\
	$\propositions$ & Set of propositions \\
    $\singleproposition$ & A single proposition \\
    $\singlepropositiondetected$ & Noisy detection  for proposition $\singleproposition$ \\
    $\observationlabel$ & Label (a subset of propositions) \\
	$\labelingfunction$ & Perfect labelling function \\
    $\noisylabelingfunction$ & Noisy labelling function \\
	$\noisysensordetections$ & Set of propositions detected by a noisy labelling function \\
    $\trace$ & Label trace \\
	$\noisytrace$ & Noisy label trace \\
    $[\noisylabelingfunction(\statee, \action, \statee')]_\singleproposition$ & Probability of observing proposition $\singleproposition$ on transition $\langle \statee, \action, \statee'\rangle$\\
	\midrule
	\multicolumn{2}{l}{\textbf{Reward Machines (\Cref{sec:background_rms})}} \\
	$\rewardmachine$ & Reward machine \\
	$\rmstates$ & Set of RM states \\
	$\initialrmstate$ & Initial RM state \\
	$\acceptingrmstate$ & Accepting RM state \\
	$\rejectingrmstate$ & Rejecting RM state \\
	$\rmtransitionfunction$ & RM state-transition function \\
	$\rmrewardtransitionfun$ & RM reward-transition function \\
	\midrule
	\multicolumn{2}{l}{\textbf{Inductive Learning of Answer Set Programs (\Cref{sec:background_ilasp})}} \\
    $B$ & Background knowledge of LAS task \\
    $\modebias$ & Hypothesis space of a LAS task \\
	$\exampleset$ & Set of WCDPI examples of a LAS task \\
	$\exid$ & Identifier of a WCDPI example \\
	$\expen$ & Penalty of a WCDPI example \\
	$\langle \exinc, \exexc\rangle$ & Partial interpretation of a WCDPI example \\
	$\exctx$ & Context of a WCDPI example \\
	$\hypothesis$ & Hypothesis (i.e., inductive solution of a LAS task) \\
	\midrule
	\multicolumn{2}{l}{\textbf{\thesystem (\Cref{sec:learning_rms_from_noisy_traces,section:rm_exploitation,sec:interleaving_learning_and_exploitation})}} \\
    $\sampletrace$ & Grounded trace (obtained by sampling from the noisy trace) \\
	$\numsamplespertrace$ & Number of grounded traces sampled from a noisy trace \\
    $\logicaltransitionfn$ & Logical transition function \\
	$\notlogicaltransitionfn$ & Negation of the logical transition function \\
    $\maxrmstates$ & Maximum set of states available to learn an RM \\
    $\maxedgesconst$ & Maximum number of edges from one state to another \\
    $\classweight$ & Class-based weight multiplier for WCDPIs, inversely proportional to class frequency \\
    $\traceoriginmultiplier$ & Trace origin multiplier\\
    $\penaltythreshold$ & Minimum WCDPI penalty\\
    $\rmstatebelief$ & RM state belief \\
	$\rmrewardtransitionfunbelief$ & Reward over RM state beliefs \\
	$\rmrelearningthreshold$ & RM relearning threshold \\
\end{longtable}
}

\section{Proof of \texorpdfstring{\Cref{proposition:rmrewardtransitionfun}}{Proposition 5.3}}
\label{derivation:simplification_with_ass_1}

\allowdisplaybreaks

Recall from \Cref{def:rmrewardtransitionfun} that the reward over RM state beliefs is:
\begin{align}
	\tag{\ref{eq:rm_reward_beliefs}}
	\rmrewardtransitionfunbelief(\rmstatebelief_t, \rmstatebelief_{t+1}) = \sum_{\substack{\rmstate_t, \rmstate_{t+1} \in \rmstates,\\ \observationlabel_t \in 2^{\propositions}}} P(\observationlabel_t \mid \statee_t, \action_t, \statee_{t+1}) \rmstatebelief_t(\rmstate_t) \mathbf{1}[\rmstatetransitionfun(\rmstate_t, \observationlabel_t) = \rmstate_{t+1}] \rmrewardtransitionfun(\rmstate_t, \rmstate_{t+1}),
\end{align}
where, under \Cref{assumption:reward}, the deterministic reward-transition function is:
\begin{align}
	\label{eq:rm_reward_deterministic_repeat}
	\rmrewardtransitionfun(\rmstate_t, \rmstate_{t+1}) =
	\begin{cases}
		1 &\text{if } \rmstate_t \neq \acceptingrmstate \text{ and } \rmstate_{t+1} = \acceptingrmstate; \\
		-1 &\text{if } \rmstate_t \neq \rejectingrmstate \text{ and } \rmstate_{t+1} = \rejectingrmstate; \\
		0 &\text{otherwise}.
	\end{cases}
\end{align}

The proof proceeds in three steps: first we expand the reward function using the deterministic reward structure, then we analyse how beliefs update for the accepting and rejecting states, and finally we combine these results to express the reward as a difference of belief updates.

\paragraph{Step 1: Expanding the Reward Function.}
We substitute $\rmrewardtransitionfun(\rmstate_t, \rmstate_{t+1})$ into $\rmrewardtransitionfunbelief(\rmstatebelief_t, \rmstatebelief_{t+1})$. Since the reward is non-zero only when $\rmstate_t \notin \{\acceptingrmstate, \rejectingrmstate\}$ and $\rmstate_{t+1} \in \{\acceptingrmstate, \rejectingrmstate\}$, we can restrict the summation:
\begin{align*}
	\rmrewardtransitionfunbelief(\rmstatebelief_t, \rmstatebelief_{t+1}) 
	&= \sum_{\substack{\rmstate_t \in \rmstates \setminus \{\acceptingrmstate, \rejectingrmstate\},\\ \rmstate_{t+1} = \acceptingrmstate,\\ \observationlabel_t \in 2^{\propositions}}} P(\observationlabel_t \mid \statee_t, \action_t, \statee_{t+1}) \rmstatebelief_t(\rmstate_t) \mathbf{1}[\rmstatetransitionfun(\rmstate_t, \observationlabel_t) = \rmstate_{t+1}] \cdot 1 \\
	&\quad + \sum_{\substack{\rmstate_t \in \rmstates \setminus \{\acceptingrmstate, \rejectingrmstate\},\\ \rmstate_{t+1} = \rejectingrmstate,\\ \observationlabel_t \in 2^{\propositions}}} P(\observationlabel_t \mid \statee_t, \action_t, \statee_{t+1}) \rmstatebelief_t(\rmstate_t) \mathbf{1}[\rmstatetransitionfun(\rmstate_t, \observationlabel_t) = \rmstate_{t+1}] \cdot (-1).
\end{align*}
Substituting the specific values $\rmstate_{t+1} = \acceptingrmstate$ and $\rmstate_{t+1} = \rejectingrmstate$ into their respective indicator functions:
\begin{align}
	\label{eq:appendix_reward_simplification}
	\rmrewardtransitionfunbelief(\rmstatebelief_t, \rmstatebelief_{t+1}) 
	&= \sum_{\substack{\rmstate_t \in \rmstates \setminus \{\acceptingrmstate, \rejectingrmstate\},\\ \observationlabel_t \in 2^{\propositions}}} P(\observationlabel_t \mid \statee_t, \action_t, \statee_{t+1}) \rmstatebelief_t(\rmstate_t) \mathbf{1}[\rmstatetransitionfun(\rmstate_t, \observationlabel_t) = \acceptingrmstate] \nonumber \\
	&\quad - \sum_{\substack{\rmstate_t \in \rmstates \setminus \{\acceptingrmstate, \rejectingrmstate\},\\ \observationlabel_t \in 2^{\propositions}}} P(\observationlabel_t \mid \statee_t, \action_t, \statee_{t+1}) \rmstatebelief_t(\rmstate_t) \mathbf{1}[\rmstatetransitionfun(\rmstate_t, \observationlabel_t) = \rejectingrmstate].
\end{align}

\paragraph{Step 2: Belief Updates for Terminal States.}
We now analyse how the belief in the accepting state evolves. From the belief update equation in \Cref{def:rm_state_belief}:
\begin{align}
	\tag{\ref{eq:rm_state_belief}}
	\rmstatebelief_{t+1}(\rmstate) = \sum_{\substack{\rmstate_t \in \rmstates,\\ \observationlabel_t \in 2^{\propositions}}}P(\observationlabel_t \mid \statee_t, \action_t, \statee_{t+1}) \rmstatebelief_t(\rmstate_t) \mathbf{1}[\rmstatetransitionfun(\rmstate_t, \observationlabel_t) = \rmstate].
\end{align}

For the accepting state $\acceptingrmstate$, we partition the summation over $\rmstate_t$ into three disjoint cases: non-terminal states, the accepting state itself, and the rejecting state:
\begin{align}
	\rmstatebelief_{t+1}(\acceptingrmstate) 
	&= \sum_{\substack{\rmstate_t \in \rmstates,\\ \observationlabel_t \in 2^{\propositions}}}P(\observationlabel_t \mid \statee_t, \action_t, \statee_{t+1}) \rmstatebelief_t(\rmstate_t) \mathbf{1}[\rmstatetransitionfun(\rmstate_t, \observationlabel_t) = \acceptingrmstate] \nonumber \\
	&= \sum_{\substack{\rmstate_t \in \rmstates \setminus \{\acceptingrmstate, \rejectingrmstate\},\\ \observationlabel_t \in 2^{\propositions}}} P(\observationlabel_t \mid \statee_t, \action_t, \statee_{t+1}) \rmstatebelief_t(\rmstate_t) \mathbf{1}[\rmstatetransitionfun(\rmstate_t, \observationlabel_t) = \acceptingrmstate] \label{eq:accepting_belief_expansion} \\
	&\quad+ \sum_{\observationlabel_t \in 2^{\propositions}} P(\observationlabel_t \mid \statee_t, \action_t, \statee_{t+1}) \rmstatebelief_t(\acceptingrmstate) \mathbf{1}[\rmstatetransitionfun(\acceptingrmstate, \observationlabel_t) = \acceptingrmstate] \nonumber \\
	&\quad+ 0. \quad \text{(no transition from $\rejectingrmstate$ to $\acceptingrmstate$)} \nonumber
\end{align}

The second term simplifies as follows:
\begin{align*}
	&\sum_{\observationlabel_t \in 2^{\propositions}} P(\observationlabel_t \mid \statee_t, \action_t, \statee_{t+1}) \rmstatebelief_t(\acceptingrmstate) \mathbf{1}[\rmstatetransitionfun(\acceptingrmstate, \observationlabel_t) = \acceptingrmstate] \\
	&= \rmstatebelief_t(\acceptingrmstate) \sum_{\observationlabel_t \in 2^{\propositions}} P(\observationlabel_t \mid \statee_t, \action_t, \statee_{t+1}) \mathbf{1}[\rmstatetransitionfun(\acceptingrmstate, \observationlabel_t) = \acceptingrmstate] \\
	&= \rmstatebelief_t(\acceptingrmstate) \sum_{\observationlabel_t \in 2^{\propositions}} P(\observationlabel_t \mid \statee_t, \action_t, \statee_{t+1}) \quad \text{(absorbing state: indicator always 1)} \\
	&= \rmstatebelief_t(\acceptingrmstate), \quad \text{(law of total probability)}
\end{align*}
where the second-to-last step uses the fact that $\rmstatetransitionfun(\acceptingrmstate, \observationlabel_t) = \acceptingrmstate$ for all $\observationlabel_t$, and the last step uses $\sum_{\observationlabel_t \in 2^{\propositions}} P(\observationlabel_t \mid \statee_t, \action_t, \statee_{t+1}) = 1$.

Substituting this back into \Cref{eq:accepting_belief_expansion} and rearranging:
\begin{align}
	\label{eq:belief_diff_ua}
	\rmstatebelief_{t+1}(\acceptingrmstate) -  \rmstatebelief_t(\acceptingrmstate) &= \sum_{\substack{\rmstate_t \in \rmstates \setminus \{\acceptingrmstate, \rejectingrmstate\},\\ \observationlabel_t \in 2^{\propositions}}} P(\observationlabel_t \mid \statee_t, \action_t, \statee_{t+1}) \rmstatebelief_t(\rmstate_t) \mathbf{1}[\rmstatetransitionfun(\rmstate_t, \observationlabel_t) = \acceptingrmstate].
\end{align}

By analogous reasoning for the rejecting state $\rejectingrmstate$:
\begin{align}
	\label{eq:belief_diff_ur}
	\rmstatebelief_{t+1}(\rejectingrmstate) -  \rmstatebelief_t(\rejectingrmstate) &= \sum_{\substack{\rmstate_t \in \rmstates \setminus \{\acceptingrmstate, \rejectingrmstate\},\\ \observationlabel_t \in 2^{\propositions}}} P(\observationlabel_t \mid \statee_t, \action_t, \statee_{t+1}) \rmstatebelief_t(\rmstate_t) \mathbf{1}[\rmstatetransitionfun(\rmstate_t, \observationlabel_t) = \rejectingrmstate].
\end{align}

\paragraph{Step 3: Final Substitution.}
Comparing \Cref{eq:appendix_reward_simplification} with \Cref{eq:belief_diff_ua,eq:belief_diff_ur}, we see that the two summations in the expression are exactly the belief differences. Therefore:
\begin{align*}
	\rmrewardtransitionfunbelief(\rmstatebelief_t, \rmstatebelief_{t+1}) = \left(\rmstatebelief_{t+1}(\acceptingrmstate) -  \rmstatebelief_t(\acceptingrmstate)\right) - \left(\rmstatebelief_{t+1}(\rejectingrmstate) -  \rmstatebelief_t(\rejectingrmstate)\right),
\end{align*}
which completes the proof.

\end{document}